\documentclass[journal]{IEEEtran}
\usepackage[utf8]{inputenc} \usepackage{amssymb} \usepackage{multirow} \usepackage{graphicx} \usepackage{tabularx} \usepackage[acronym]{glossaries} \usepackage{makecell} \usepackage{rotating} \usepackage{threeparttable} \usepackage{booktabs} \usepackage{ragged2e} \usepackage{array}
\usepackage{cite}
\usepackage{bm}
\usepackage{mathrsfs}
\usepackage{amsfonts}
\usepackage{amssymb}
\usepackage{amsmath}
\usepackage{latexsym}
\usepackage{graphicx}
\usepackage{subfigure}
\usepackage{amsthm}
\usepackage{hyperref}
\usepackage{indentfirst}
\usepackage{algorithm} 
\usepackage{algorithmic} 
\usepackage{multirow} 
\usepackage{xcolor}
\usepackage{color}
\usepackage{longtable}
\usepackage{graphics}
\usepackage{graphicx}
\usepackage{epsfig}
\usepackage[switch]{lineno}
\usepackage{hyperref}

\newcommand{\tabincell}[2]{\begin{tabular}{@{}#1@{}}#2\end{tabular}}

\usepackage{hyperref}
\usepackage[hyphenbreaks]{breakurl}

\usepackage{stfloats}

\theoremstyle{plain}

\theoremstyle{definition}

\theoremstyle{remark}

\ifCLASSINFOpdf
\else
\fi
\begin{document}
%
\title{Residual-Sparse Fuzzy \emph{C}-Means Clustering Incorporating Morphological Reconstruction and Wavelet frames}
%
%
%

\author{Cong~Wang,
Witold Pedrycz,~\IEEEmembership{Fellow,~IEEE},
ZhiWu Li,~\IEEEmembership{Fellow,~IEEE},
MengChu Zhou,~\IEEEmembership{Fellow,~IEEE},
and Jun Zhao,~\IEEEmembership{Member,~IEEE} 
\thanks{This work was supported in part by the Doctoral Students' Short Term Study Abroad Scholarship
Fund of Xidian University, in part by the National Natural Science Foundation of China under Grant Nos. 61873342, 61672400, in part by the Recruitment Program of Global Experts, and in part by the Science and Technology Development Fund, MSAR, under Grant No. 0012/2019/A1.} 
\thanks{C. Wang is with the School of Electro-Mechanical Engineering, Xidian University, Xi'an 710071, China (e-mail: wangc0705@stu.xidian.edu.cn).}
\thanks{W. Pedrycz is with the Department of Electrical and Computer Engineering, University of Alberta, Edmonton, AB T6R 2V4, Canada, the School of Electro-Mechanical Engineering, Xidian University, Xi'an 710071, China, and also with the Faculty of Engineering, King Abdulaziz University, Jeddah 21589, Saudi Arabia (e-mail: wpedrycz@ualberta.ca).}
\thanks{Z. Li is with the School of Electro-Mechanical Engineering, Xidian University, Xi'an 710071, China, and also with the Institute of Systems Engineering, Macau University of Science and Technology, Macau, China (e-mail: zhwli@xidian.edu.cn).}
\thanks{M. Zhou is with the Institute of Systems Engineering, Macau University  of Science and Technology, Macau 999078, China and also with the Helen and John C. Hartmann Department of Electrical and Computer Engineering, New Jersey Institute of Technology, Newark, NJ 07102 USA (e-mail: zhou@njit.edu).}
\thanks{J. Zhao is with School of Computer Science and Engineering, Nanyang Technological University, Singapore (e-mail: junzhao@ntu.edu.sg).}}

\maketitle
\begin{abstract}
Instead of directly utilizing an observed image including some outliers, noise or intensity inhomogeneity, the use of its ideal value (e.g. noise-free image) has a favorable impact on clustering. Hence, the accurate estimation of the residual (e.g. unknown noise) between the observed image and its ideal value is an important task. To do so, we propose an $\ell_0$ regularization-based Fuzzy \emph{C}-Means (FCM) algorithm incorporating a morphological reconstruction operation and a tight wavelet frame transform. To achieve a sound trade-off between detail preservation and noise suppression, morphological reconstruction is used to filter an observed image. By combining the observed and filtered images, a weighted sum image is generated. Since a tight wavelet frame system has sparse representations of an image, it is employed to decompose the weighted sum image, thus forming its corresponding feature set. Taking it as data for clustering, we present an improved FCM algorithm by imposing an $\ell_0$ regularization term on the residual between the feature set and its ideal value, which implies that the favorable estimation of the residual is obtained and the ideal value participates in clustering. Spatial information is also introduced into clustering since it is naturally encountered in image segmentation. Furthermore, it makes the estimation of the residual more reliable. To further enhance the segmentation effects of the improved FCM algorithm, we also employ the morphological reconstruction to smoothen the labels generated by clustering. Finally, based on the prototypes and smoothed labels, the segmented image is reconstructed by using a tight wavelet frame reconstruction operation. Experimental results reported for synthetic, medical, and color images show that the proposed algorithm is effective and efficient, and outperforms other algorithms.
\end{abstract}

%
\IEEEpeerreviewmaketitle

\section{Introduction}
\label{I}

\IEEEPARstart{S}{ince} its inception, a Fuzzy \emph{C}-Means (FCM) algorithm \cite{Dunn1971,Bezdek1981} has achieved much attention, and been applied to a wide range of research fields, such as granular computing \cite{Zhuxiubin2017}, pattern recognition \cite{Celik2013} and image analysis \cite{Bai2019}. However, the conventional FCM has a substantial flaw as it is not robust to observed images. To improve its robustness, its modified versions have been put forward by mainly introducing spatial information into its objective function \cite{Ahmed2002,Chen2004,Szilagyi2003,Cai2007,Krinidis2010} and substituting the Euclidean distance by kernel distances (functions) \cite{Gong2013,Elazab2015,Zhao2013,Guo2016,Zhao2014,Wang2019,Lin2014,Zhu2017}. As the first improvement, some classic FCM-related algorithms, such as FCM\_S \cite{Ahmed2002}, FCM\_S1 \cite{Chen2004}, FCM\_S2 \cite{Chen2004}, EnFCM \cite{Szilagyi2003} and FGFCM \cite{Cai2007}, have been proposed. Especially, Krinidis et al. \cite{Krinidis2010} report a fuzzy local information \emph{C}-means algorithm (FLICM) with the assistance of a fuzzy factor, which brings a simplified parameter setting. It yields better segmentation performance than previous algorithms. Nevertheless, only non-robust Euclidean distance is adopted in it, which is not effective for copying with the spatial information of images. In order to enhance its robustness, the second improvement has been investigated by using kernel distances. The essence of kernel distances (functions) is to transform the original data space into a new one. By making full use of superior properties of the new space, image data can be analyzed and manipulated easily. As a result, the use of kernel distances gives rise to such well-known FCM-related algorithms as KWFLICM \cite{Gong2013}, ARKFCM \cite{Elazab2015}, KGFCM \cite{Zhao2013}, NDFCM \cite{Guo2016} and NWFCM \cite{Zhao2014}. In particular, Wang et al. \cite{Wang2019} propose a wavelet frame-based FCM algorithm (WFCM) for addressing image segmentation problems defined in regular Euclidean and irregular domains. By considering tight wavelet frames as a kernel function, the image data characteristics are fully analyzed.

Recently, some comprehensive FCM-related algorithms have been presented \cite{Gharieb2017,Lei2018,Gu2018}, which involve various techniques. For instance, Gharieb et al. \cite{Gharieb2017} introduce a FCM framework by using Kullback-Leibler divergence to control the membership distance between a pixel and its neighbors. However, their algorithm is time-consuming and its segmentation effects can be further improved. Lei et al. \cite{Lei2018} report a new algorithm, namely FRFCM, by augmenting morphological reconstruction and membership filtering. It is fast thanks to using gray level histograms. However, its performance is sometimes unstable. More recently, Gu et al. \cite{Gu2018} introduce a fuzzy double C-Means algorithm (FDCM) incorporating sparse representation. It deals with two datasets simultaneously. The one is the basic feature set coming from an observed image, and the other is the feature set learned from a spare self-representation model. Although FDCM is robust to noise, its computational efficiency is low.

In fact, there usually exist some outliers, noise or intensity inhomogeneity in an observed image, which are produced by interferences during image acquisition and transmission. Such objects are usually modeled as the residual between the observed image and its ideal value (e.g. noise-free image). Intuitively, using the ideal value may benefit to segmentation effects of FCM. By briefly reviewing the literature, it has been found that most of the existing FCM-related algorithms utilize spatial information in observed images to suppress the residual present in fuzzy clustering. Yet no studies focus on in-depth analysis and manipulation of data components to develop an FCM algorithm. In other words, the residual is often not introduced into the objective function of FCM by separating it from the observed image. Moreover, a large proportion of image data has small or zero number of outliers, noise or intensity inhomogeneity. Therefore, the residual is very sparse. To take the sparsity of the residual into consideration, Zhang et al. \cite{Zhang2019} attempt to impose $\ell_{1} $ regularization on the residual, thus resulting in two alternative clustering algorithms, namely DSFCM and DSFCM\_N. Once spatial information has been encountered, DSFCM upgrades to DSFCM\_N that makes full use of spatial information. However, both algorithms fail to fully analyze the sparsity in the residual. As a result, their segmentation performance remains to be improved.

Motivated by \cite{Zhang2019}, we propose an $\ell_{0} $ regularization-based FCM algorithm incorporating a morphological reconstruction (MR) operation \cite{Najman1996,Vincent1993} and a tight wavelet frame transform \cite{Cai2012,Dong2010}.  The framework of the proposed algorithm is given in Fig. 1. It has the following steps:
\begin{figure}[htb]
\centering
\begin{minipage}[t]{0.9\linewidth}
\centering
\includegraphics[width=1\textwidth]{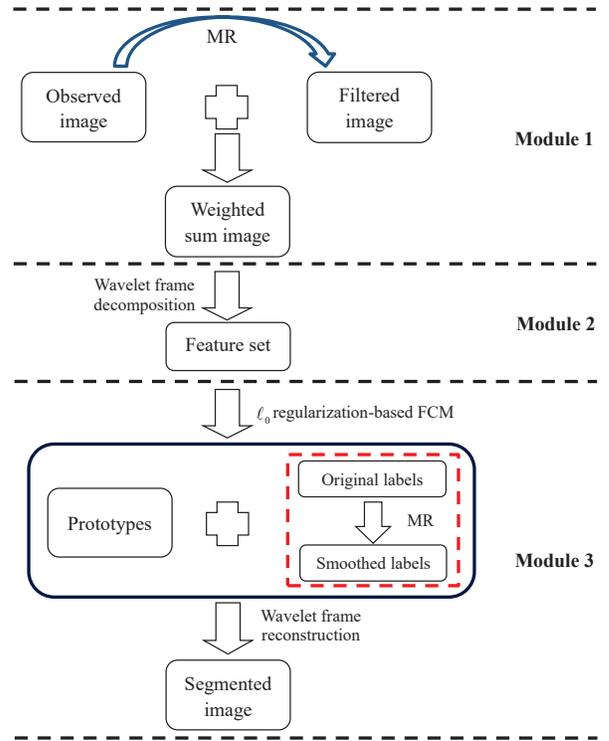}
\end{minipage}
\caption{The framework of the proposed algorithm.}
\end{figure}

As shown in Fig. 1, to achieve a good trade-off between detail preservation and noise suppression, MR is used to filter an observed image. By combining the observed and filtered images, a weighted sum image is generated, which contains less noise than the observed image but more features than the filtered image. To acquire sparse representations of the weighted sum image, a tight wavelet frame system is employed to decompose an image, thus resulting in the formation of its feature set. Considering the feature set as data to be clustered, we present an improved FCM algorithm by imposing an $\ell_{0}$ regularization term in the residual between the feature set and its ideal value, which implies that the ideal value estimated from the observed value participates in clustering in the light of the sparsity of the residual. Moreover, due to its capacity of noise suppression, the spatial information of image pixels is also considered into the objective function of FCM since it is naturally encountered in image segmentation. To further enhance the segmentation performance of the improved algorithm, MR is employed to smoothen the generated labels in clustering. Finally, by combining the prototypes obtained by the improved FCM algorithm and the smoothed labels, a segmented image is reconstructed by using a tight wavelet frame reconstruction operation.

This study makes three contributions to improve the segmentation performance of FCM. The first one is that MR is used to filter an observed image, thus forming a weighted sum image having some good properties. MR is applicable to cope with different types of noise, thus the noise is mainly suppressed and image feature details are retained well. Moreover, MR makes the distribution characteristic of pixels in the weighted sum image more favorable to fuzzy clustering, which is also positive to improve the speed of clustering.

The second contribution is to employ tight wavelet frames to form a feature set of a weighted sum image, which overcomes the drawback of the direct use of image pixels. The set is composed of redundant high and low frequency information. Therefore, underlying image details can be fully analyzed and manipulated. In fact, tight wavelet frames realize the transformation from a time domain to frequency one. Hence, they can be regarded as a kernel function, which means that the proposed approach is a kernel-based FCM algorithm.

Finally, we introduce an $\ell_{0} $ regularization term in the residual between the observed feature set corresponding to a weighted sum image and its ideal value into the objective function of FCM. The term makes the residual \mbox{accurately-estimated} by using its sparsity. It is equivalent to saying that the ideal value instead of just the observed value is used in clustering. Thus, the segmentation performance is improved in comparison with others without using it. Moreover, the spatial information contained in an image is simultaneously considered to modify the objective function, which aims to guarantee the close relationship between a target pixel and its neighbors.

In addition, we complete a label smoothing step by prudently using MR. Consequently, the segmentation performance is further improved. Therefore, this step can also be viewed as a contribution.

Overall, the originality of this work is to propose a comprehensive FCM algorithm by precisely estimating the residual and the assistance of various techniques. The essence of the proposed algorithm is a \mbox{kernel-based} FCM method with the aid of tight wavelet frames, which has better ability to identify features and noise in images. As a pre-processing step, MR removes a large proportion of noise and preserves main features, thus making a weighted sum image more favorable to clustering. Moreover, since $\ell_{0} $ regularization exhibits good properties related to the sparsity of the residual, the estimation of the residual is precisely realized. That is to say, the proposed algorithm conducts image segmentation by indirectly using the ideal value of an observed image.

This paper is organized as follows. FCM and a tight wavelet frame transform are briefed in Section \ref{II}. Section \ref{III} details the proposed algorithm. Section \ref{IV} reports experimental results for a set of images. Conclusions are drawn in Section \ref{V}.

\section{Preliminaries}
\label{II}
\subsection{Fuzzy C-Means (FCM) algorithm}
Given a data pattern ${\bm X}=\{ {\bm x}_{j}: j=1,2,\cdots,K\} \subset \mathbb{R}^{K} .$ Since each object ${\bm x}_{j} $ has $L$ attributes (channels), it is denoted as an $L$-dimensional vector $(x_{j1} ,x_{jl} ,\cdots ,x_{jL} )^{T} $. FCM splits $\bm X$ into $c$ prototypes by minimizing:
\begin{equation} \label{GrindEQ__1_}
J({\bm U},{\bm V})=\sum_{i=1}^{c}\sum_{j=1}^{K}u_{ij}^{m} \|{\bm x}_{j} -{\bm v}_{i} \|^{2}   ,
\end{equation}
subject to
\[\sum_{i=1}^{c}u_{ij}=1{\kern 1pt} {\kern 1pt} {\kern 1pt} {\kern 1pt} {\kern 1pt} {\kern 1pt}{\kern 1pt} \forall j\in \{ 1,2,\cdots ,K\} ,\]
where ${\bm U}=[u_{ij} ]_{c\times K} $ with $0\le u_{ij} \le 1$ is a partition matrix, ${\bm V}=\{\bm v_{i}\}_{i=1,2,\cdots ,c} $ is a set of $c$ prototypes, $\|\cdot\|$ denotes the Euclidean distance, and $m$ is a fuzzification coefficient ($m>1$).

The FCM algorithm provides an alternating iteration scheme to minimize \eqref{GrindEQ__1_}. Each iteration can be realized as follows \cite{Bezdek1984}:
\[u_{ij}^{(t+1)} =\frac{(\|{\bm x}_{j} -{\bm v}_{i}^{(t)} \|^{2} )^{-\frac{1}{m-1} } }{\sum\limits_{q=1}^{c}(\|{\bm x}_{j} -{\bm v}_{q}^{(t)} \|^{2} )^{-\frac{1}{m-1} } } ,\]
\[v_{il}^{(t+1)} =\frac{\sum\limits_{j=1}^{K}\left(u_{ij}^{(t+1)} \right)^{m} x_{jl} }{\sum\limits_{j=1}^{K}\left(u_{ij}^{(t+1)} \right)^{m} } .\]
Here, $t$ stands for the $t$-th iteration. By presetting a threshold $\varepsilon$, the iterative process is terminated when $\|{\bm U}^{(t+1)} -{\bm U}^{(t)} \|<\varepsilon $.

\subsection{Tight Wavelet Frame Transform}

Since tight wavelet frames can provide redundant representations of image data and exhibit substantial ability for feature/texture extraction, they have been successfully applied to various research areas, such as image segmentation \cite{Wang2019,Liu2015}, image denoising \cite{Dong2017,Wang2020denoising,Wang2017}, image restoration \cite{Cai2012,Ma2018}, and mesh surface reconstruction \cite{Dong2016,Yang2017}. For simplicity, we present the main idea of a tight wavelet frame transform concisely. Its more details can be found in \cite{Dong2010,Cai2012}. Generally speaking, it consists of two operators, i.e., decomposition ${\rm {\mathcal W}}$ and reconstruction ${\rm {\mathcal W}}^{T}$. By presetting a set of filters (masks), some sub-filtering operators are generated, i.e., ${\rm {\mathcal W}}_{0} ,{\rm {\mathcal W}}_{1} ,{\rm {\mathcal W}}_{2} \cdots $, which make up decomposition ${\rm {\mathcal W}}$. More specifically, ${\rm {\mathcal W}}_{0} $ is a low-pass filtering operator and the rest are high-pass filtering operators. According to unitary extension principle \cite{ARon1997}, reconstruction ${\rm {\mathcal W}}^{T} $ is available. Therefore, we have ${\rm {\mathcal W}}^{T} {\rm {\mathcal W}}={\rm {\mathcal I}}$, where ${\rm {\mathcal I}}$ is an identity matrix \cite{Dong2010}.

\section{Proposed Methodology}
\label{III}
\subsection{Image Filtering via MR}
Being superior to many usual filtering operations, such as mean filtering, median filtering, and Gaussian filtering, an MR operation exhibits the sound noise-immunity and retention capacity of image details. In light of superiority of MR, we employ it to filter an observed image in advance, thus resulting in reduction in the impact of the residual on clustering as much as possible. Formally speaking, the residual $\eta $ between an observed image $g$ and its ideal value $\tilde{g}$ is expressed as
\[\eta =g-\tilde{g}.\]

To reduce $\eta $ from $g$, MR is used. Generally speaking, MR consists of two basic operators, i.e., dilation and erosion reconstructions \cite{Chen2012}. Based on their combination, two reconstruction operators are usually obtained, i.e., morphological opening reconstruction and morphological closing one. Since the latter is more applicable for smoothing texture details, it is employed to filter $g$. Here, we denote its morphological closing reconstruction as
\begin{equation} \label{GrindEQ__2_}
{\rm {\mathcal R}}^{C} (g)={\rm {\mathcal R}}_{{\rm {\mathcal R}}_{g}^{{\rm {\mathcal D}}} ({\rm {\mathcal E}}(g))}^{{\rm {\mathcal E}}} ({\rm {\mathcal D}}({\rm {\mathcal R}}_{g}^{{\rm {\mathcal D}}} ({\rm {\mathcal E}}(g)))),
\end{equation}
where ${\rm {\mathcal R}}_{}^{{\rm {\mathcal D}}} $ is the dilation reconstruction of an image based on a dilation operation ${\rm {\mathcal D}}$, ${\rm {\mathcal R}}_{}^{{\rm {\mathcal E}}} $ is the erosion reconstruction of an image based on an erosion operation ${\rm {\mathcal E}}$. ${\rm {\mathcal D}}$ and ${\rm {\mathcal E}}$ are realized by using a flat structuring element. The readers are referred to \cite{Vincent1993,Chen2012}. In the sequel, we denote the filtered image as $\bar{g}={\rm {\mathcal R}}^{C} (g)$. To exhibit the performance of MR, we refer to Fig. 2 as an example. Here, a square of size $3\times 3$ is took as the structuring element.

\begin{figure}[htb]
\centering
\begin{minipage}[t]{0.19\linewidth}
\centering
\includegraphics[width=1\textwidth]{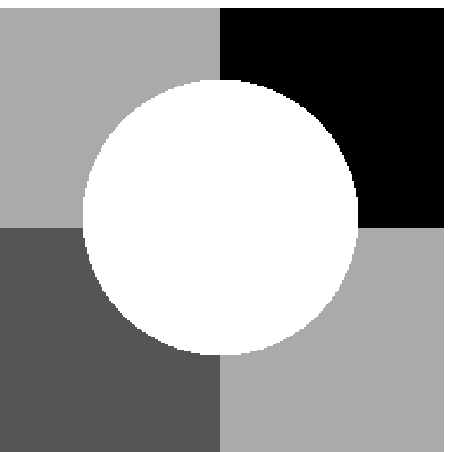}
\end{minipage}
\begin{minipage}[t]{0.19\linewidth}
\centering
\includegraphics[width=1\textwidth]{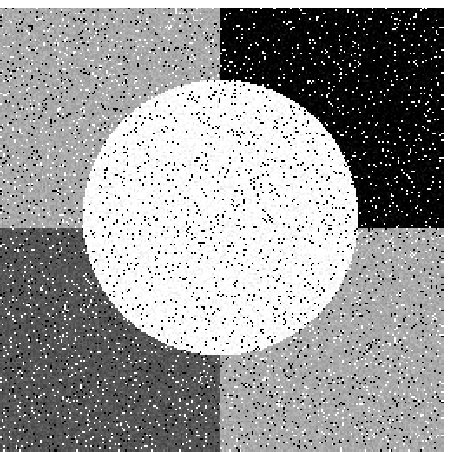}
\end{minipage}
\begin{minipage}[t]{0.19\linewidth}
\centering
\includegraphics[width=1\textwidth]{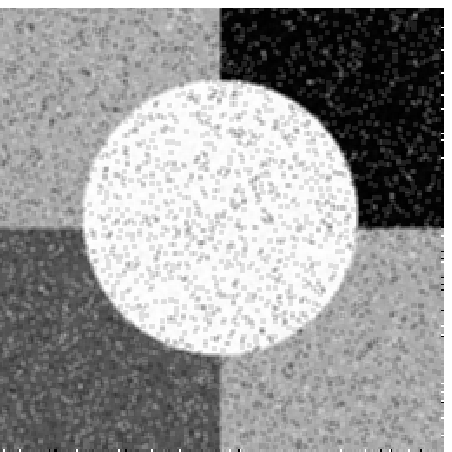}
\end{minipage}
\begin{minipage}[t]{0.19\linewidth}
\centering
\includegraphics[width=1\textwidth]{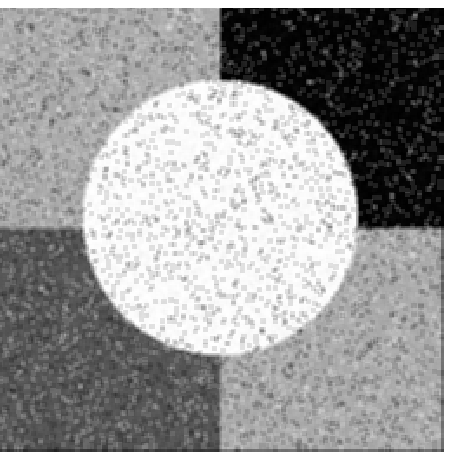}
\end{minipage}
\begin{minipage}[t]{0.19\linewidth}
\centering
\includegraphics[width=1\textwidth]{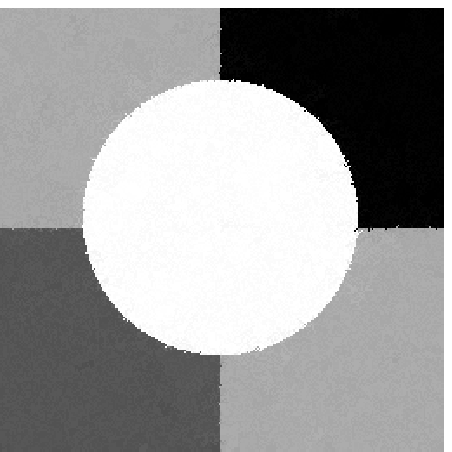}
\end{minipage}
\begin{minipage}[t]{0.19\linewidth}
\centering
\includegraphics[width=1\textwidth]{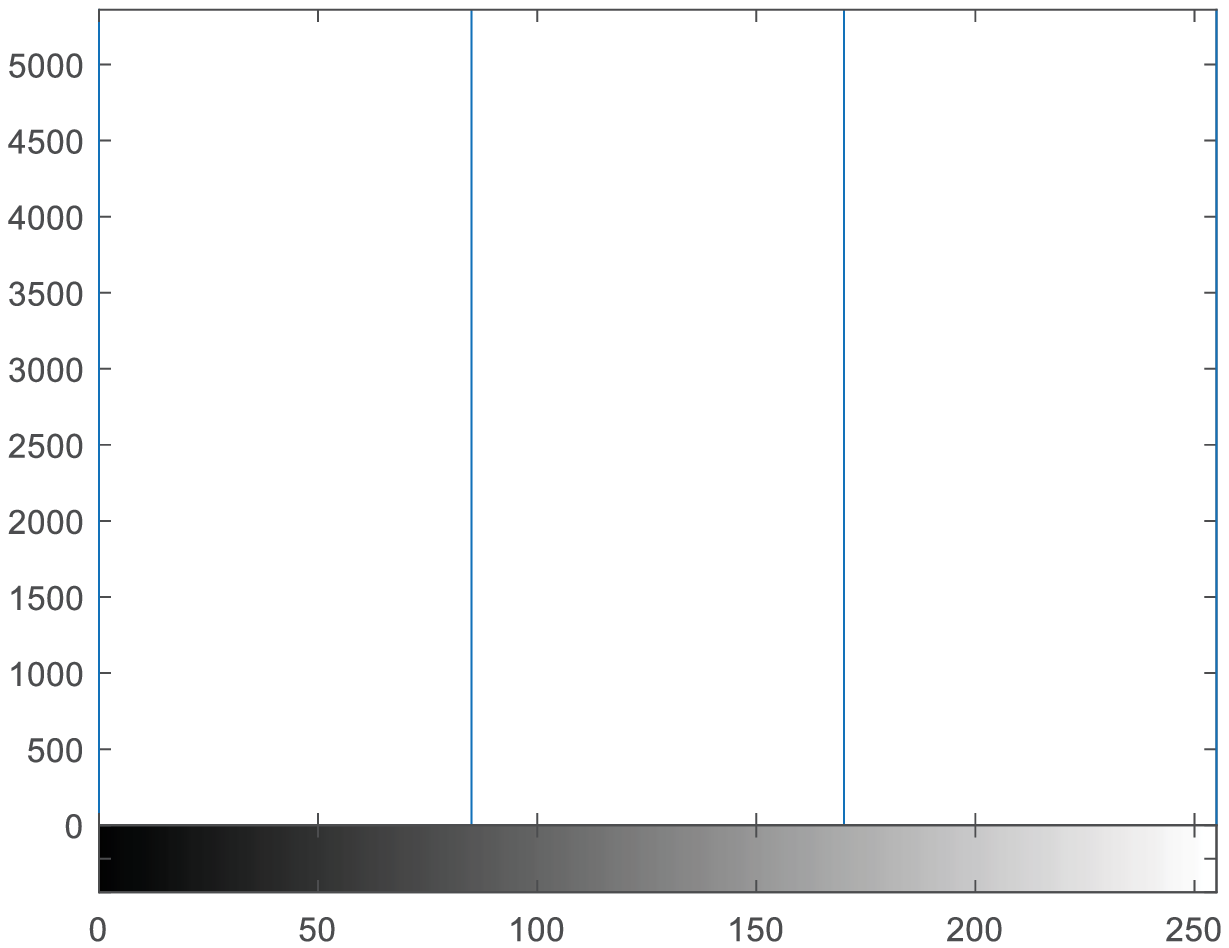}
\centerline{ \footnotesize (a)}
\end{minipage}
\begin{minipage}[t]{0.19\linewidth}
\centering
\includegraphics[width=1\textwidth]{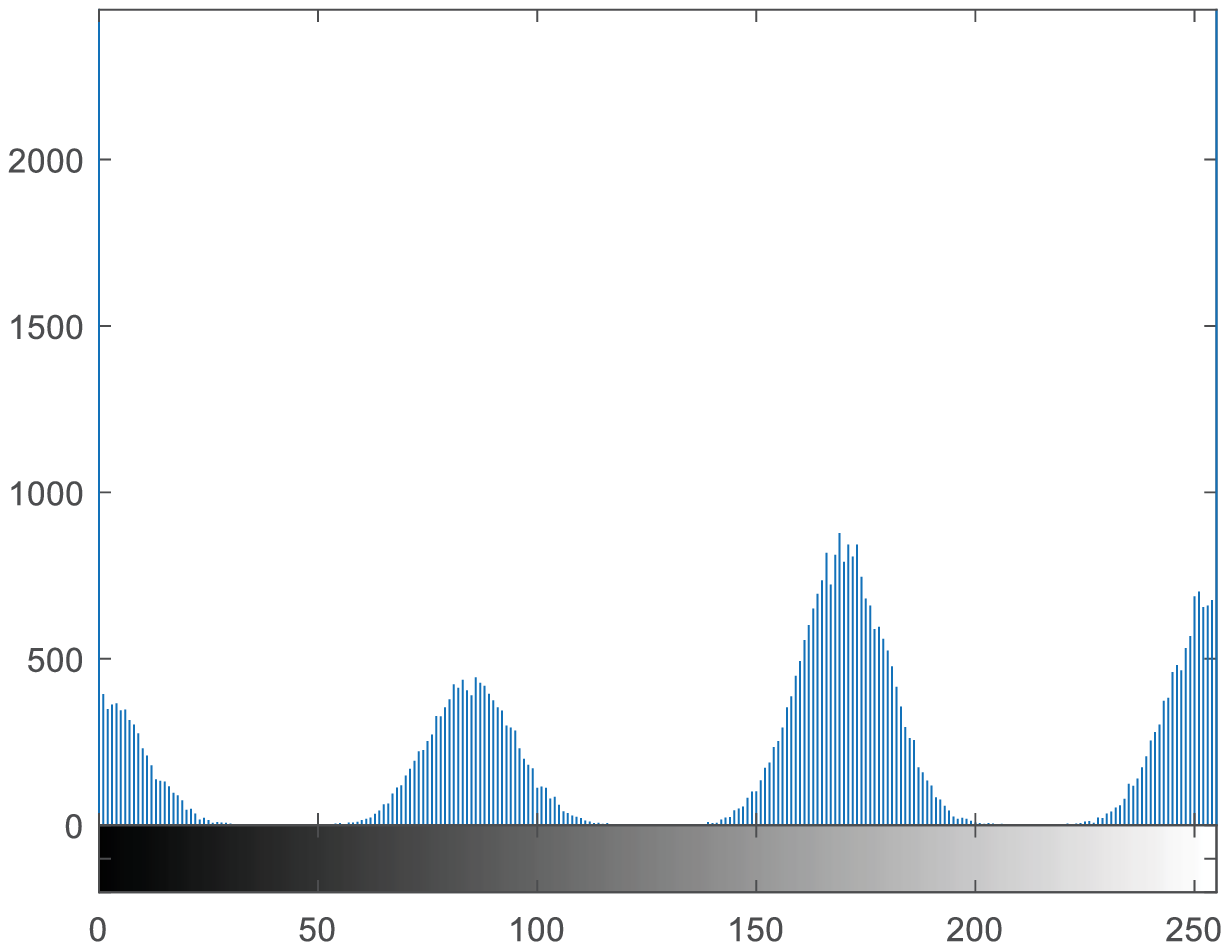}
\centerline{ \footnotesize (b)}
\end{minipage}
\begin{minipage}[t]{0.19\linewidth}
\centering
\includegraphics[width=1\textwidth]{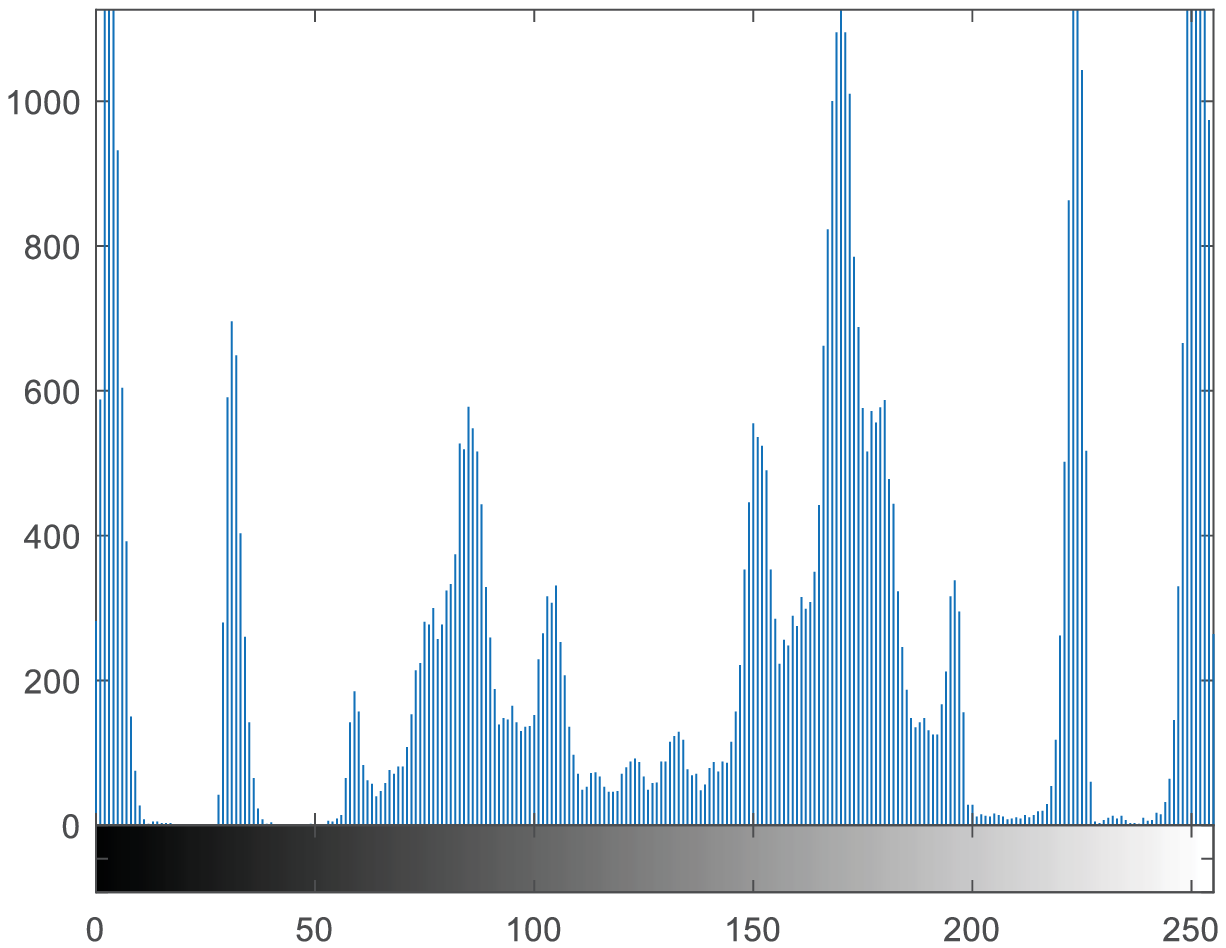}
\centerline{ \footnotesize (c)}
\end{minipage}
\begin{minipage}[t]{0.19\linewidth}
\centering
\includegraphics[width=1\textwidth]{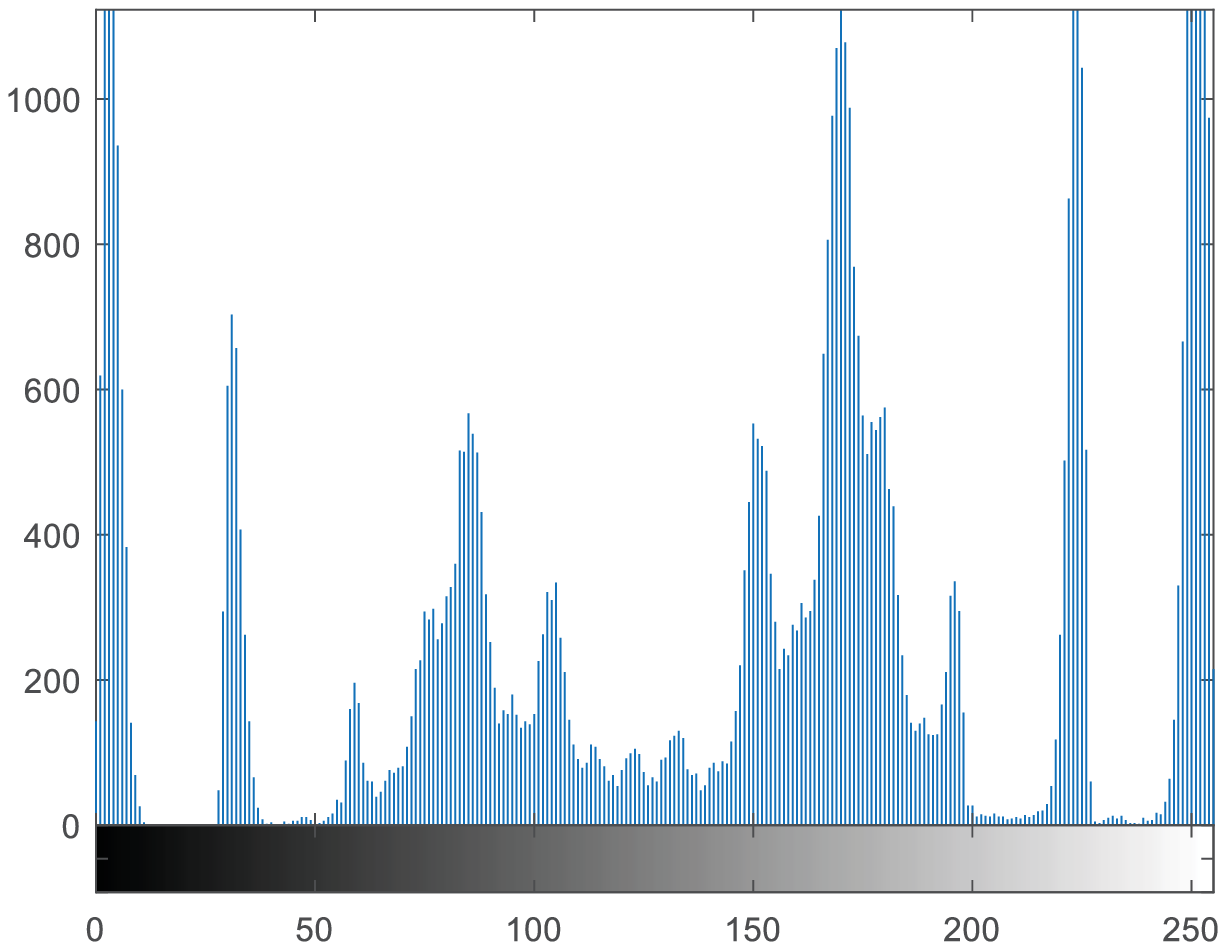}
\centerline{ \footnotesize (d)}
\end{minipage}
\begin{minipage}[t]{0.19\linewidth}
\centering
\includegraphics[width=1\textwidth]{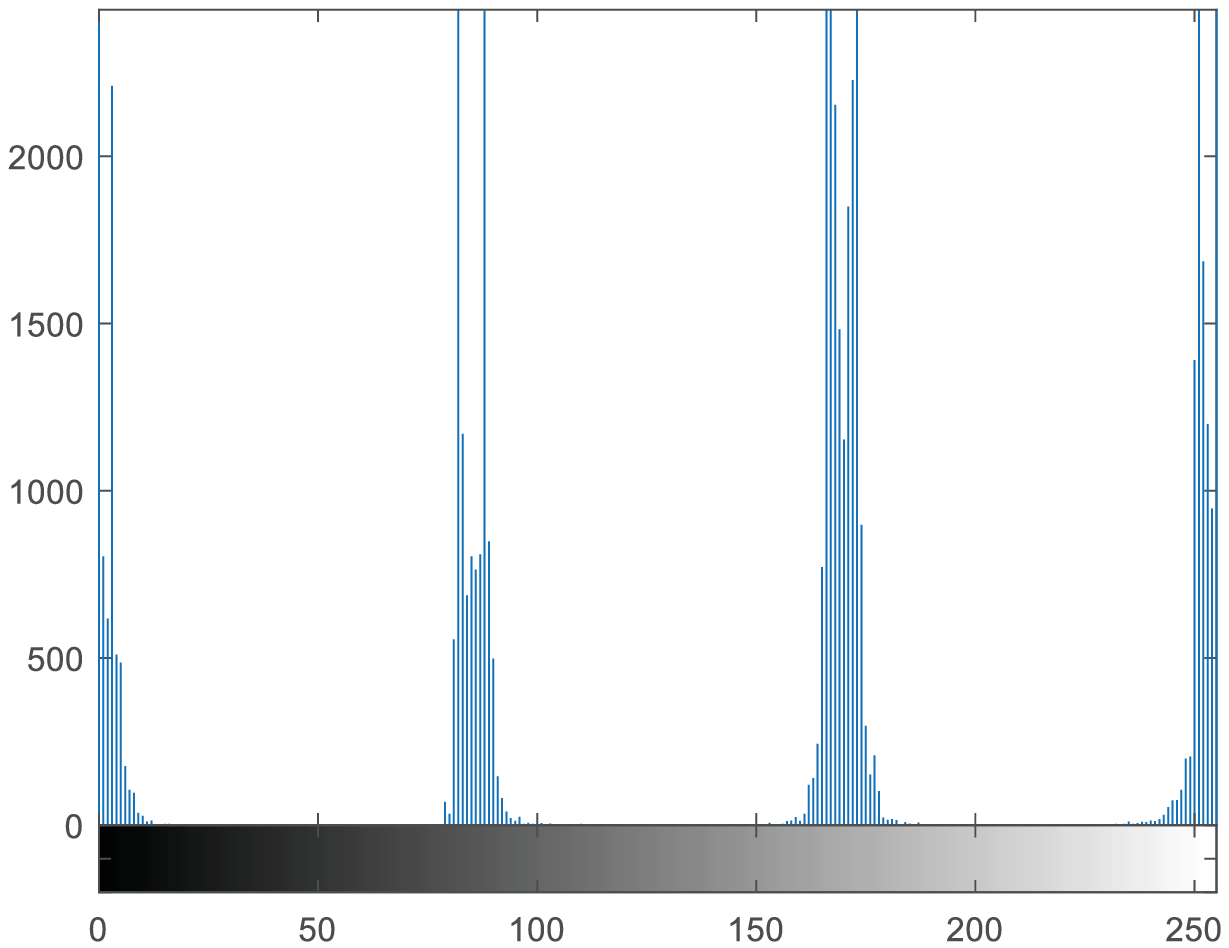}
\centerline{ \footnotesize (e)}
\end{minipage}
\caption{Comparison with different filtering methods. (a) Original image. (b) Observed image. (c) Filtered result using a mean filter. (d) Filtered result using a median filter. (e) Filtered result using MR.}
\end{figure}

As shown in Fig. 2, the second row stands for the corresponding gray level histograms of five images presented in the first row. Obviously, the original image includes four gray levels, i.e., 0, 85, 170, and 255, while its gray level~histogram has four obvious peaks. The observed image is corrupted by mixed Gaussian and impulse noise ($\textrm{standard~deviation}=10$, $\textrm{density}=10\%$). Its histogram has only two apparent peaks, i.e., 0 and 255. As Figs. 2(c)--(e) indicate, MR clusters image pixels into 4 groups while both mean and median filters produce too many peaks. In summary, MR is superior to usual filters since it can effectively remove noise and retain image details.

In this work, by combining $g$ and $\bar{g}$, we define a weighted sum image $\hat{g}$ as
\begin{equation} \label{GrindEQ__3_}
\hat{g}=\frac{g+\alpha \bar{g}}{1+\alpha } .
\end{equation}
Generally speaking, $\hat{g}$ contains less noise than $g$ and more features than $\bar{g}$. In addition, to express the composition of $\hat{g}$ explicitly, we rewrite $\hat{g}$ as
\begin{equation} \label{GrindEQ__4_}
\hat{g}=\eta '+g',
\end{equation}
where $g'$ stands for the ideal value of $\hat{g}$ and $\eta '$ is the residual between $\hat{g}$ and $g'$.

\subsection{Feature Extraction via Wavelet Frames}
After weighted sum image $\hat{g}$ is obtained, we generate its feature set by using a tight wavelet frame system. Due to its simplicity and practicability, a piecewise linear B-spline tight frame system \cite{ARon1997,Yang2018} is popular for feature extraction and redundant representations of images. We choose this system in which three one-dimensional filters are discretized as
$$
a_{0} =\left[\frac{1}{4},\frac{1}{2} ,\frac{1}{4} \right],a_{1} =\left[-\frac{1}{4} ,\frac{1}{2} ,-\frac{1}{4} \right],a_{2} =\left[\frac{\sqrt{2} }{4} ,0,-\frac{\sqrt{2} }{4} \right].
$$

By employing the above filters, we generate nine \mbox{two-dimensional} filters so as to construct nine corresponding sub-filtering operators ${\rm {\mathcal W}}_{0} {\rm {\mathcal W}}_{1} \cdots {\rm {\mathcal W}}_{8} $. They make up a tight wavelet frame decomposition operator ${\rm {\mathcal W}}.$ More specifically, ${\rm {\mathcal W}}_{0} $ is a low-pass filtering operator, and the remaining are high-pass filtering ones. As a result, we use ${\rm {\mathcal W}}$ to form the feature set associated with the weighted sum image, i.e.,
\begin{equation} \label{GrindEQ__5_}
{\bm X}={\rm {\mathcal W}}\hat{g}.
\end{equation}

For image processing, ${\rm {\mathcal W}}_{0} \hat{g}$ is a wavelet coefficient that represents low frequency information; while the rest represents high frequency one. The wavelet coefficients are acquired by using the 1-level tight wavelet frame decomposition. When the level becomes higher, the decomposition operation is repeatedly applied to the obtained coefficients. To exhibit the effect of wavelet frames, we show an example in Fig. 3.
\begin{figure}[htb]
\centering
\begin{minipage}[t]{0.19\linewidth}
\centering
\includegraphics[width=1\textwidth]{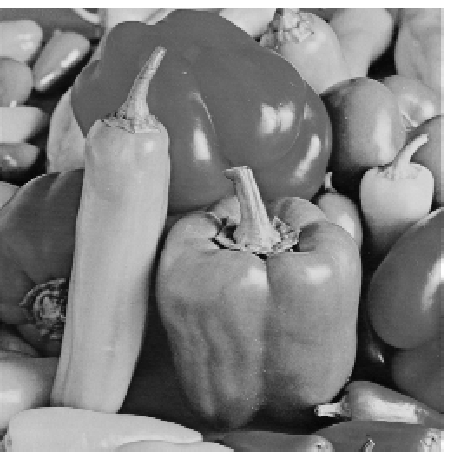}
\centerline{\footnotesize (a)}
\end{minipage}
\begin{minipage}[t]{0.19\linewidth}
\centering
\includegraphics[width=1\textwidth]{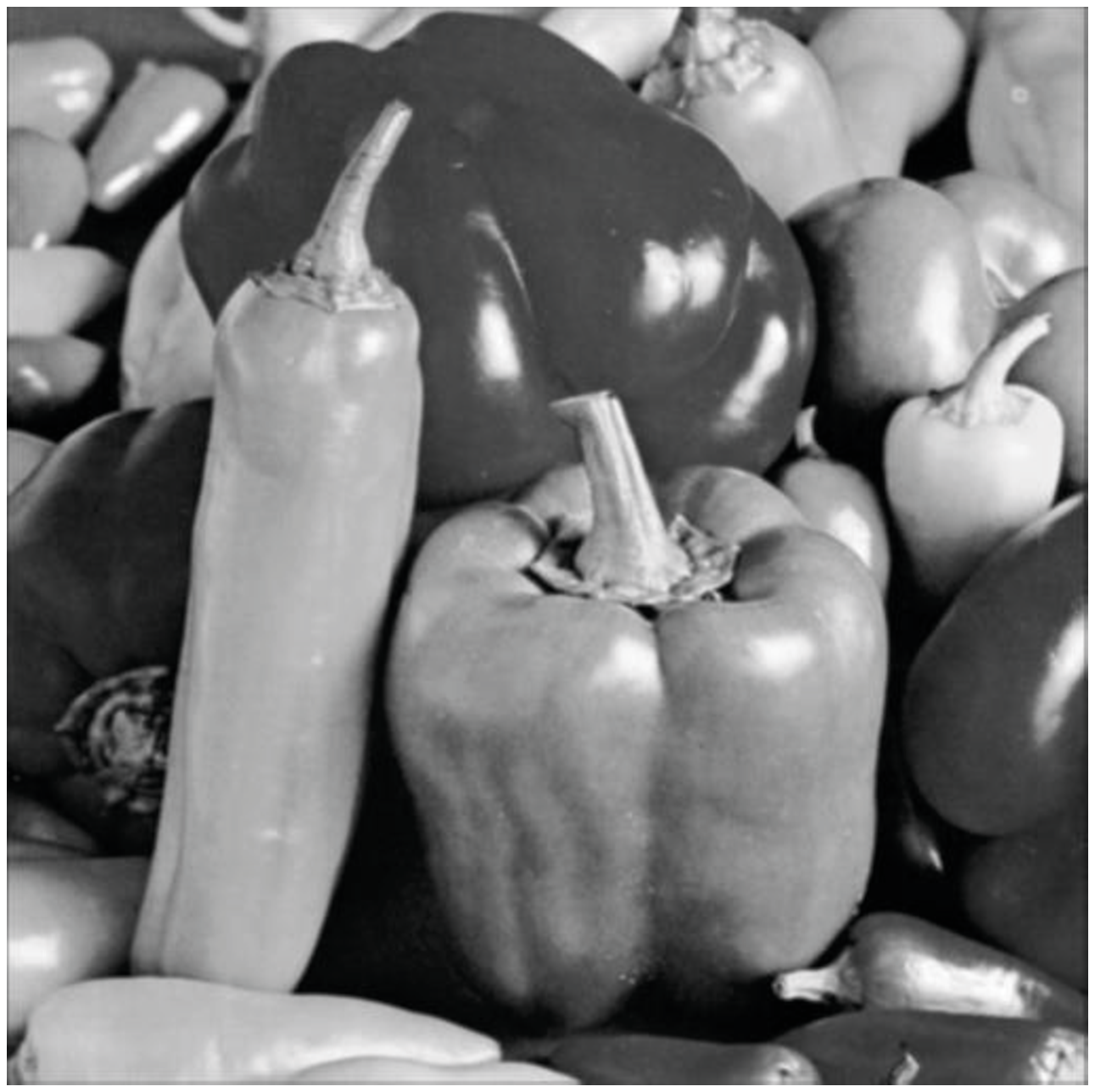}
\centerline{\footnotesize (b)}
\end{minipage}
\begin{minipage}[t]{0.19\linewidth}
\centering
\includegraphics[width=1\textwidth]{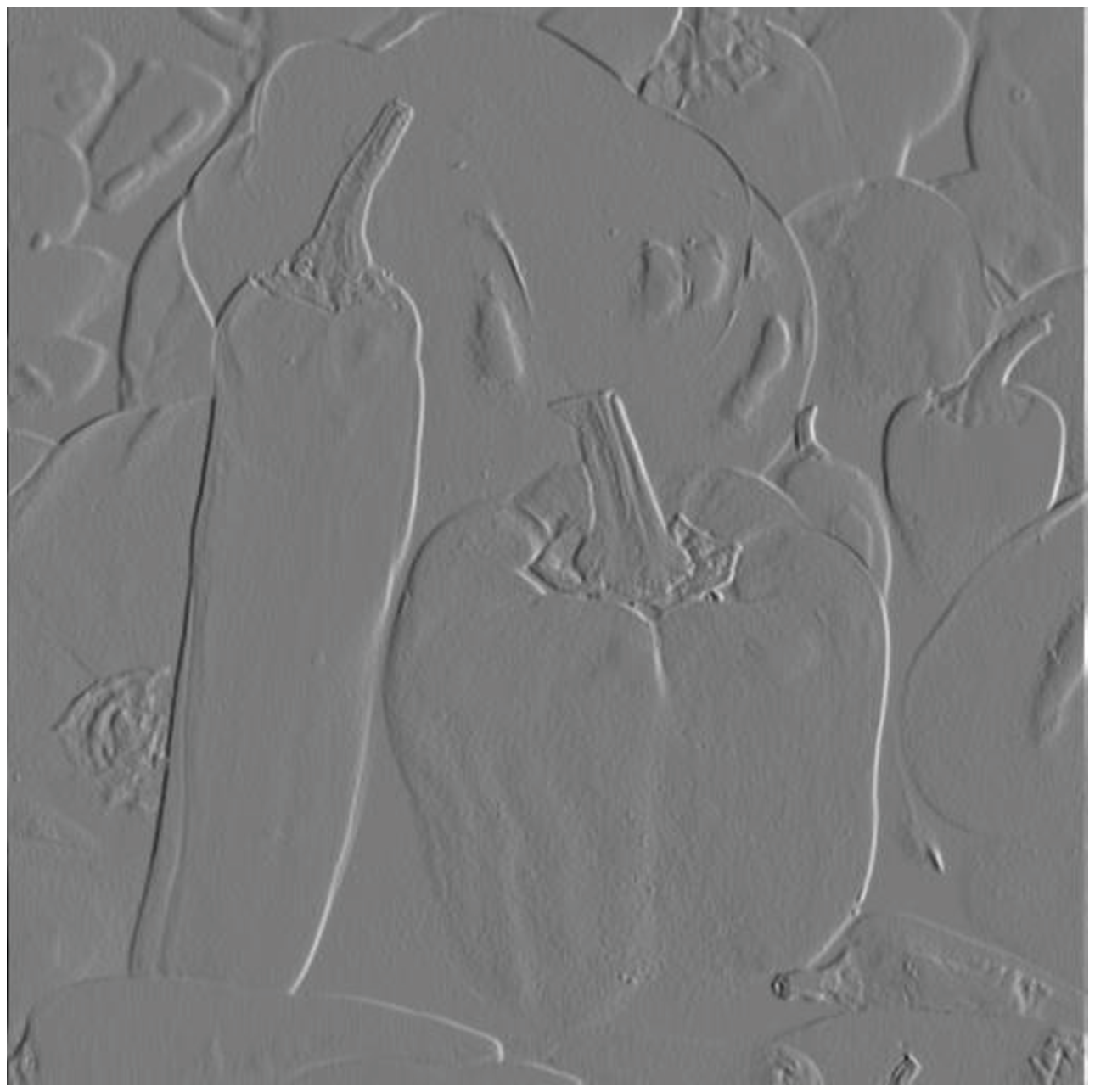}
\centerline{\footnotesize (c)}
\end{minipage}
\begin{minipage}[t]{0.19\linewidth}
\centering
\includegraphics[width=1\textwidth]{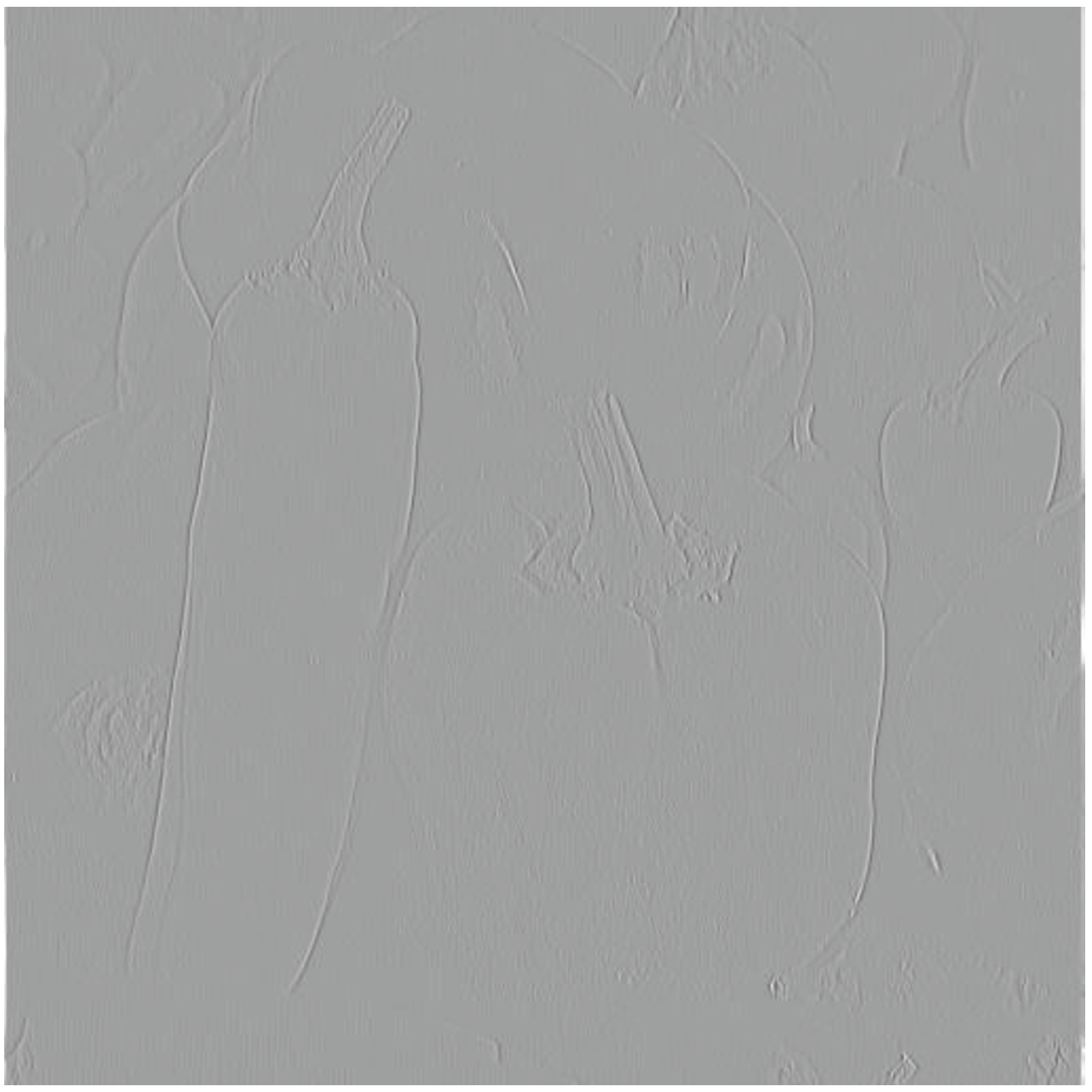}
\centerline{\footnotesize (d)}
\end{minipage}
\begin{minipage}[t]{0.19\linewidth}
\centering
\includegraphics[width=1\textwidth]{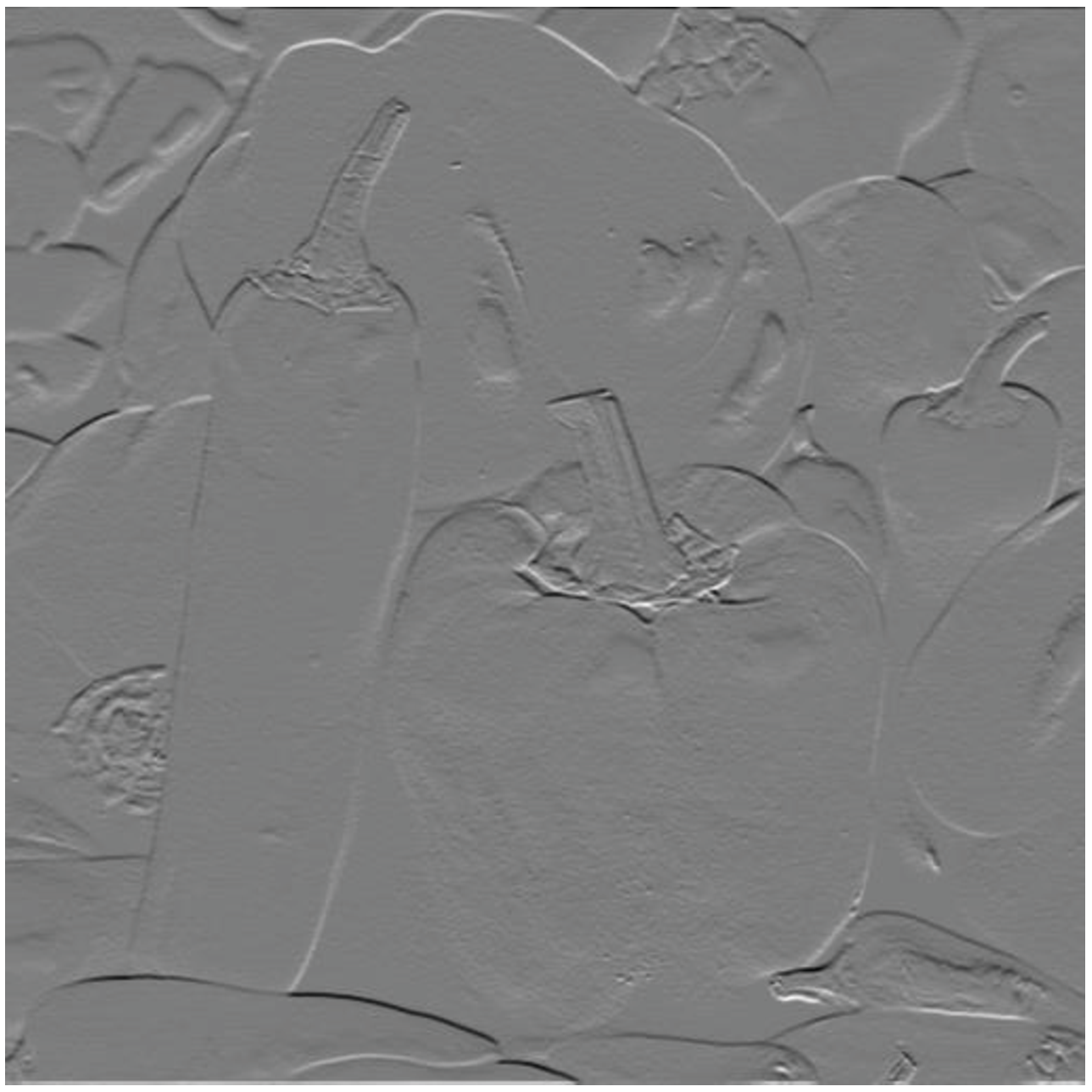}
\centerline{\footnotesize (e)}
\end{minipage}
\begin{minipage}[t]{0.19\linewidth}
\centering
\includegraphics[width=1\textwidth]{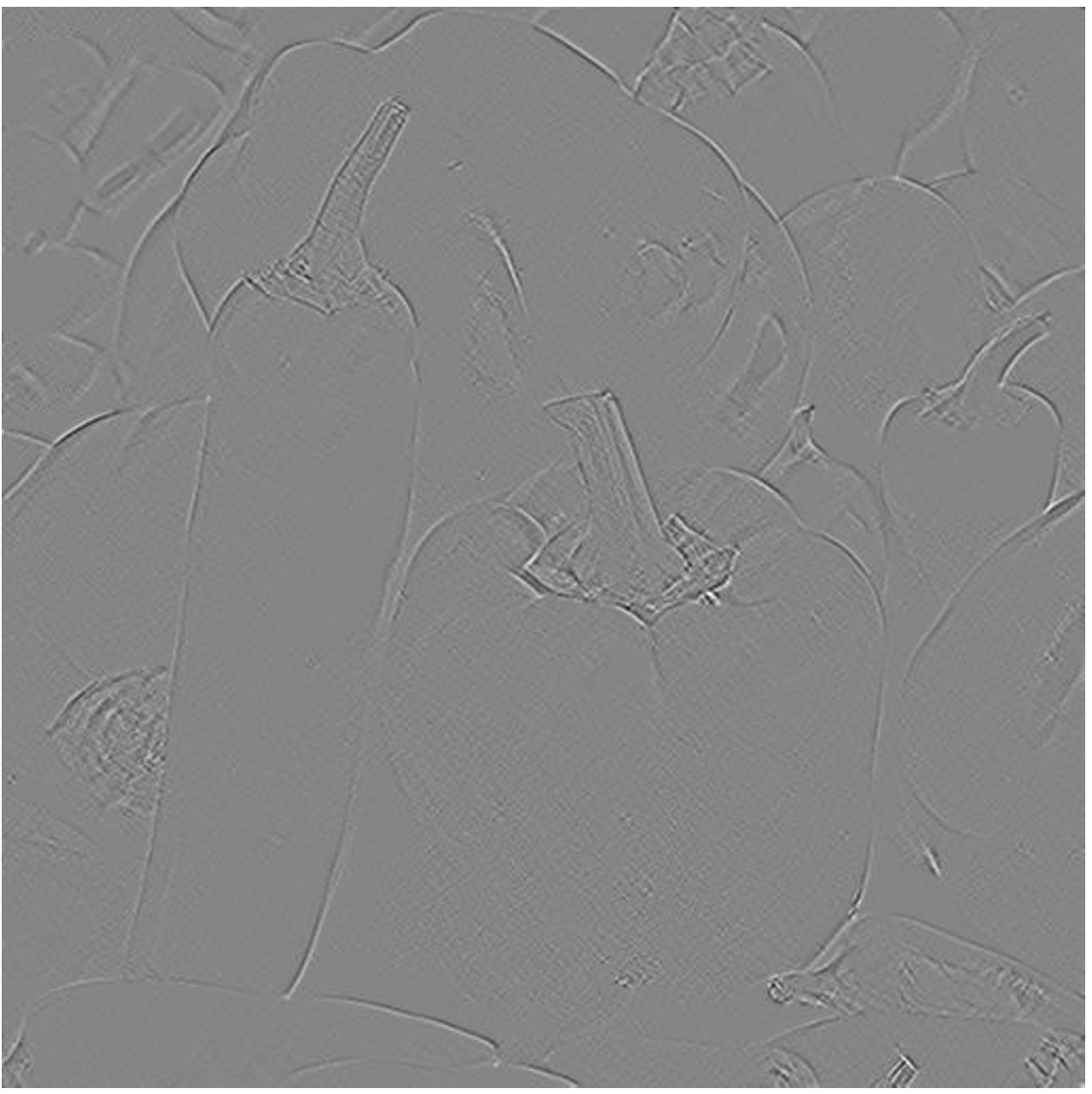}
\centerline{ \footnotesize (f)}
\end{minipage}
\begin{minipage}[t]{0.19\linewidth}
\centering
\includegraphics[width=1\textwidth]{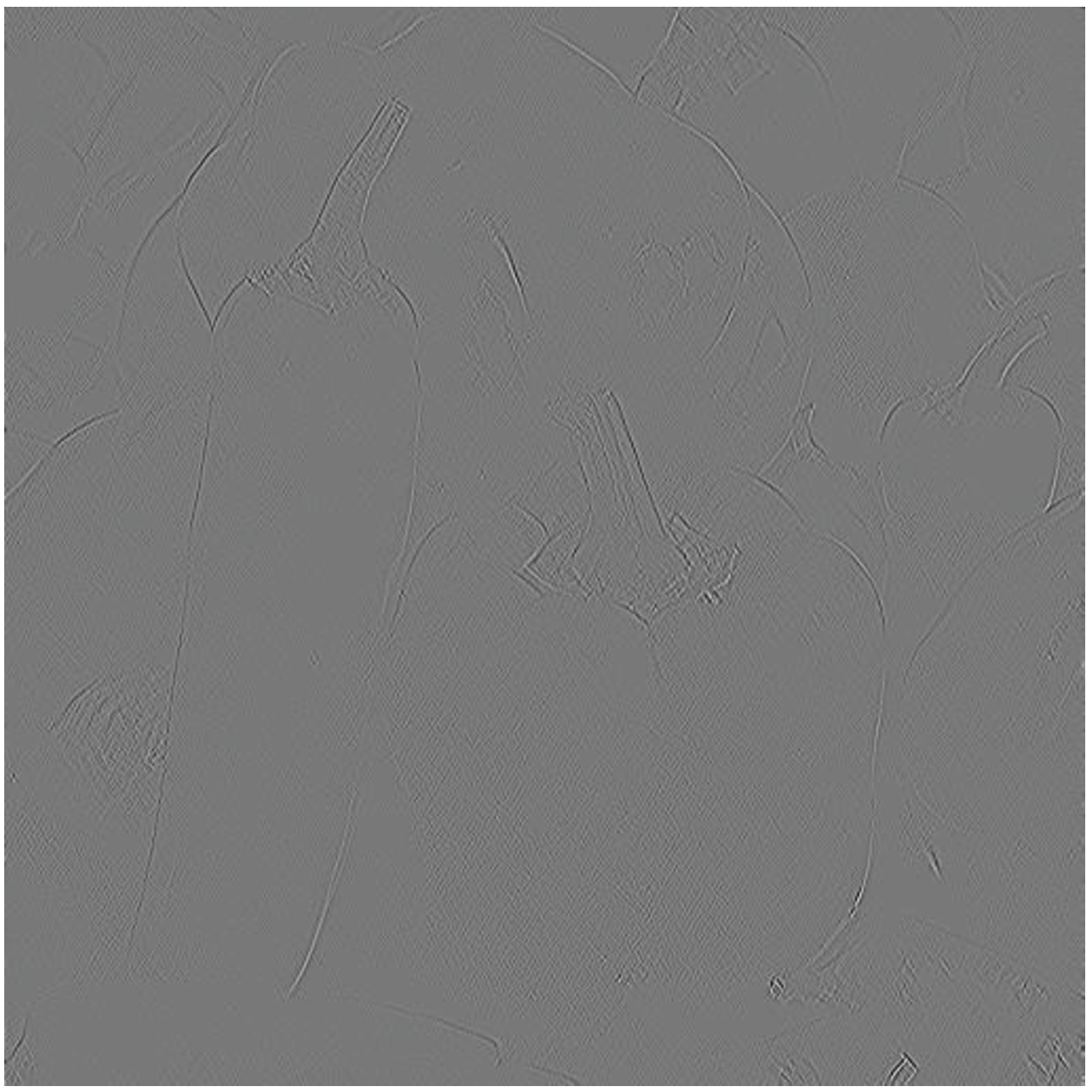}
\centerline{ \footnotesize (g)}
\end{minipage}
\begin{minipage}[t]{0.19\linewidth}
\centering
\includegraphics[width=1\textwidth]{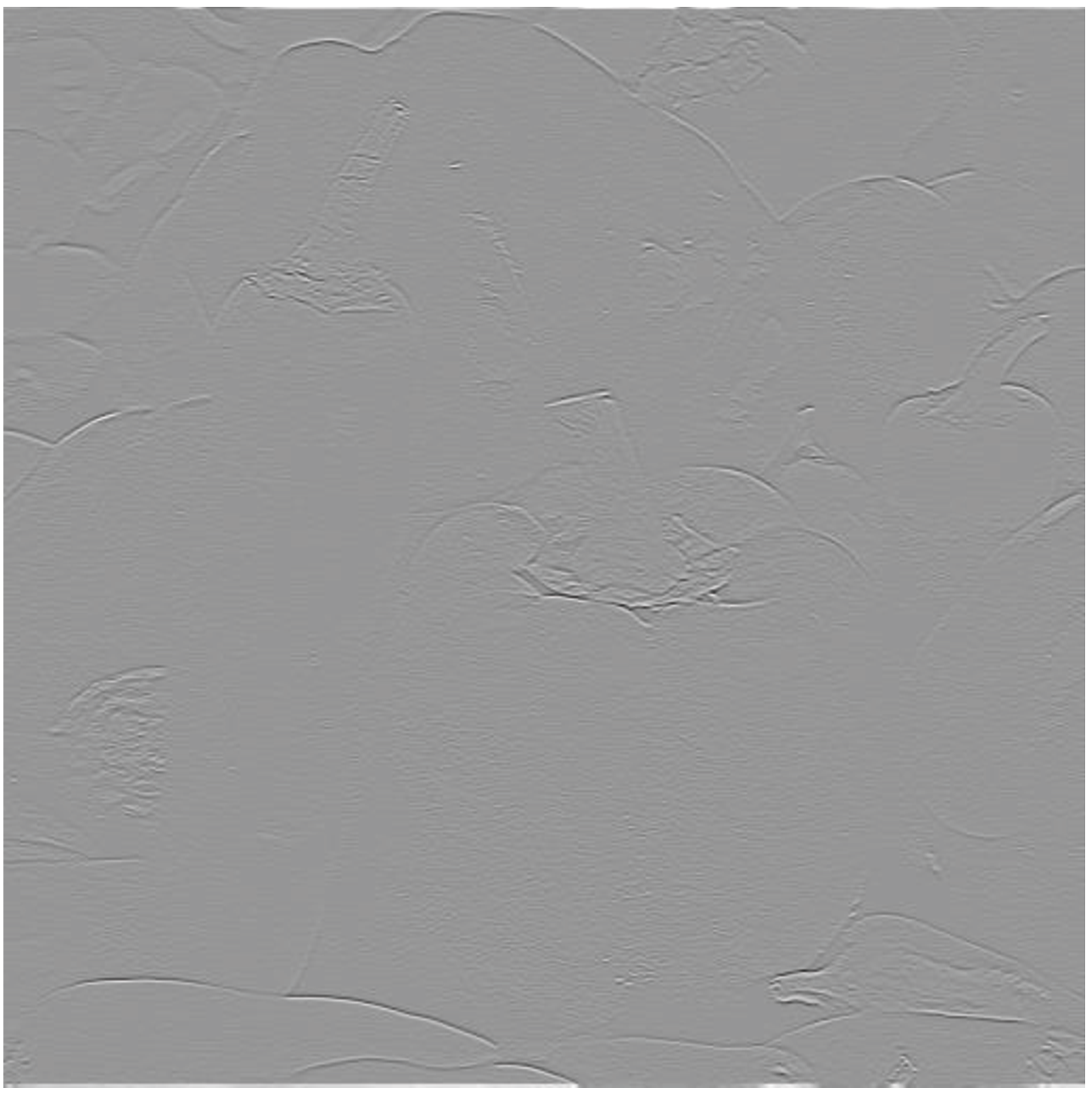}
\centerline{ \footnotesize (h)}
\end{minipage}
\begin{minipage}[t]{0.19\linewidth}
\centering
\includegraphics[width=1\textwidth]{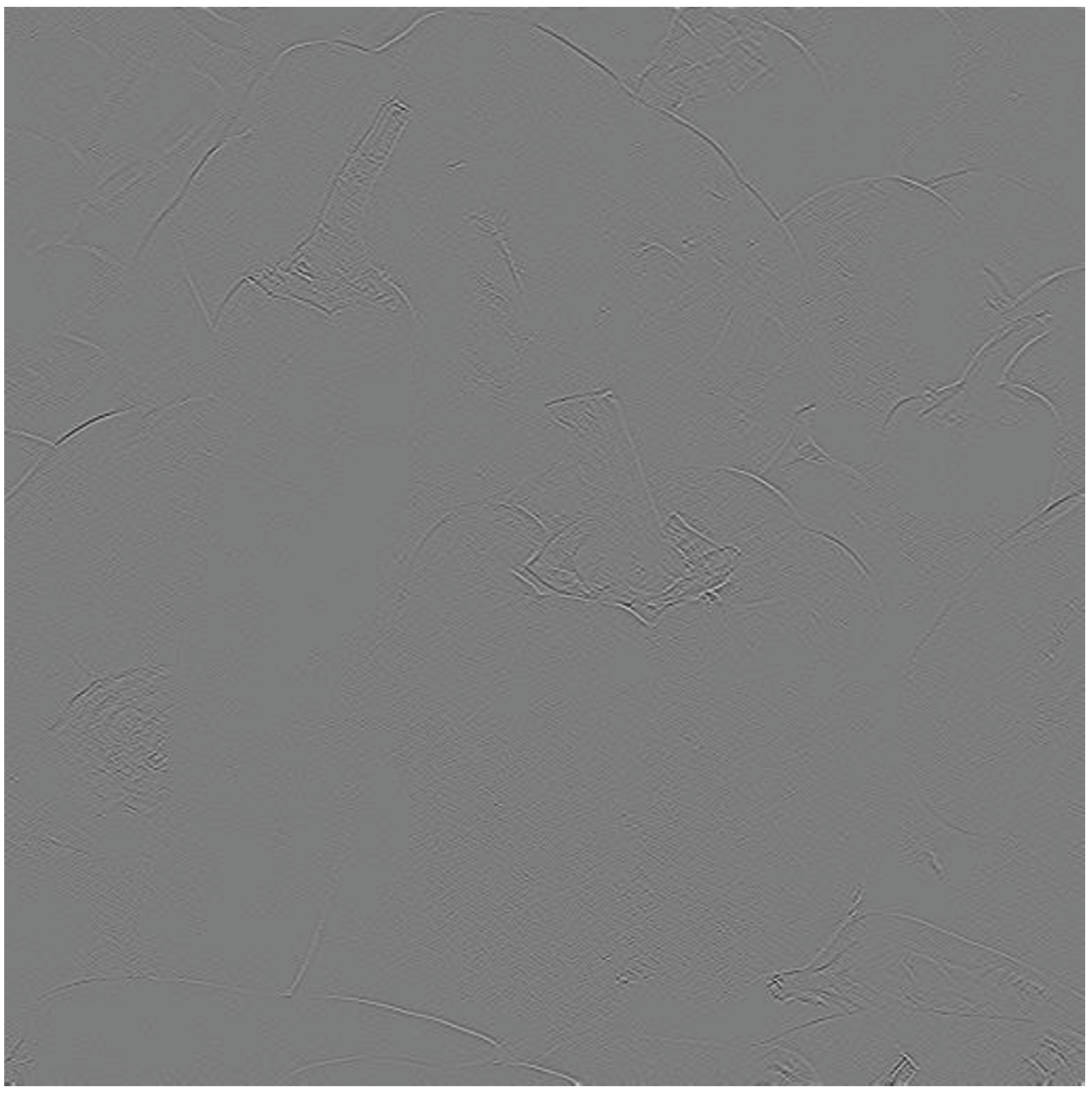}
\centerline{ \footnotesize (i)}
\end{minipage}
\begin{minipage}[t]{0.19\linewidth}
\centering
\includegraphics[width=1\textwidth]{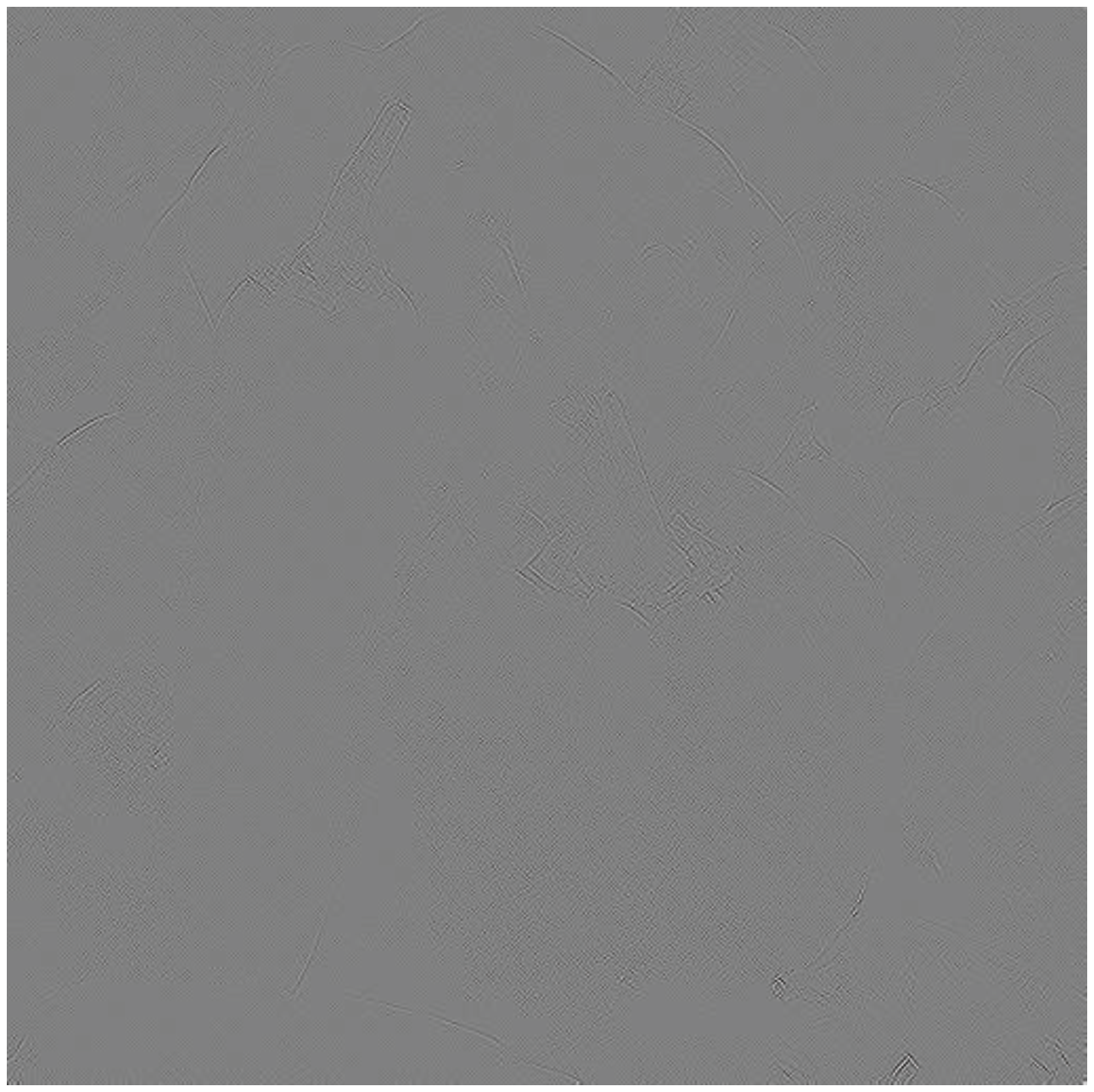}
\centerline{ \footnotesize (j)}
\end{minipage}
\caption{Illustration for feature extraction. (a) Original image. (b)--(j) Wavelet coefficients.}
\end{figure}

The 1-level tight wavelet frame decomposition gives redundant representations of an original image, thus forming its feature set. Specifically, each image pixel contains nine underlying attributes (channels), i.e., one low frequency and eight high ones. Formally speaking, for an image with $K$ pixels, the size of its feature set $\bm X$ is $9\times K$. Therefore, for image segmentation, the dimensionality of data for clustering is expanded when comparing with the direct use of image pixels.

\subsection{$\ell_{0}$ regularization-based FCM}
To achieve better segmentation effects, it is necessary to take the ideal value of an observed image as data for clustering, which means that the residual between them can be considered into clustering. As a result, we can augment the FCM algorithm by introducing a sparse regularization term on the residual into its objective function.

Before presenting the proposed algorithm, we respectively reformulate feature sets associated with $\hat{g}$, $g'$ and $\eta '$ in \eqref{GrindEQ__4_} as
\[{\bm X}={\rm {\mathcal W}}\hat{g}=\{ {\bm x}_{1} ,{\bm x}_{2} ,\cdots ,{\bm x}_{K} \} ,\]
\[\widetilde{{\bm X}}={\rm {\mathcal W}}g'=\{ \tilde{{\bm x}}_{1} ,\tilde{{\bm x}}_{2} ,\cdots ,\tilde{{\bm x}}_{K} \} ,\]
and
\[{\bm R}={\rm {\mathcal W}}\eta '=\{ {\bm r}_{1} ,{\bm r}_{2} ,\cdots ,{\bm r}_{K} \} .\]
Here, we have
\begin{equation} \label{GrindEQ__6_}
\widetilde{{\bm X}}={\bm X}-{\bm R}.
\end{equation}
In addition, $\bm{R}$ can also be rewritten as $\{\bm{R}_{l}: l=1,2,\cdots,L\}$, which means that each component $\bm{R}_{l}$ contains $L$ channels. In this work, $L=9$. Then we introduce an $\ell_{p} $ regularization term on ${\bm R}$ into the objective function of FCM. In the sequel, the modified objective function is expressed as
\begin{small}
\begin{equation} \label{GrindEQ__7_}
J({\bm U},{\bm V},{\bm R})=\sum_{i=1}^{c}\sum _{j=1}^{K}u_{ij}^{m} \|{\bm x}_{j} -{\bm r}_{j} -{\bm v}_{i} \|^{2}   +\sum_{l=1}^{L}\beta_{l} \|{\bm R}_{l} \|_{\ell_{p}}^{p}  ,
\end{equation}
\end{small}where $p\ge 0,$ ${\bm \beta }=\left\{\beta_{l}: l=1,2,\cdots,L\right\}$ is a parameter set that controls the impact of the $\ell_{p} $ regularization term on FCM, and
\begin{equation} \label{GrindEQ__8_}
\|{\bm R}_{l} \|_{\ell _{p} }^{p} =\left\{\begin{array}{l} {\sum\limits_{j=1}^{K}|r_{jl} |^{p}  ,{\kern 1pt} {\kern 1pt} {\kern 1pt} {\kern 1pt} {\kern 1pt} {\kern 1pt} p>0} \\ {\sum\limits_{j=1}^{K}|r_{jl} |_{0}  ,{\kern 1pt} {\kern 1pt} {\kern 1pt} {\kern 1pt} {\kern 1pt} {\kern 1pt} {\kern 1pt} p=0} \end{array}\right.
\end{equation}
with
\[|r_{jl} |_{0} =\left\{\begin{array}{l} {1,{\kern 1pt} {\kern 1pt} {\kern 1pt} {\kern 1pt} {\kern 1pt} {\kern 1pt} r_{jl} \ne 0} \\ {0,{\kern 1pt} {\kern 1pt} {\kern 1pt} {\kern 1pt} {\kern 1pt} r_{jl} =0} \end{array}\right. .\]
Here, $\|\cdot \|_{\ell_{p} } $ is the $\ell_{p} $ vector norm. Especially, $\|\cdot \|_{\ell _{0} } $ denotes the $\ell _{0} $ vector norm, and $\|{\bm R}_{l}\|_{\ell_{0}} $ represents the number of nonzero entries in ${\bm R}_{l} $. When $p\in [0,1)$, the minimization of \eqref{GrindEQ__7_} is a nonconvex problem. As to $p\in \left\{{\rm 4/3,3/2,2}\right\}$, the closed-form solution to minimizing \eqref{GrindEQ__7_} are derived in \cite{Chaux2007} and \cite{Combettes2005}. In particular, for $p=1$, the closed-form solution is given in \cite{Zhang2019} by using a general soft-thresholding operator.

Generally speaking, a large proportion of image data has a small or zero number of outliers, noise or intensity inhomogeneity. Therefore, ${\bm R}$ tends to be very sparse. To take the sparsity of ${\bm R}$ into consideration, in this work we focus on the case $p=0$. The main difference between this work and the above cases is in the form of the norms used for $\bm R$. Even though the use of $\ell_{0} $ norm gives rise to the difficulty for designing effective algorithms to solve the underlying optimization problems, it is beneficial to cope with a variety of cases \cite{Dong2013L0}. In addition, the use of spatial information is beneficial to improve FCM's robustness. Therefore, if the distance between an image pixel and its neighbors is small, there exists a large possibility that they belong to the same cluster. To further improve the segmentation performance, we introduce spatial information into the objective function of FCM.

As a result, by substituting \eqref{GrindEQ__8_} into \eqref{GrindEQ__7_} and considering spatial information, the modified objective function can be defined as:
\begin{equation} \label{GrindEQ__9_}
\begin{array}{l} {J({\bm U},{\bm V},{\bm R})=\sum\limits_{i=1}^{c}\sum\limits_{j=1}^{K}u_{ij}^{m} \left(\sum\limits_{n\in {\rm {\mathcal N}}_{j} }\frac{\|{\bm x}_{n} -{\bm r}_{n} -{\bm v}_{i} \|^{2} }{1+d_{nj}}  \right)} \\ \quad\quad\quad\quad\quad\quad\,\, {+\sum\limits_{l=1}^{L}\beta_{l} \sum\limits_{j=1}^{K}\sum\limits_{n\in {\rm {\mathcal N}}_{j} }\frac{|r_{nl} |_{0} }{1+d_{nj} }} \end{array},
\end{equation}where a pixel is sometimes loosely represented by its corresponding index while this is not ambiguous. Thus, ${\rm {\mathcal N}}_{j} $ stands for a local window centralized in $j$, $n$ is a neighbor pixel of $j$, and $d_{nj} $ represents the Euclidean distance between $n$ and $j$.

In the sequel, the Lagrangian multiplier method is applied to minimize \eqref{GrindEQ__9_}. The augmented Lagrangian function is:
\begin{equation*}
\begin{array}{l} {{\rm {\mathcal L}}_{\Lambda } ({\bm U},{\bm V},{\bm R})=\sum\limits_{i=1}^{c}\sum\limits_{j=1}^{K}u_{ij}^{m} \left(\sum\limits_{n\in {\it {\mathcal N}}_{j} }\frac{\|{\bm x}_{n} -{\bm r}_{n} -{\bm v}_{i} \|^{2} }{1+d_{nj} }  \right)} \\ \quad\quad\quad\quad { +\sum\limits_{l=1}^{L}\beta _{l} \sum\limits_{j=1}^{K}\sum\limits_{n\in {\it {\mathcal N}}_{j} }\frac{|r_{nl} |_{0} }{1+d_{nj} }+\sum\limits_{j=1}^{K}\lambda_{j} \left(\sum\limits_{i=1}^{c}u_{ij}-1\right)} \end{array},
\end{equation*}
where $\Lambda =\{ \lambda _{j} :j=1,2,\cdots ,K\} $ stands for a set of Lagrangian multipliers. The solution to the minimization of \eqref{GrindEQ__9_} can be produced in an iterative manner by handling the following three subproblems:
\begin{equation} \label{GrindEQ__10_}
\left\{\begin{array}{l} {{\bm U}^{(t+1)} =\arg {\mathop{\min}\limits_{{\bm U}}} \,\mathcal{L}_{\Lambda } ({\bm U},{\bm V}^{(t)} ,{\bm R}^{(t)} )} \\ {{\bm V}^{(t+1)} =\arg {\mathop{\min }\limits_{{\bm V}}} \, {\rm {\mathcal L}}_{\Lambda } ({\bm U}^{(t+1)} ,{\bm V},{\bm R}^{(t)} )} \\ {{\bm R}^{(t+1)} =\arg {\mathop{\min }\limits_{{\bm R}}}\, {\rm {\mathcal L}}_{\Lambda } ({\bm U}^{(t+1)} ,{\bm V}^{(t+1)} ,{\bm R})} \end{array}\right. .
\end{equation}

Each of the subproblem of \eqref{GrindEQ__10_} has a closed-form solution. We adopt an alternative optimization scheme similar to that used in the FCM algorithm to conduct the optimization of the partition matrix $\bm U$ and prototypes $\bm V$. The iterative updates of $\bm U$ and $\bm V$ are easily given as:
\begin{equation} \label{GrindEQ__11_}
u_{ij}^{(t+1)} =\frac{\left(\sum\limits_{n\in {\rm {\mathcal N}}_{j} }\frac{\|{\bm x}_{n} -{\bm r}_{n}^{(t)} -{\bm v}_{i}^{(t)} \|^{2} }{1+d_{nj}}  \right)^{-\frac{1}{m-1} } }{\sum\limits_{q=1}^{c}\left(\sum\limits_{n\in {\rm {\mathcal N}}_{j} }\frac{\|{\bm x}_{n} -{\bm r}_{n}^{(t)} -{\bm v}_{q}^{(t)} \|^{2} }{1+d_{nj} }  \right)^{-\frac{1}{m-1} } } ,
\end{equation}
\begin{equation} \label{GrindEQ__12_}
{\bm v}_{i}^{(t+1)} =\frac{\sum\limits_{j=1}^{K}\left(\left(u_{ij}^{(t+1)} \right)^{m} \sum\limits_{n\in {\rm {\mathcal N}}_{j} }\frac{{\bm x}_{n} -{\bm r}_{n}^{(t)} }{1+d_{nj} }  \right) }{\sum\limits_{j=1}^{K}\left(\left(u_{ij}^{(t+1)} \right)^{m} \sum\limits_{n\in {\rm {\mathcal N}}_{j} }\frac{1}{1+d_{nj} }  \right) } .
\end{equation}

When optimizing $\bm R$, it is obvious that both ${\bm r}_{j} $ and ${\bm r}_{n} $ are in \eqref{GrindEQ__9_}. Since ${\bm r}_{j} $ is not independent from ${\bm r}_{n} $, ${\bm r}_{n} $ cannot be treated as a constant vector. If $n$ is one of neighbors of $j$, $j$ is also one of neighbors of $n$ symmetrically. In the sequel, $n\in {\rm {\mathcal N}}_{j} $ is equivalent to $j\in {\rm {\mathcal N}}_{n} $. Then we have
\begin{small}
\begin{equation} \label{GrindEQ__13_}
\sum\limits_{j=1}^{K}u_{ij}^{m} \left(f({\bm r}_{j} )+\sum\limits_{\begin{array}{l} {n\in {\rm {\mathcal N}}_{j} } \\ {{\kern 1pt} {\kern 1pt} n\ne j} \end{array}}f({\bm r}_{n} ) \right) =\sum\limits_{j=1}^{K}\sum\limits_{n\in {\rm {\mathcal N}}_{j} }u_{in}^{m} (f({\bm r}_{j} ))  ,
\end{equation}
\end{small}where $f$ stands for a function in terms of ${\bm r}_{j} $ or ${\bm r}_{n} $. According to \eqref{GrindEQ__13_}, \eqref{GrindEQ__9_} is rewritten as:
\begin{equation} \label{GrindEQ__14_}
\begin{array}{l} {J({\bm U},{\bm V},{\bm R})=\sum\limits_{i=1}^{c}\sum\limits_{j=1}^{K}\sum\limits_{n\in {\rm {\mathcal N}}_{j} }\frac{u_{in}^{m} \|{\bm x}_{j} -{\bm r}_{j} -{\bm v}_{i} \|^{2} }{1+d_{nj} }} \\ \quad\quad\quad\quad\quad\quad\,\, {+\sum\limits_{l=1}^{L}\beta_{l} \sum\limits_{j=1}^{K}\sum\limits_{n\in {\rm {\mathcal N}}_{j} }\frac{|r_{jl} |_{0} }{1+d_{nj} }} \end{array}.
\end{equation}

Based on \eqref{GrindEQ__14_}, once $\bm U$ and $\bm V$ are updated, the third subproblem of \eqref{GrindEQ__10_} is separable and the optimization of $\bm R$ can be decomposed into $K\times L$ subproblems as follows:
\begin{footnotesize}
\begin{equation} \label{GrindEQ__15_}
\begin{array}{l} {r_{jl}^{(t+1)} =\arg {\mathop{\min }\limits_{r_{jl} }} \sum\limits_{i=1}^{c}\left(\sum\limits_{n\in {\rm {\mathcal N}}_{j} }\frac{\left(u_{in}^{(t+1)} \right)^{m} \|x_{jl} -r_{jl} -v_{il}^{(t+1)} \|^{2} }{1+d_{nj} }  \right)} \\ \quad\quad\quad\quad\quad\quad\quad\,\, {+\sum\limits_{n\in {\rm {\mathcal N}}_{j} }\frac{\beta_{l} |r_{jl} |_{0} }{1+d_{nj} }} \end{array}.
\end{equation}
\end{footnotesize}
We can employ a well-known hard-thresholding operator to solve \eqref{GrindEQ__15_}. The iterative formula of residuals is expressed as
\begin{footnotesize}
\begin{equation} \label{GrindEQ__16_}
r_{jl}^{(t+1)} =\frac{{\rm {\mathcal H}}_{\sum\limits_{n\in {\rm {\mathcal N}}_{j} }\frac{\beta_{l} }{1+d_{nj} }  } \left(\sum\limits_{i=1}^{c}\sum\limits_{n\in {\rm {\mathcal N}}_{j} }\frac{\left(u_{in}^{(t{\rm +}1)} \right)^{m} \left(x_{jl} -v_{il}^{(t+1)} \right)}{1+d_{nj} }   \right)}{\sum\limits_{i=1}^{c}\sum\limits_{n\in {\rm {\mathcal N}}_{j} }\frac{\left(u_{in}^{(t{\rm +}1)} \right)^{m} }{1+d_{nj} }   } ,
\end{equation}
\end{footnotesize}where ${\rm {\mathcal H}}$ is a hard-thresholding operator defined as:
\[{\rm {\mathcal H}}_{\sigma } (\xi )=\left\{\begin{array}{l} {\xi ,{\kern 1pt} {\kern 1pt} {\kern 1pt} {\kern 1pt} {\kern 1pt} {\kern 1pt} \xi \ge \sqrt{\sigma } } \\ {0,{\kern 1pt} {\kern 1pt} {\kern 1pt} {\kern 1pt} {\kern 1pt} {\kern 1pt} {\kern 1pt} \xi <\sqrt{\sigma } } \end{array}\right. .\]

\subsection{Label Smoothing via MR}

To further enhance the segmentation effects of the $\ell_{0}$ regularization-based FCM algorithm, we also use MR to smoothen the obtained labels of pixels. We define the label of the $j$-th pixel as $\phi_{j}=label(u_{ij})$:
\[u_{ij} =\arg\max \left\{u_{1j} ,u_{2j} ,\cdots ,u_{cj} \right\},\]
where $label$ denotes the location of maximum $u_{ij} $, i.e., $\phi_{j} =i$. This means that the $j$-th pixel belongs to the $i$-th cluster. Thus, we can define the label set of $K$ pixels as
\begin{equation} \label{GrindEQ__17_}
\Phi =\{\phi _{j}:j=1,2,\cdots ,K\}.
\end{equation}

In the sequel, $\Phi $ is arranged into a matrix of the same size as $\hat{g}$, thus generating a label image $\Phi _{im} $. We employ MR to smoothen the obtained label image $\Phi_{im} $ so as to generate a smoothed label image that is formulated as
\begin{equation} \label{GrindEQ__18_}
\overline{\Phi }_{im} ={\rm {\mathcal R}}^{C} (\Phi _{im} ).
\end{equation}

\subsection{Reconstruction of Segmented Image}

Based on the smoothed label image $\overline{\Phi }_{im} $ and the obtained prototypes $\bm V$, the segmented feature set $\widehat{{\bm X}}$ is obtained. Then we use the wavelet frame reconstruction operator ${\rm {\mathcal W}}^{T} $ to reconstruct a segmented image $g''$:
\begin{equation} \label{GrindEQ__19_}
g''={\rm {\mathcal W}}^{T} (\widehat{{\bm X}}).
\end{equation}
The proposed method is realized in \textbf{Algorithm 1}.

\begin{algorithm}[htbp]
\caption{$\ell_{0}$ regularization-based FCM algorithm incorporating MR and wavelet frames (LRFCM)}
\begin{algorithmic}[1]
\REQUIRE Observed image $g$, fuzzification coefficient $m$, number of clusters $c$, and threshold $\varepsilon$.
\ENSURE Segmented image $g''$.
\STATE Calculate the filtered image $\bar{g}$ via \eqref{GrindEQ__2_}
\STATE Calculate the weighted sum image $\hat{g}$ via \eqref{GrindEQ__3_}
\STATE Generate the feature set $\bm X$ via \eqref{GrindEQ__5_}
\STATE Initialize randomly the prototypes ${\bm V}^{(0)} $
\STATE $t\leftarrow 0$
\REPEAT
\STATE Calculate the partition matrix ${\bm U}^{(t+1)} $ by using the residuals ${\bm R}^{(t)} $ and the prototypes ${\bm V}^{(t)} $ via \eqref{GrindEQ__11_}
\STATE Update the prototypes ${\bm V}^{(t+1)} $ by using the partition matrix ${\bm U}^{(t+1)} $ and the residuals ${\bm R}^{(t)} $ via \eqref{GrindEQ__12_}
\STATE Update the residuals ${\bm R}^{(t{\rm +}1)} $ by using the partition matrix ${\bm U}^{(t+1)} $ and the prototypes ${\bm V}^{(t+1)} $ via \eqref{GrindEQ__16_}
\STATE $t\leftarrow t+1$
\UNTIL {$ \|\bm U^{(t+1)}-\bm U^{(t)}\|< \varepsilon $}
\STATE \textbf{return} partition matrix $\bm U$, prototypes $\bm V$, and residuals $\bm R$\\
\STATE Generate the labels of image pixels via \eqref{GrindEQ__17_}
\STATE Smoothen the labels via \eqref{GrindEQ__18_}
\STATE Calculate the segmented image $g''$ via \eqref{GrindEQ__19_}
\end{algorithmic}
\end{algorithm}

\section{Experimental Studies}
\label{IV}
In this section, we proceed with numerical experiments to investigate the effectiveness and efficiency of the proposed algorithm (namely LRFCM). Numerical results reported for synthetic, medical, and color images are provided. In addition, we compare the proposed algorithm with the four classic algorithms in the literature, i.e., `FCM\_S1' \cite{Chen2004}, `FCM\_S2' \cite{Chen2004}, `FGFCM' \cite{Cai2007}, and `FLICM' \cite{Krinidis2010}, and five recently proposed algorithms including `KWFLICM' \cite{Gong2013}, `ARKFCM' [12], `FRFCM' \cite{Lei2018}, `WFCM' \cite{Wang2019}, and `DSFCM\_N' \cite{Zhang2019}. Finally, we conduct ablation studies and analyze the impact of each component in LRFCM.

\subsection{Parameter Settings}

Prior to accomplishing all numerical experiments, we report parameter settings of all algorithms. A local window of size $3\times 3$ is fairly chosen since spatial information is considered in all algorithms. We set the fuzzification coefficient $m=2,$ and threshold $\varepsilon =1\times 10^{-6} $ across all algorithms. Moreover, the proper number of clusters $c$ is assumed to be known, and how to decide $c$ is introduced in each experiment.

Except for usual parameters $m,$ $\varepsilon ,$ and $c,$ there are no more parameters in FLICM, KWFLICM and ARKFCM. Nevertheless, there exist different parameters in the remaining algorithms. For FCM\_S1 and FCM\_S2, $\alpha $ is uniformly set to 3.8, which aims to constrain the neighbor term. In FGFCM, the spatial scale factor $\lambda _{s} $ and gray-level scale factor $\lambda _{g} $ are respectively set to 3 and 5. For FRFCM, according to \cite{Lei2018}, we select the observed image as the mask image, and generate the marker image with the aid of a square structuring element of size $3\times 3.$ Moreover, a median filter of size $3\times 3$ is applied to the membership filtering. As to WFCM, we experimentally select $\mu \in [0.55,0.65]$, which is used to control the effect of spatial information. For DSFCM\_N, the parameter vector $\lambda $ is selected according to the standard deviation of each channel of image data.

For LRFCM, we apply MR to image pixel filtering and label smoothing based on the same setting as that in FRFCM. For fair comparisons, we set $\alpha $ to 3.8, which controls the impact of the filtered image. The level of tight wavelet frame transform is set to 1. Moreover, there only exists one parameter ${\bm \beta }$ in LRFCM. Since the main difference between LRFCM and DSFCM\_N is the norms used for residuals $\bm{R}$, the parameter setting for the sparse regularization term in DSFCM\_N could be referenced. ${\bm \beta }$ is associated with the standard deviation of each channel of image data since the standard deviation can reflect the noise level to some extent \cite{Zhang2019}. After massive experiments, ${\bm \beta }=\left\{\beta_{l}:l=1,2,\cdots ,L\right\}$ is recommended to be as:
$$
\beta _{l} =70\delta _{l},
$$
where $\delta_{l}$ is the standard deviation of the $l$-th channel of image data.

\subsection{Visual Comparisons and Analysis}
In the first experiments, we segment two synthetic images shown in Fig. 4(a) and Fig. 5(a). The first synthetic image with size $256 \times 256$ has four gray levels, i.e., 0, 85, 170, and 255. The number of clusters is set to 4. We impose the Gaussian noise of 30 intensity on the image. The visual comparison results are shown in Fig. 4.

\begin{figure}[htb]
\centering
\begin{minipage}[t]{0.23\linewidth}
\centering
\includegraphics[width=1\textwidth]{s2.eps}
\centerline{(a)}
\end{minipage}
\begin{minipage}[t]{0.23\linewidth}
\centering
\includegraphics[width=1\textwidth]{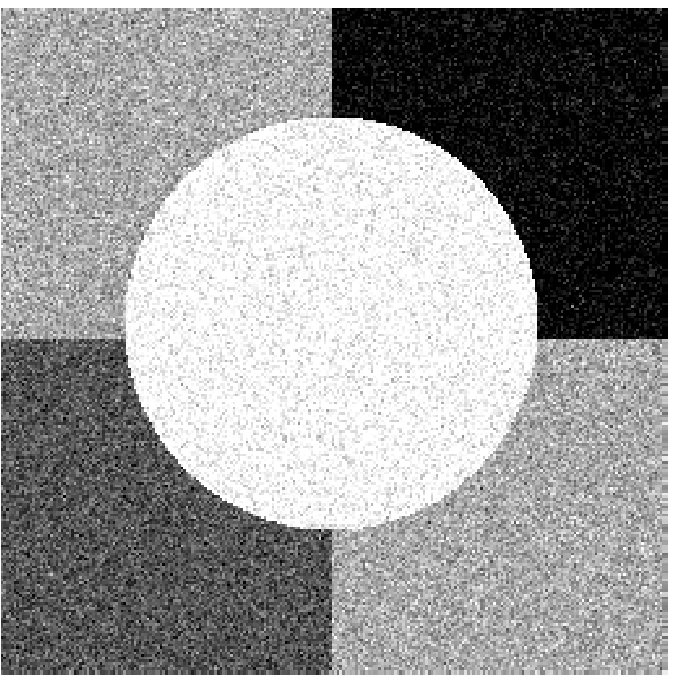}
\centerline{(b)}
\end{minipage}
\begin{minipage}[t]{0.23\linewidth}
\centering
\includegraphics[width=1\textwidth]{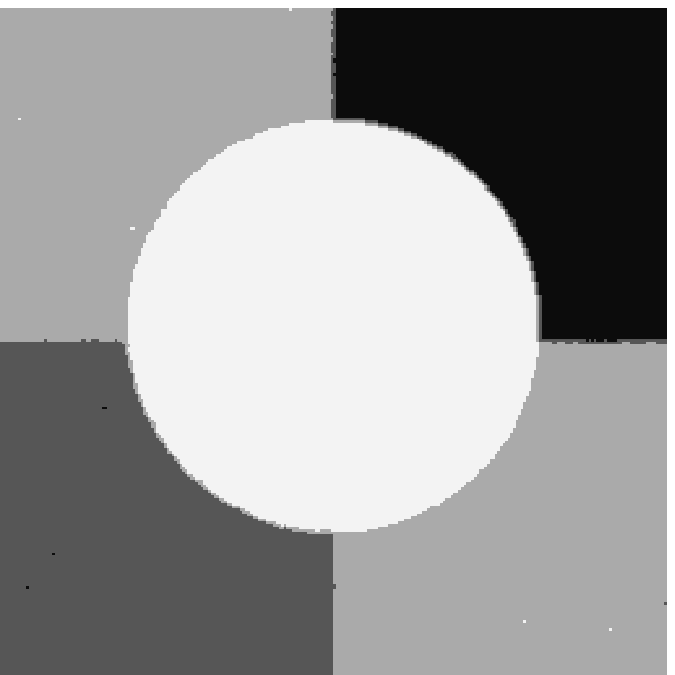}
\centerline{(c)}
\end{minipage}
\begin{minipage}[t]{0.23\linewidth}
\centering
\includegraphics[width=1\textwidth]{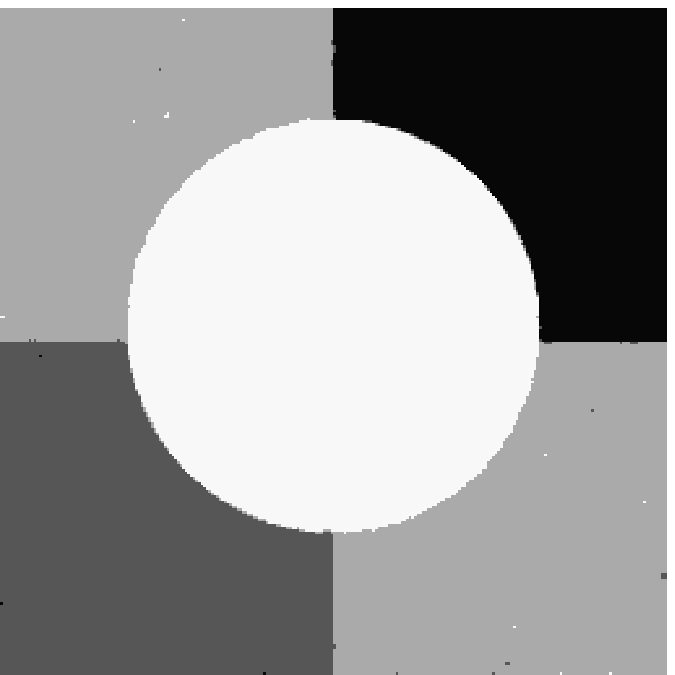}
\centerline{(d)}
\end{minipage}
\begin{minipage}[t]{0.23\linewidth}
\centering
\includegraphics[width=1\textwidth]{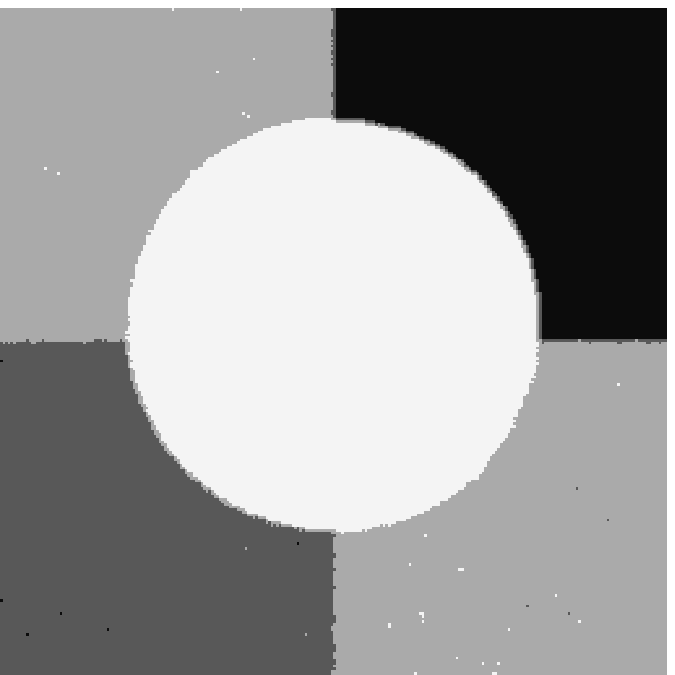}
\centerline{(e)}
\end{minipage}
\begin{minipage}[t]{0.23\linewidth}
\centering
\includegraphics[width=1\textwidth]{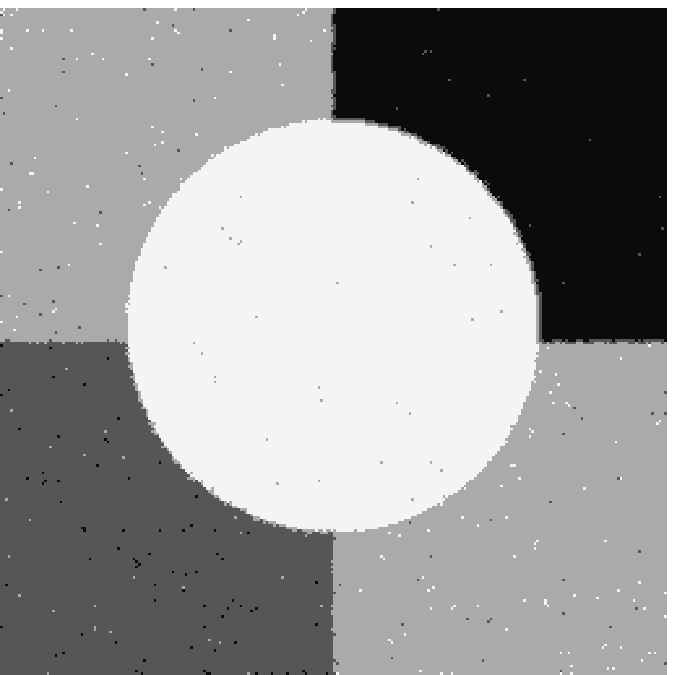}
\centerline{(f)}
\end{minipage}
\begin{minipage}[t]{0.23\linewidth}
\centering
\includegraphics[width=1\textwidth]{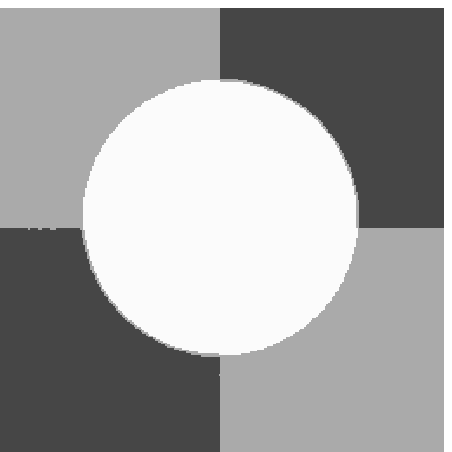}
\centerline{(g)}
\end{minipage}
\begin{minipage}[t]{0.23\linewidth}
\centering
\includegraphics[width=1\textwidth]{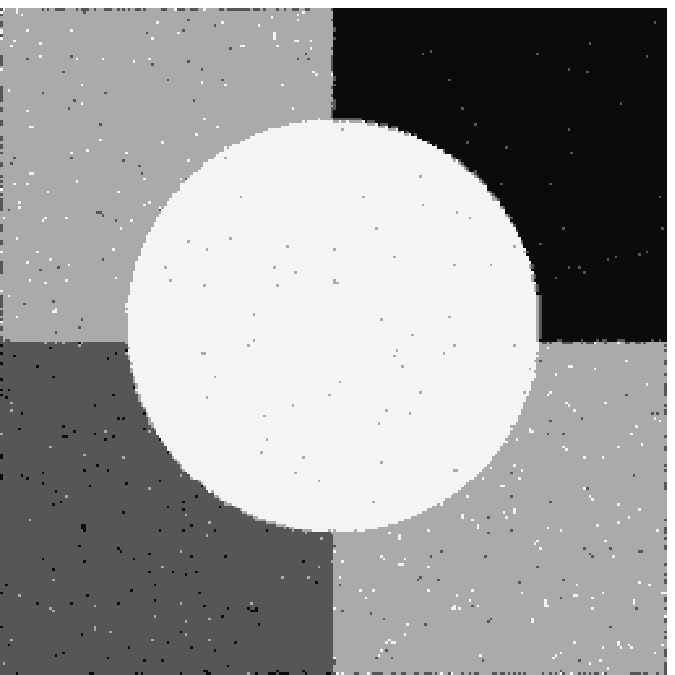}
\centerline{(h)}
\end{minipage}
\begin{minipage}[t]{0.23\linewidth}
\centering
\includegraphics[width=1\textwidth]{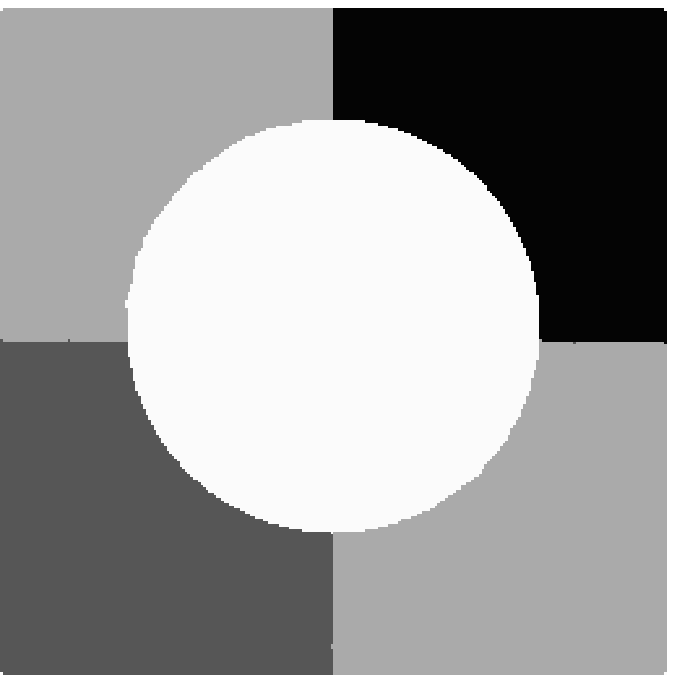}
\centerline{(i)}
\end{minipage}
\begin{minipage}[t]{0.23\linewidth}
\centering
\includegraphics[width=1\textwidth]{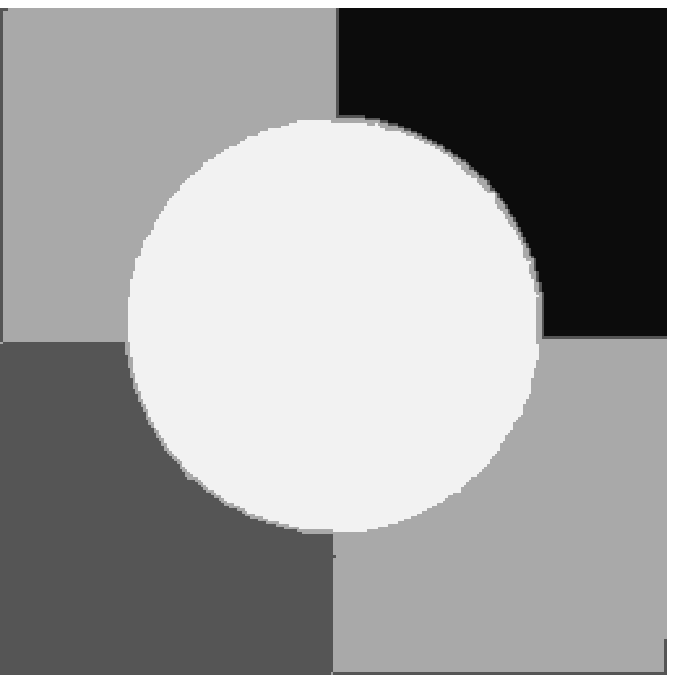}
\centerline{(j)}
\end{minipage}
\begin{minipage}[t]{0.23\linewidth}
\centering
\includegraphics[width=1\textwidth]{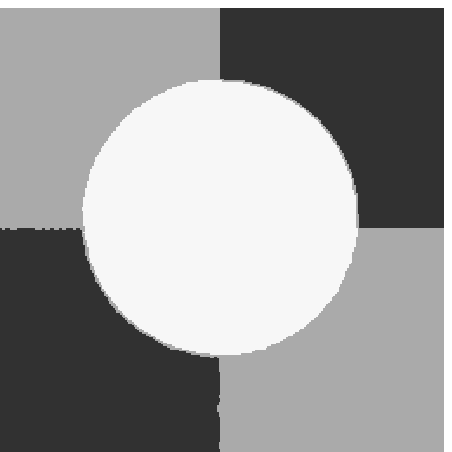}
\centerline{(k)}
\end{minipage}
\begin{minipage}[t]{0.23\linewidth}
\centering
\includegraphics[width=1\textwidth]{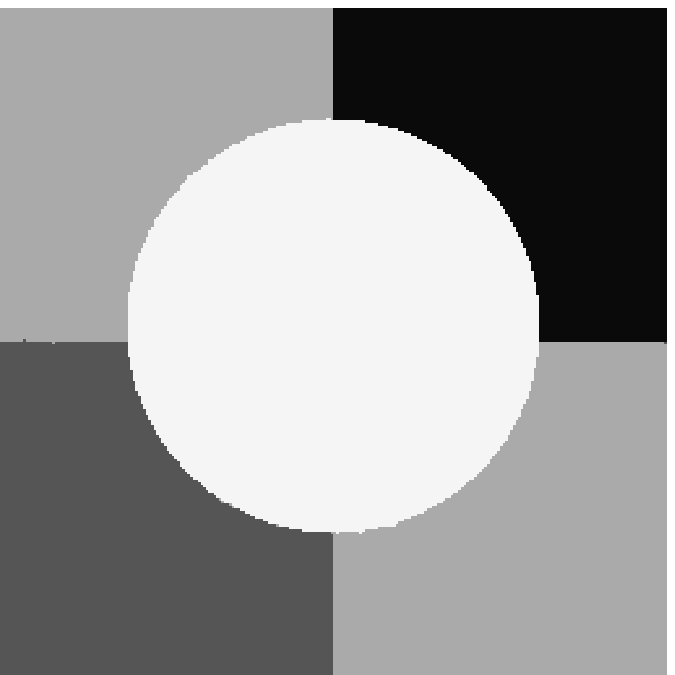}
\centerline{(l)}
\end{minipage}
\caption{Segmentation results for the first synthetic image. From (a) to (l): original image, noisy image and results of FCM\_S1, FCM\_S2, FGFCM, FLICM, KWFLICM, ARKFCM, FRFCM, WFCM, DSFCM\_N, and LRFCM.}
\end{figure}

As illustrated in Fig. 4, FLICM and ARKFCM have poor performance in suppressing Gaussian noise. Although FCM\_S1, FCM\_S2 and FGFCM can remove a large proportion of Gaussian noise, there is still a small amount of noise in their segmentation results. On the contrary, KWFLICM, FRFCM, WFCM and DSFCM\_N have a good capacity of noise suppression. However, they produce several topology changes such as merging and splitting. For example, there exist many unclear contours and some shadows attached to the edge of the circle in Fig. 4(g) and Figs. 4(i)--(k). Superior to its nine peers, LRFCM is robust to Gaussian noise and retains more image features.

The second synthetic image with size $308\times 242$ is without ground truth. The number of clusters is set to 2. To exhibit LRFCM's robustness to different types of noise, we impose impulse noise of 30\% density on the image. In particular, we consider only salt and pepper impulse noise since it is one of the most common types of impulse noise. The visual comparison results are drawn in Fig. 5.
\begin{figure}[htb]
\centering
\begin{minipage}[t]{0.23\linewidth}
\centering
\includegraphics[width=1\textwidth]{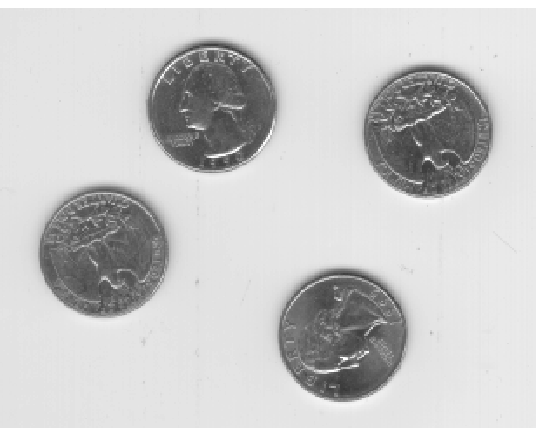}
\centerline{(a)}
\end{minipage}
\begin{minipage}[t]{0.23\linewidth}
\centering
\includegraphics[width=1\textwidth]{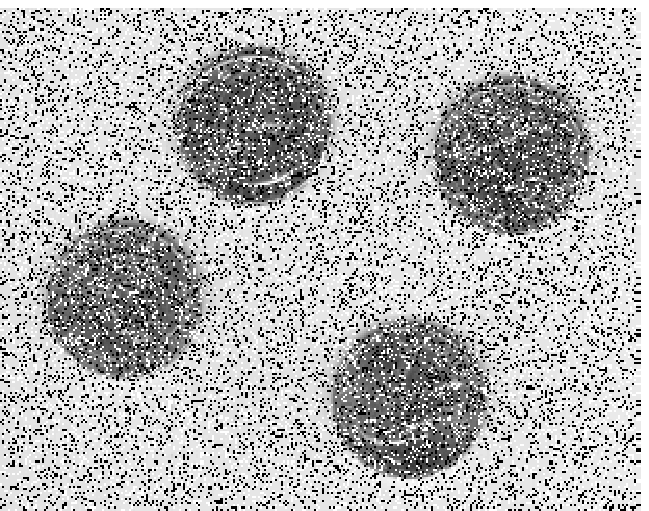}
\centerline{(b)}
\end{minipage}
\begin{minipage}[t]{0.23\linewidth}
\centering
\includegraphics[width=1\textwidth]{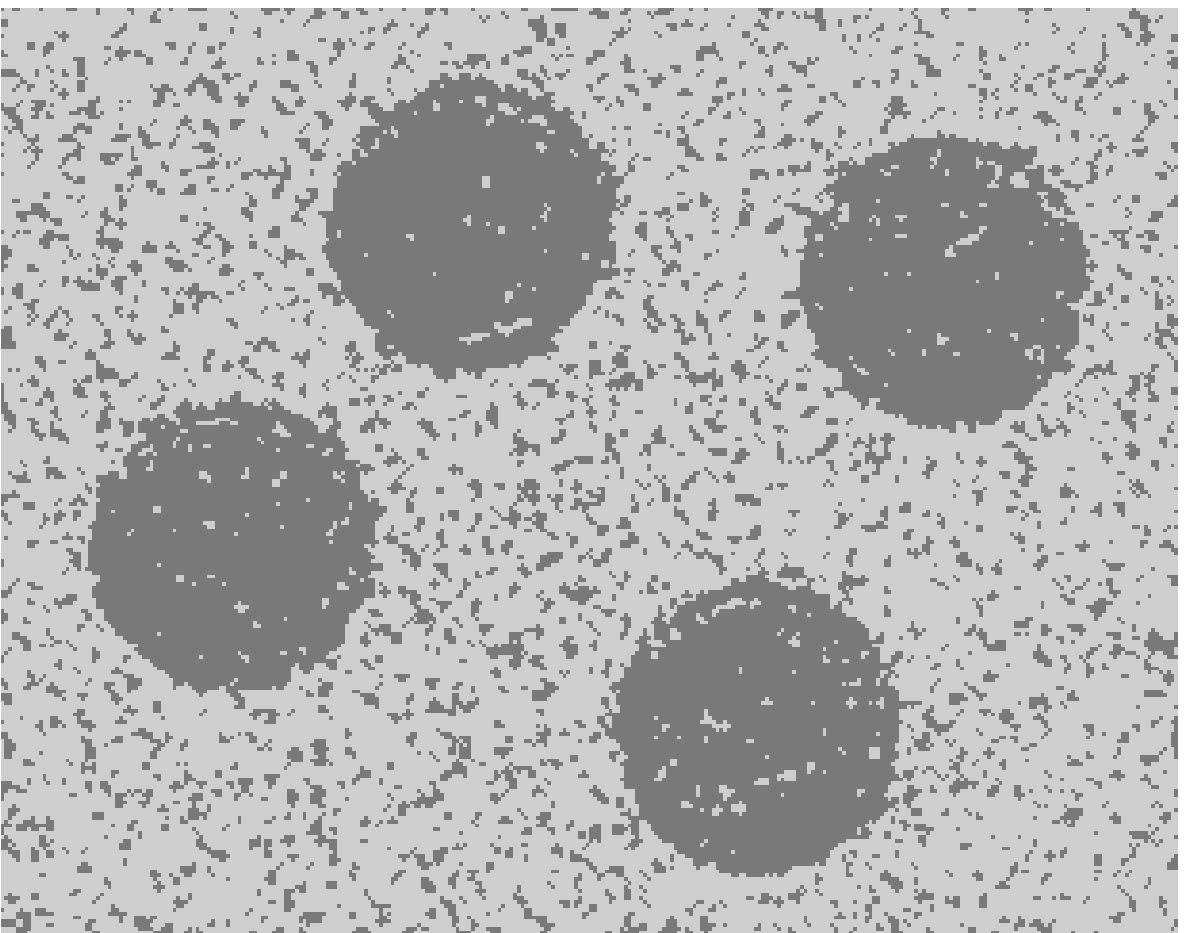}
\centerline{(c)}
\end{minipage}
\begin{minipage}[t]{0.23\linewidth}
\centering
\includegraphics[width=1\textwidth]{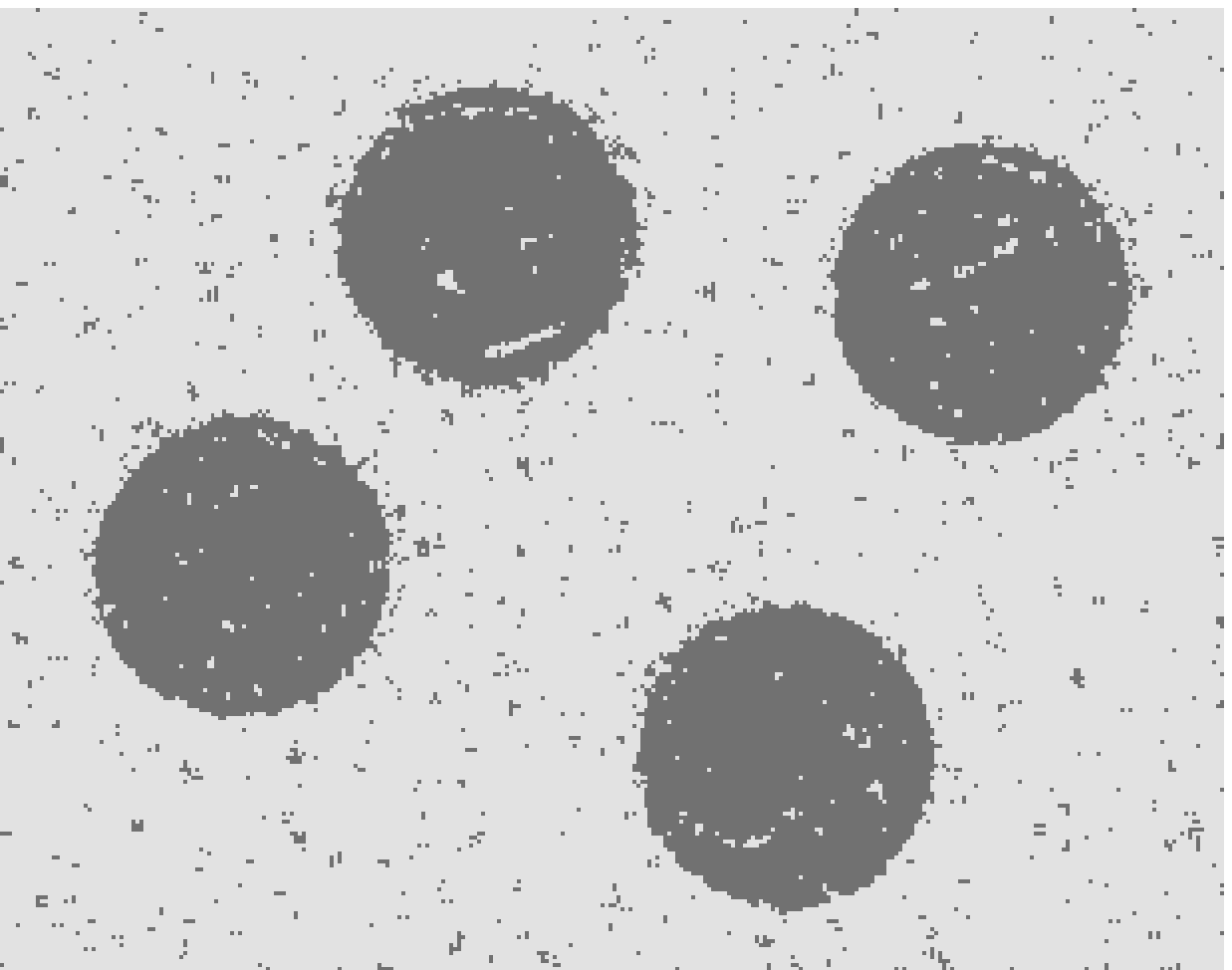}
\centerline{(d)}
\end{minipage}
\begin{minipage}[t]{0.23\linewidth}
\centering
\includegraphics[width=1\textwidth]{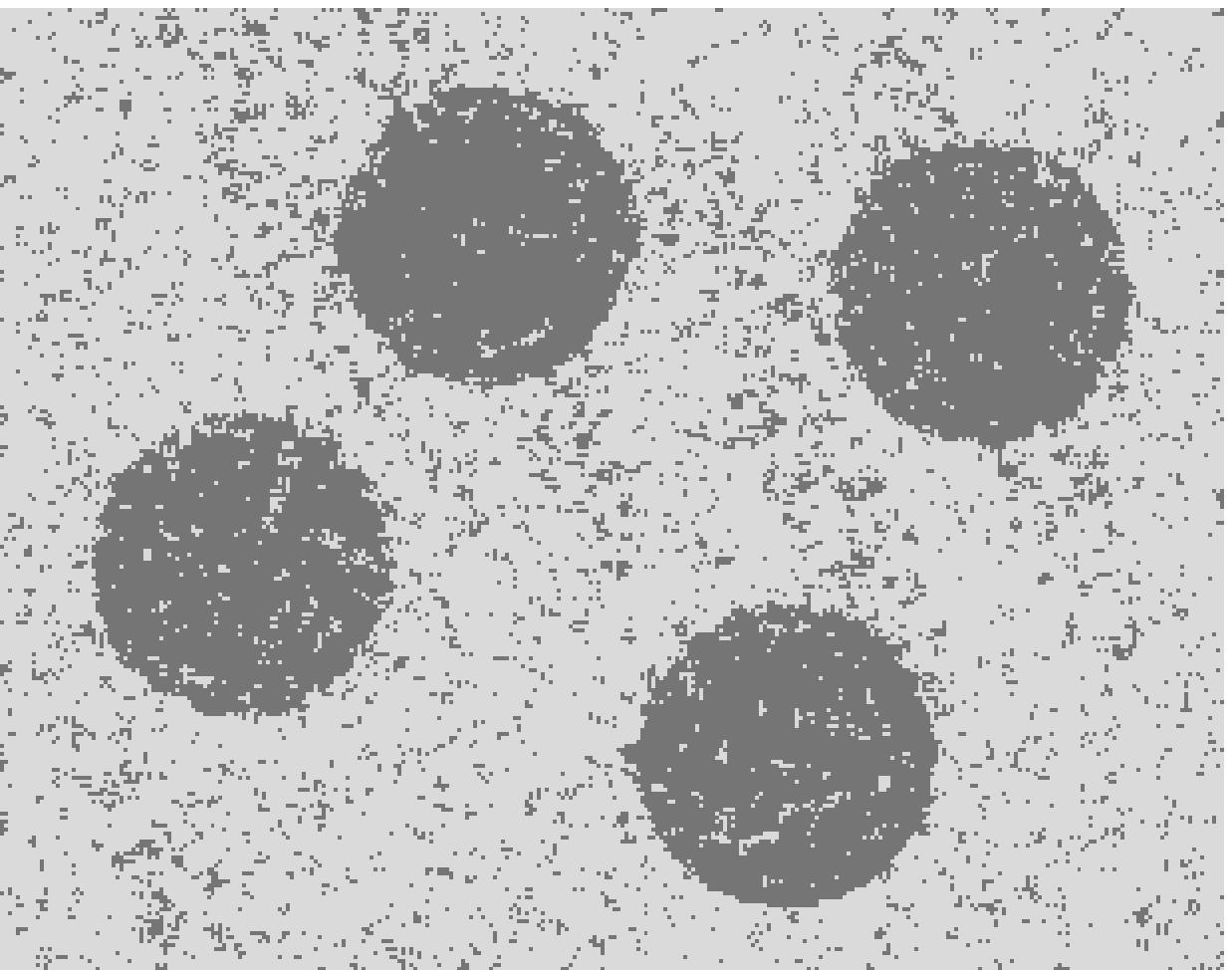}
\centerline{(e)}
\end{minipage}
\begin{minipage}[t]{0.23\linewidth}
\centering
\includegraphics[width=1\textwidth]{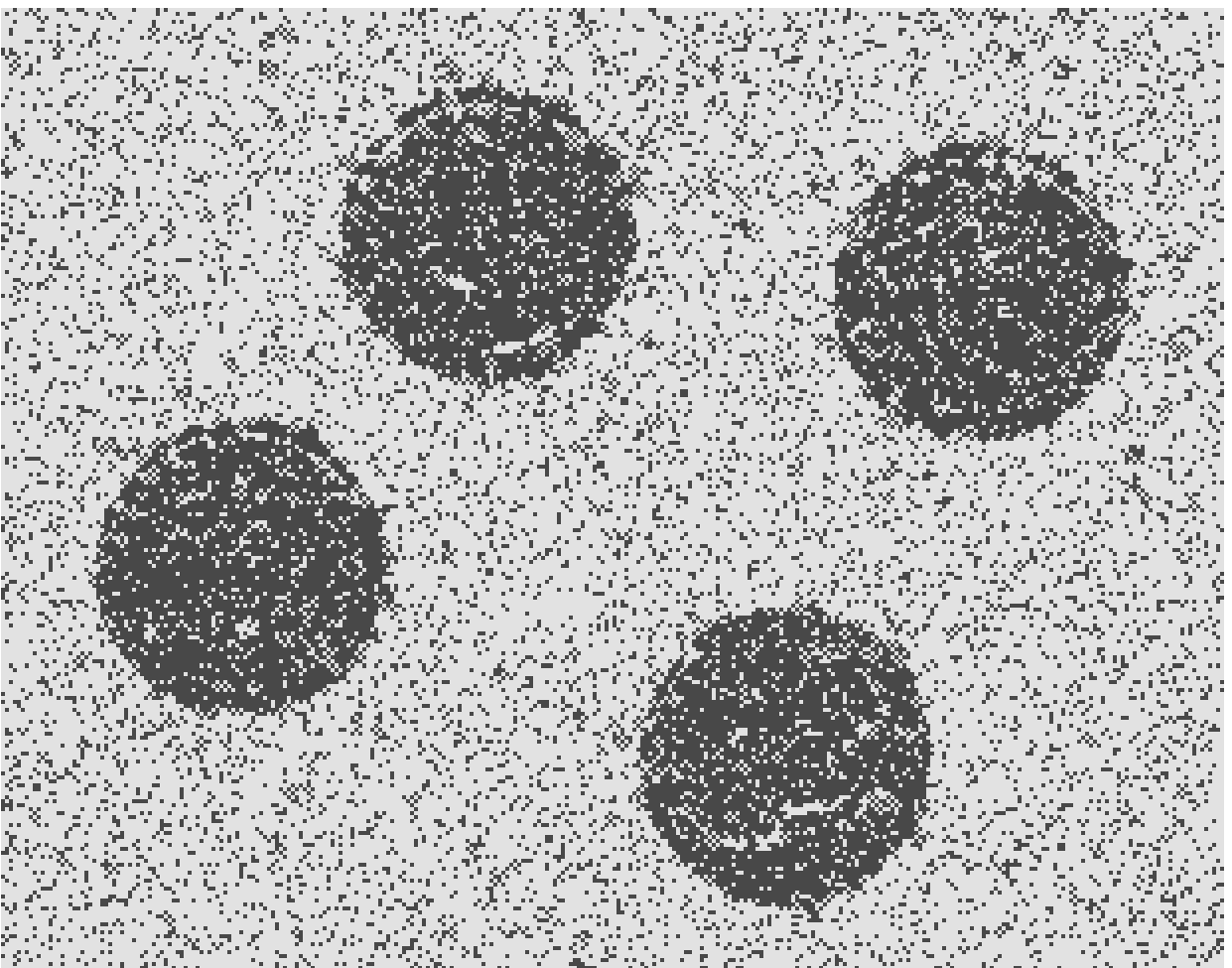}
\centerline{(f)}
\end{minipage}
\begin{minipage}[t]{0.23\linewidth}
\centering
\includegraphics[width=1\textwidth]{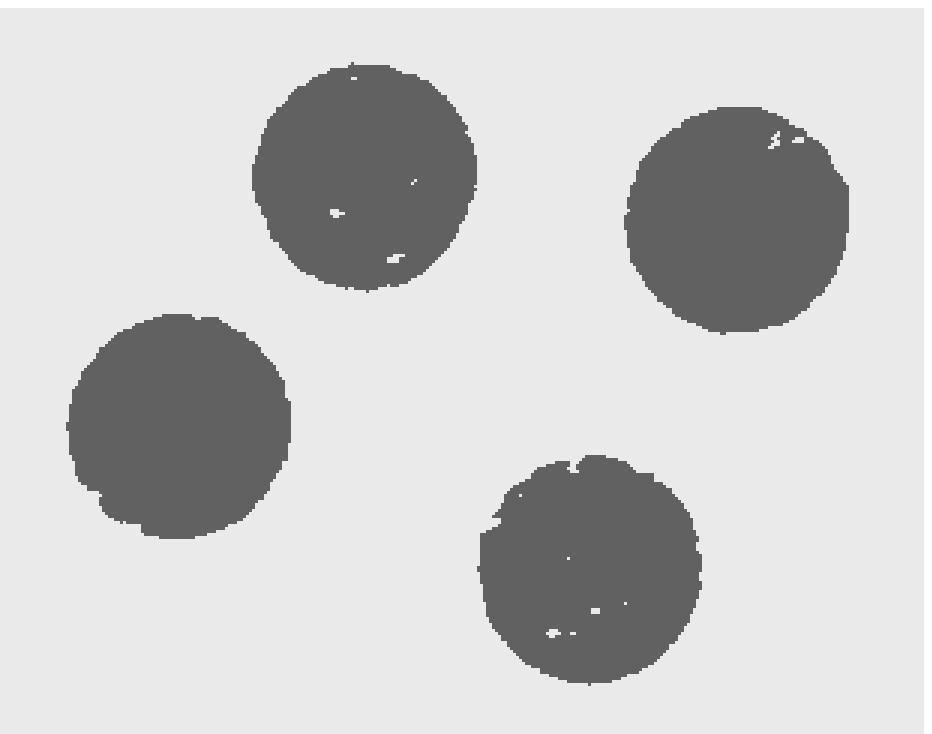}
\centerline{(g)}
\end{minipage}
\begin{minipage}[t]{0.23\linewidth}
\centering
\includegraphics[width=1\textwidth]{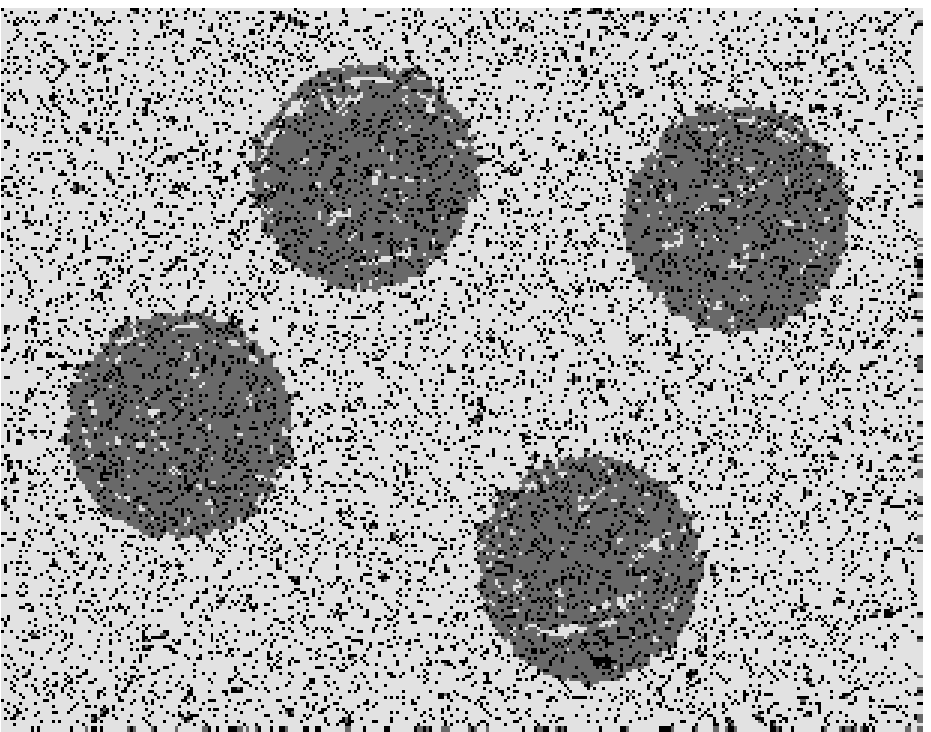}
\centerline{(h)}
\end{minipage}
\begin{minipage}[t]{0.23\linewidth}
\centering
\includegraphics[width=1\textwidth]{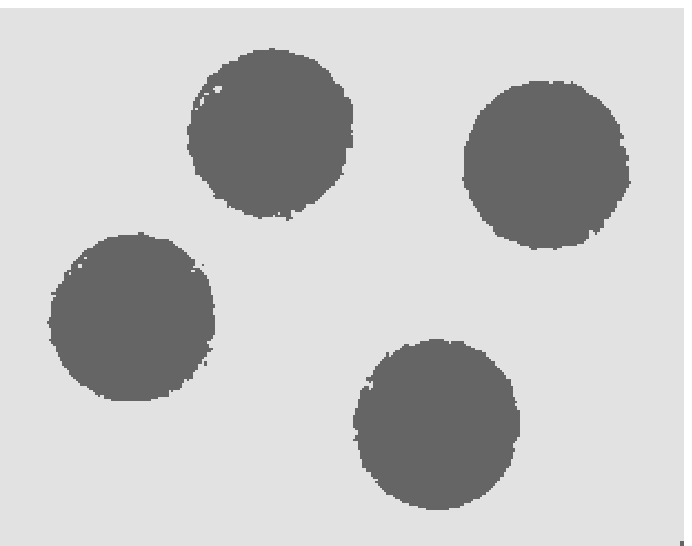}
\centerline{(i)}
\end{minipage}
\begin{minipage}[t]{0.23\linewidth}
\centering
\includegraphics[width=1\textwidth]{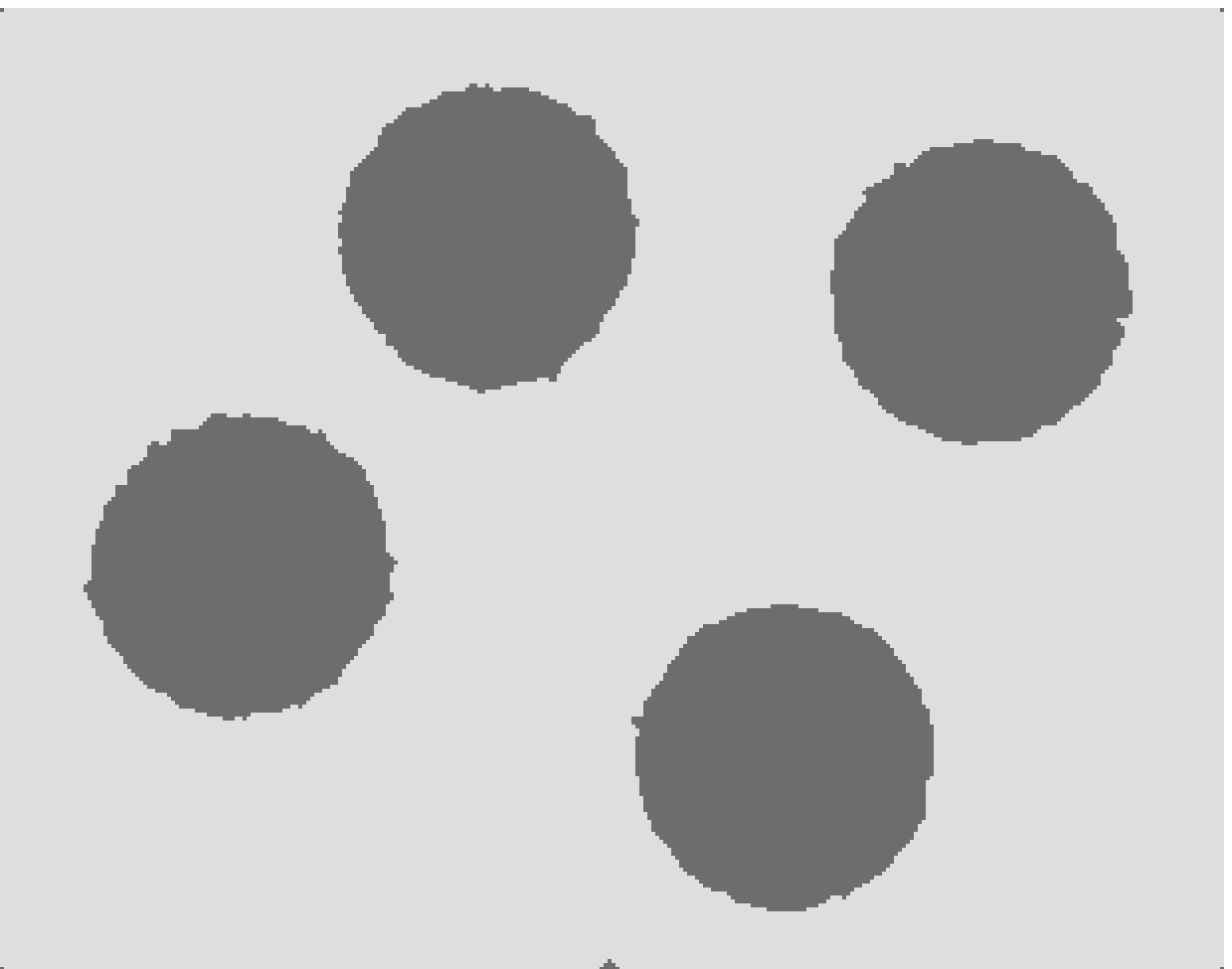}
\centerline{(j)}
\end{minipage}
\begin{minipage}[t]{0.23\linewidth}
\centering
\includegraphics[width=1\textwidth]{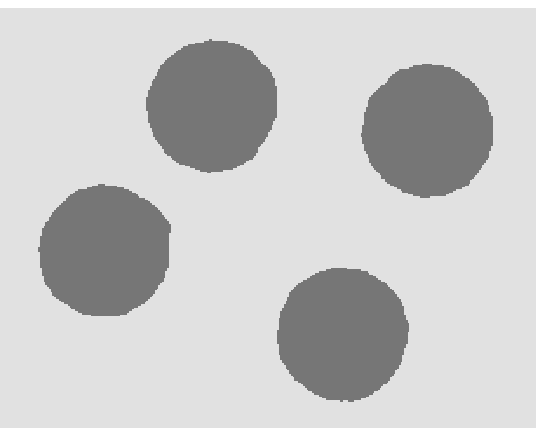}
\centerline{(k)}
\end{minipage}
\begin{minipage}[t]{0.23\linewidth}
\centering
\includegraphics[width=1\textwidth]{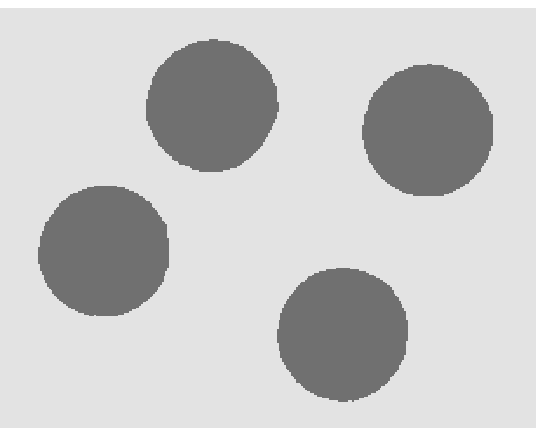}
\centerline{(l)}
\end{minipage}
\caption{Segmentation results for the second synthetic image. From (a) to (l): original image, noisy image and results of FCM\_S1, FCM\_S2, FGFCM, FLICM, KWFLICM, ARKFCM, FRFCM, WFCM, DSFCM\_N, and LRFCM.}
\end{figure}

As shown in Fig. 5, except WFCM and DSFCM\_N, other comparative algorithms cannot fully remove impulse noise. In particular, the segmentation results of FCM\_S1, FCM\_S2, FGFCM, FLICM and ARKFCM are far from being satisfactory. When focusing on the results of WFCM and DSFCM\_N, we find that there are some unsmooth edges. Nevertheless, DSFCM\_N performs better than WFCM. Compared with DSFCM\_N, LRFCM yields slightly better results since it achieves smoother edges.

In the second experiments, we segment two medical images coming from a simulated brain database (BrianWeb): \url{http://www.bic.mni.mcgill.ca/brainweb/}. The two images are generated by T1 modality with slice thickness of 1mm resolution, 9\% noise and 20\% intensity non-uniformity. Here, the two images are represented two slices in the axial plane with the sequence of 100 and 110. Moreover, there are golden standard segmentations in the dataset. We set the numbers of clusters to 4. The visual comparisons are illustrated in Figs. 6 and 7.

\begin{figure}[htbp]
\centering
\begin{minipage}[t]{0.20\linewidth}
\centering
\includegraphics[width=1\textwidth]{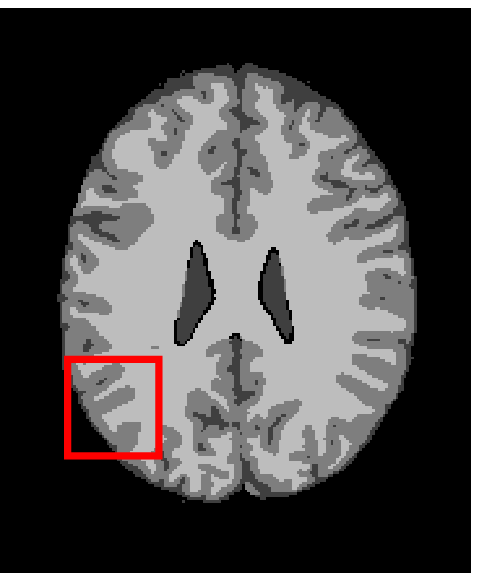}
\centerline{(a)}
\end{minipage}
\begin{minipage}[t]{0.20\linewidth}
\centering
\includegraphics[width=1\textwidth]{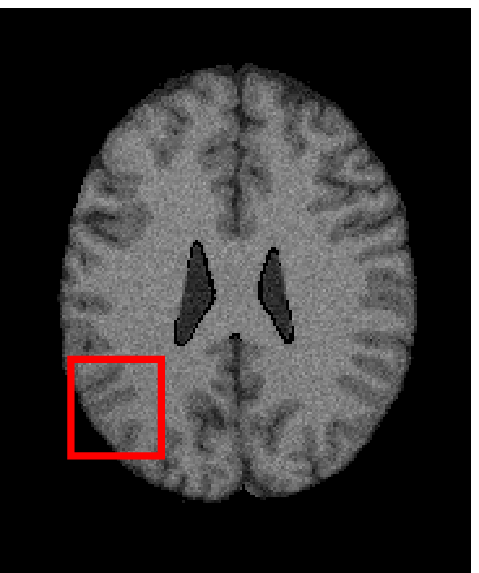}
\centerline{(b)}
\end{minipage}
\begin{minipage}[t]{0.20\linewidth}
\centering
\includegraphics[width=1\textwidth]{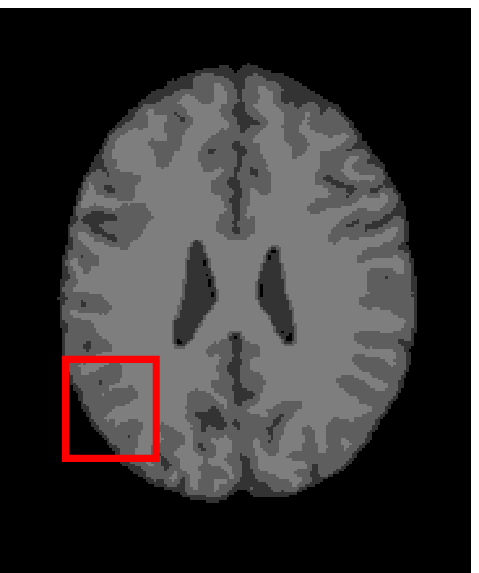}
\centerline{(c)}
\end{minipage}
\begin{minipage}[t]{0.20\linewidth}
\centering
\includegraphics[width=1\textwidth]{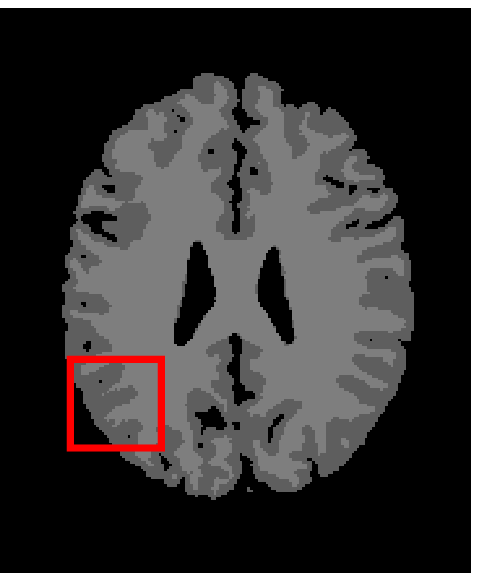}
\centerline{(d)}
\end{minipage}
\begin{minipage}[t]{0.20\linewidth}
\centering
\includegraphics[width=1\textwidth]{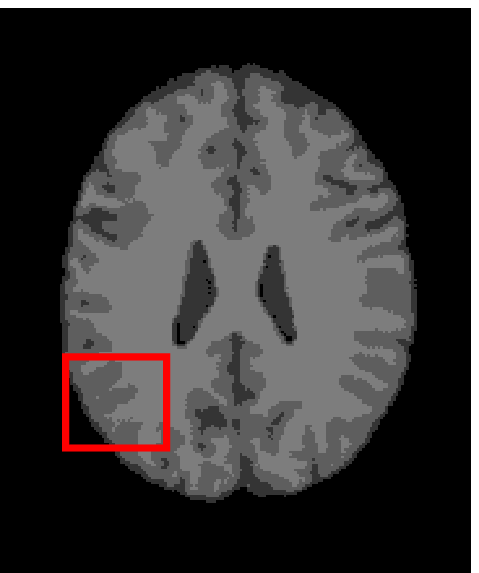}
\centerline{(e)}
\end{minipage}
\begin{minipage}[t]{0.20\linewidth}
\centering
\includegraphics[width=1\textwidth]{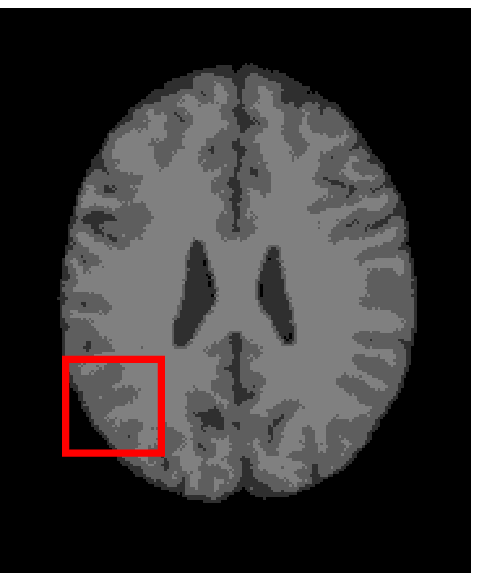}
\centerline{(f)}
\end{minipage}
\begin{minipage}[t]{0.20\linewidth}
\centering
\includegraphics[width=1\textwidth]{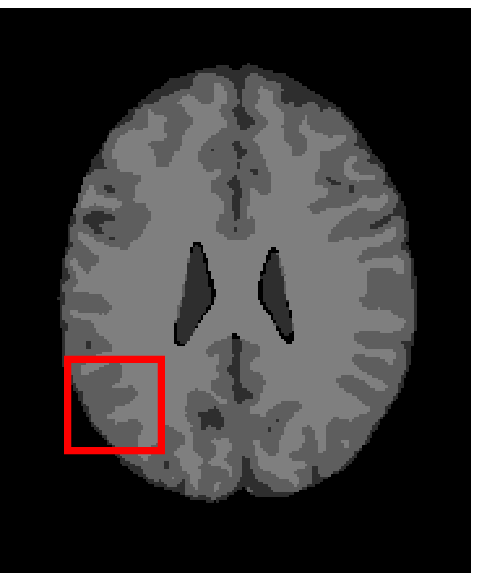}
\centerline{(g)}
\end{minipage}
\begin{minipage}[t]{0.20\linewidth}
\centering
\includegraphics[width=1\textwidth]{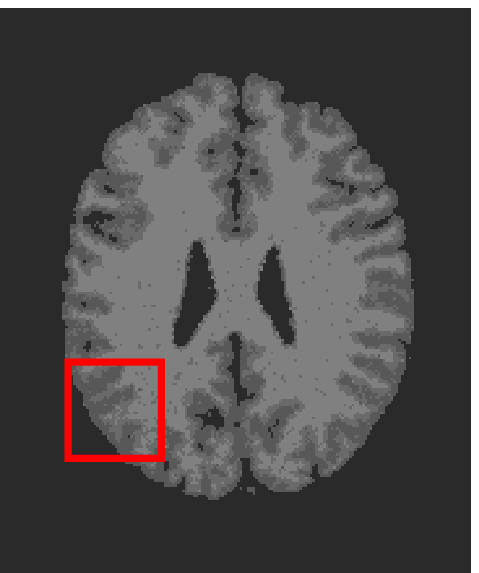}
\centerline{(h)}
\end{minipage}
\begin{minipage}[t]{0.20\linewidth}
\centering
\includegraphics[width=1\textwidth]{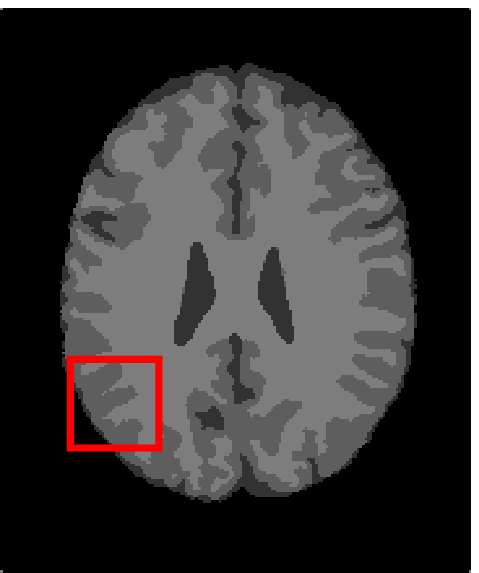}
\centerline{(i)}
\end{minipage}
\begin{minipage}[t]{0.20\linewidth}
\centering
\includegraphics[width=1\textwidth]{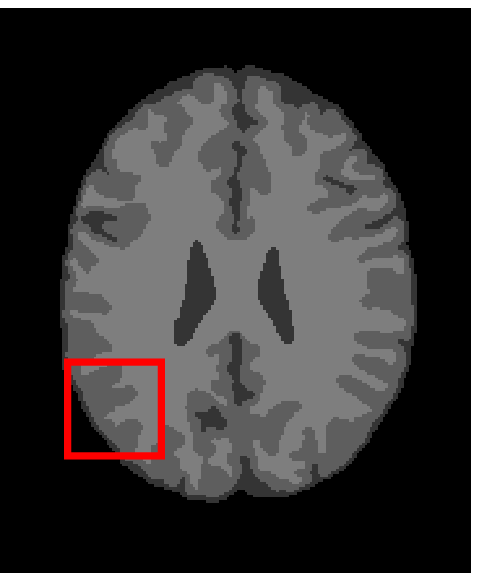}
\centerline{(j)}
\end{minipage}
\begin{minipage}[t]{0.20\linewidth}
\centering
\includegraphics[width=1\textwidth]{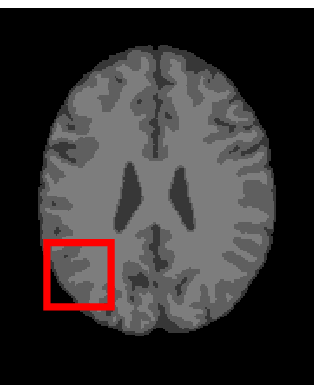}
\centerline{(k)}
\end{minipage}
\begin{minipage}[t]{0.20\linewidth}
\centering
\includegraphics[width=1\textwidth]{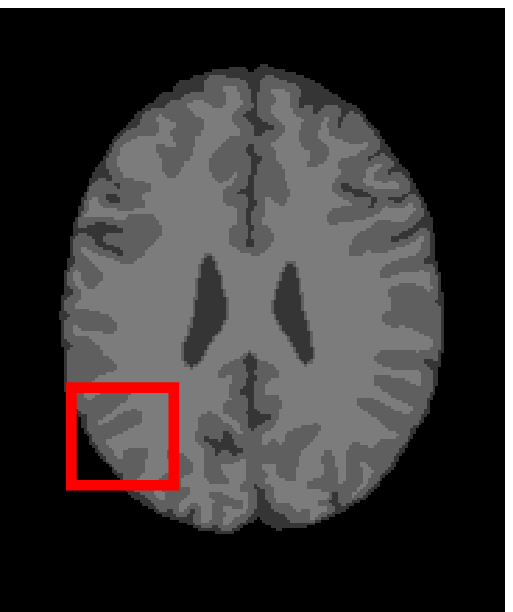}
\centerline{(l)}
\end{minipage}
\caption{Segmentation results for the first medical image. From (a) to (l): ground truth, noisy image and results of FCM\_S1, FCM\_S2, FGFCM, FLICM, KWFLICM, ARKFCM, FRFCM, WFCM, DSFCM\_N, and LRFCM.}
\end{figure}

\begin{figure}[htb]
\centering
\begin{minipage}[t]{0.20\linewidth}
\centering
\includegraphics[width=1\textwidth]{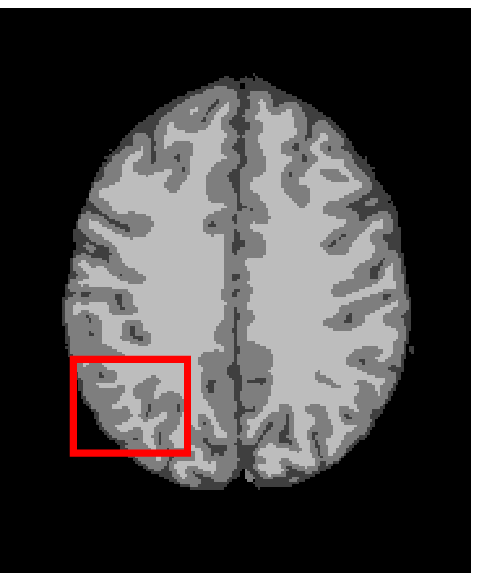}
\centerline{(a)}
\end{minipage}
\begin{minipage}[t]{0.20\linewidth}
\centering
\includegraphics[width=1\textwidth]{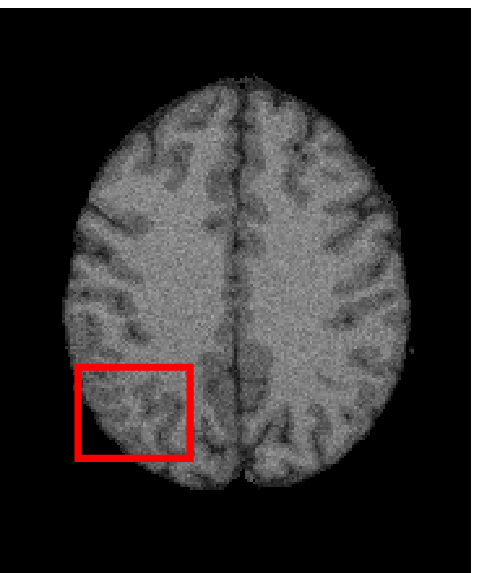}
\centerline{(b)}
\end{minipage}
\begin{minipage}[t]{0.20\linewidth}
\centering
\includegraphics[width=1\textwidth]{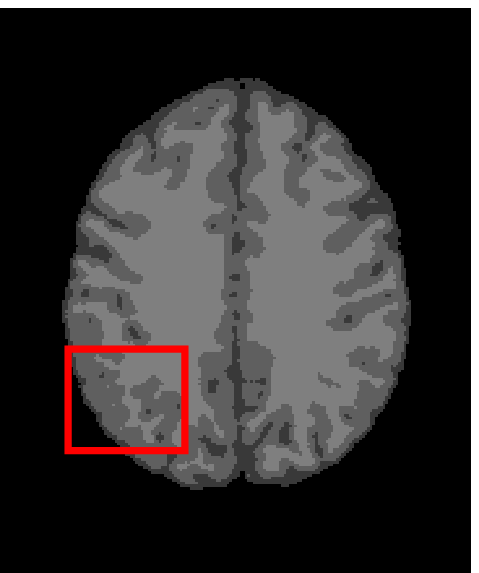}
\centerline{(c)}
\end{minipage}
\begin{minipage}[t]{0.20\linewidth}
\centering
\includegraphics[width=1\textwidth]{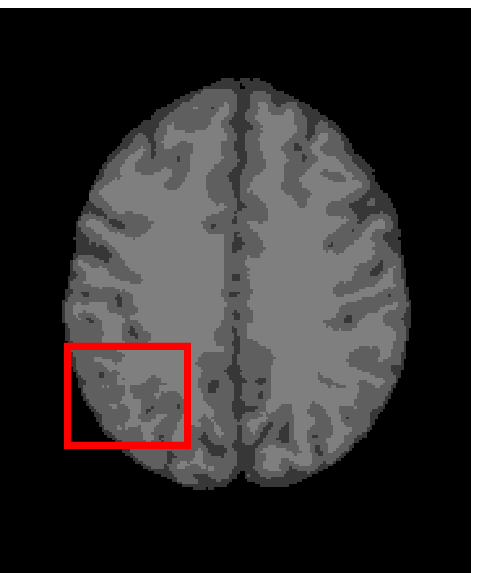}
\centerline{(d)}
\end{minipage}
\begin{minipage}[t]{0.20\linewidth}
\centering
\includegraphics[width=1\textwidth]{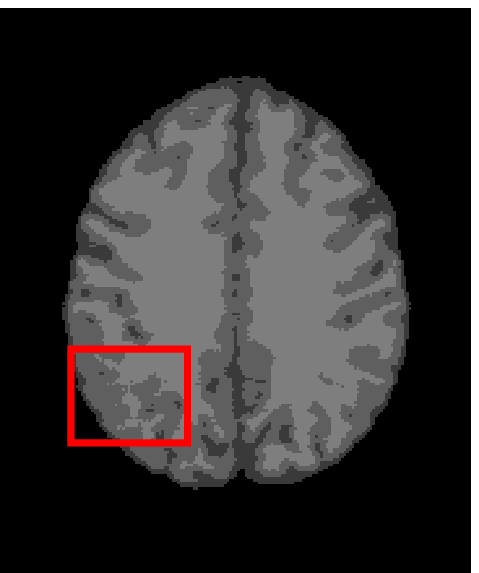}
\centerline{(e)}
\end{minipage}
\begin{minipage}[t]{0.20\linewidth}
\centering
\includegraphics[width=1\textwidth]{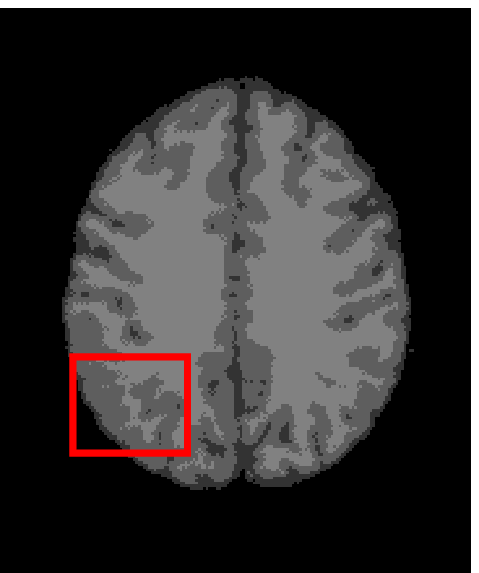}
\centerline{(f)}
\end{minipage}
\begin{minipage}[t]{0.20\linewidth}
\centering
\includegraphics[width=1\textwidth]{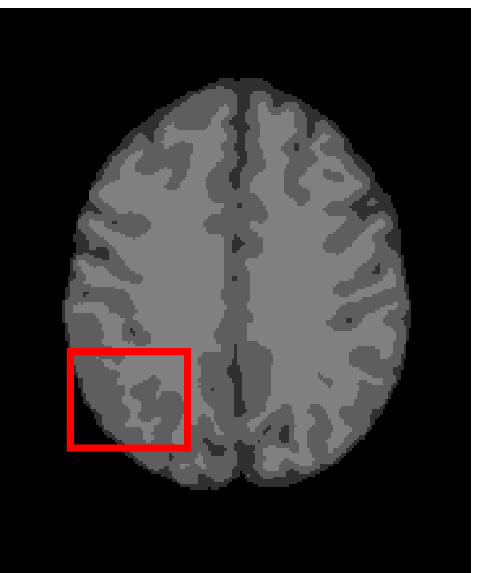}
\centerline{(g)}
\end{minipage}
\begin{minipage}[t]{0.20\linewidth}
\centering
\includegraphics[width=1\textwidth]{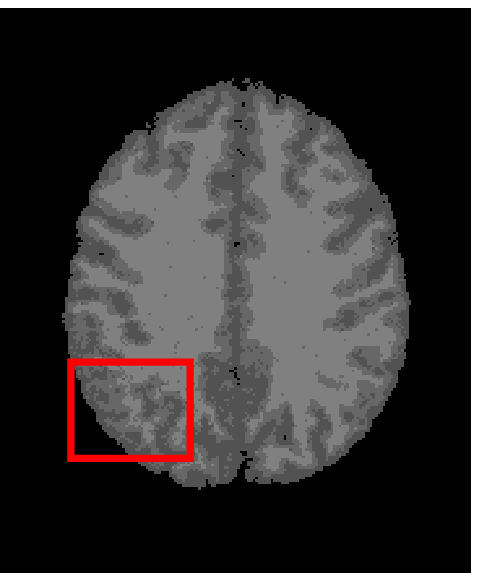}
\centerline{(h)}
\end{minipage}
\begin{minipage}[t]{0.20\linewidth}
\centering
\includegraphics[width=1\textwidth]{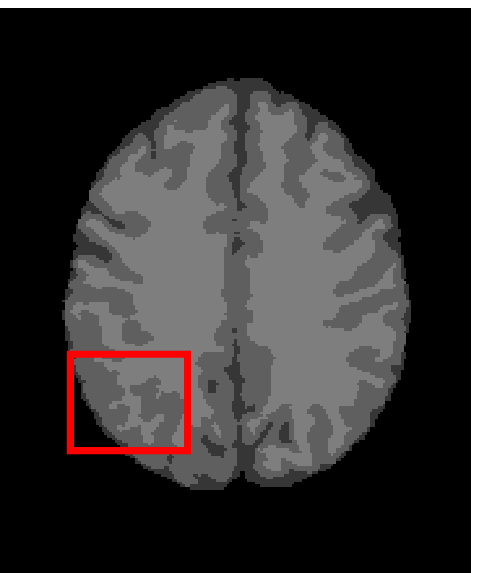}
\centerline{(i)}
\end{minipage}
\begin{minipage}[t]{0.20\linewidth}
\centering
\includegraphics[width=1\textwidth]{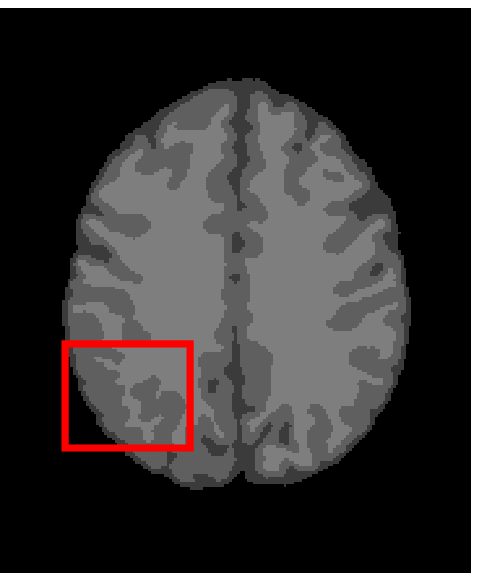}
\centerline{(j)}
\end{minipage}
\begin{minipage}[t]{0.20\linewidth}
\centering
\includegraphics[width=1\textwidth]{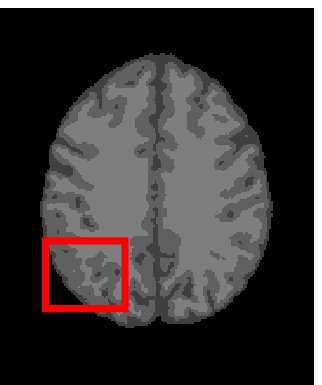}
\centerline{(k)}
\end{minipage}
\begin{minipage}[t]{0.20\linewidth}
\centering
\includegraphics[width=1\textwidth]{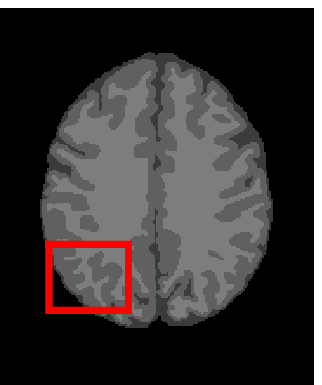}
\centerline{(l)}
\end{minipage}
\caption{Segmentation results for the second medical image. From (a) to (l): ground truth, noisy image and results of FCM\_S1, FCM\_S2, FGFCM, FLICM, KWFLICM, ARKFCM, FRFCM, WFCM, DSFCM\_N, and LRFCM.}
\end{figure}

By focusing on the marked red square in Figs. 6 and 7, we easily find that FCM\_S1, FCM\_S2, FGFCM and ARKFCM are sensitive to noise. FLICM and KWFLICM are vulnerable to severe intensity inhomogeneity. FRFCM brings overly smooth results due to the use of gray level histograms. WFCM and DSFCM\_N cause several contours to change. However, LRFCM acquires clear contours and suppresses noise adequately. Moreover, we find that the segmentation result of LRFCM is closer to ground truth.

In the last experiments, we segment some Red-Green-Blue (RGB) color images. It is easy to extend LRFCM to color image segmentation. We apply the multivariate MR to color images \cite{Lei2017}. Moreover, we conduct the tight wavelet frame transform in each channel of an RGB color image. The dimensionality of the obtained feature set is three times higher than that of a gray image. The remaining settings are similar to those in gray image segmentation. In the following, we choose two sets of color images.

First, we segment four color images coming from the Berkeley Segmentation Dataset (BSDS300): \url{https://www2.eecs.berkeley.edu/Research/Projects/CS/vision/bsds/BSDS300/html/dataset/images.html}. The numbers of clusters are set to 3, 2, 2, and 2, respectively. The segmentation results are shown in Fig. 8.
\begin{figure}[h]
\centering
\begin{minipage}[t]{0.23\linewidth}
\centering
\includegraphics[width=1\textwidth]{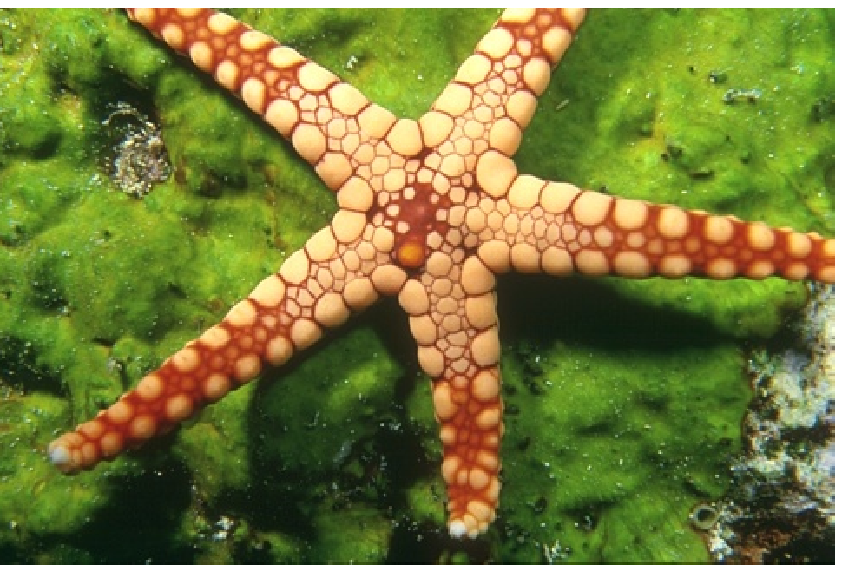}
\end{minipage}
\begin{minipage}[t]{0.23\linewidth}
\centering
\includegraphics[width=1\textwidth]{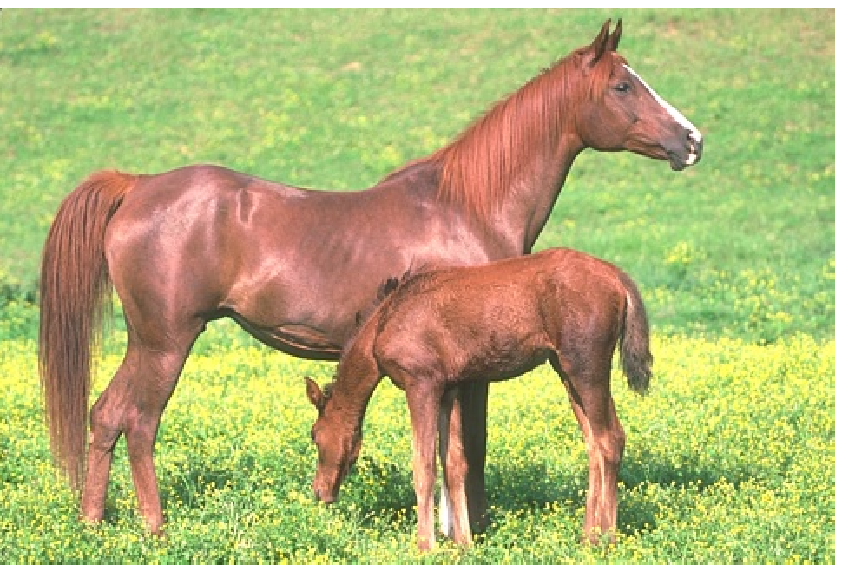}
\end{minipage}
\begin{minipage}[t]{0.23\linewidth}
\centering
\includegraphics[width=1\textwidth]{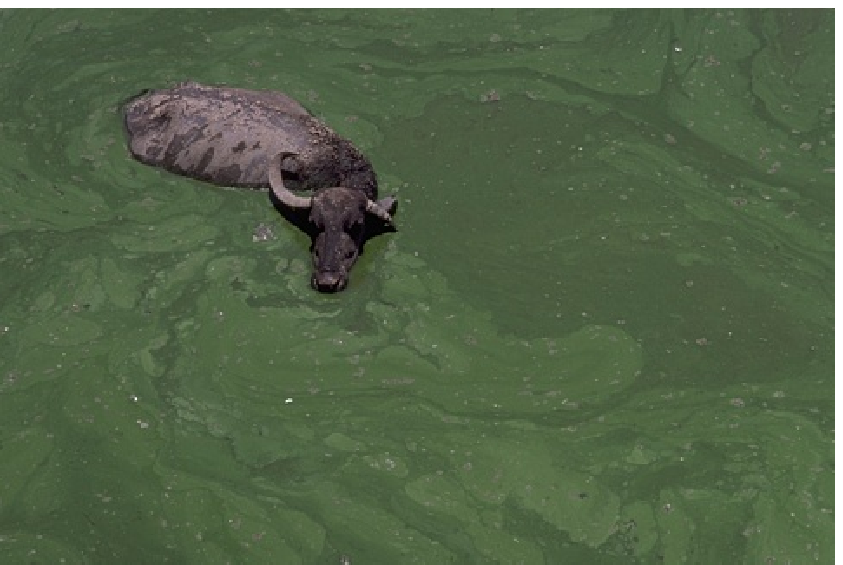}
\end{minipage}
\begin{minipage}[t]{0.23\linewidth}
\centering
\includegraphics[width=1\textwidth]{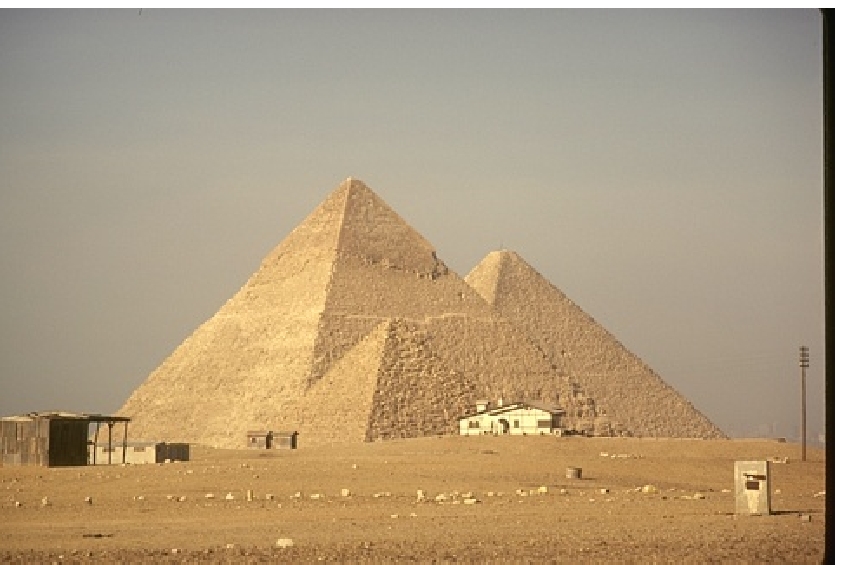}
\end{minipage}\\
\begin{minipage}[t]{0.23\linewidth}
\centering
\includegraphics[width=1\textwidth]{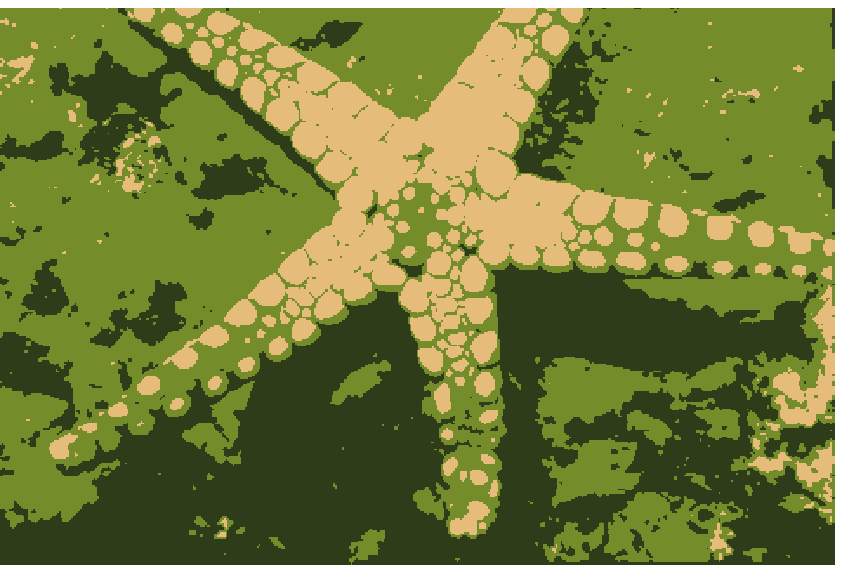}
\end{minipage}
\begin{minipage}[t]{0.23\linewidth}
\centering
\includegraphics[width=1\textwidth]{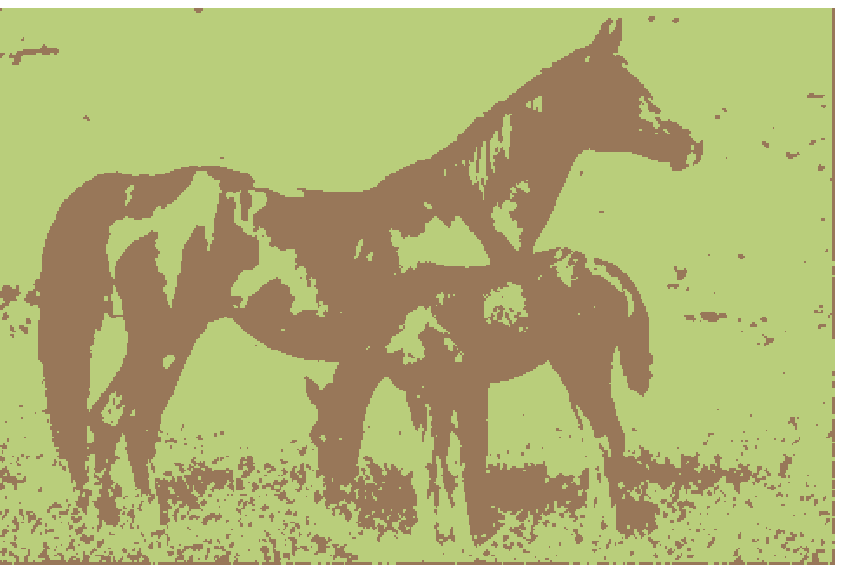}
\end{minipage}
\begin{minipage}[t]{0.23\linewidth}
\centering
\includegraphics[width=1\textwidth]{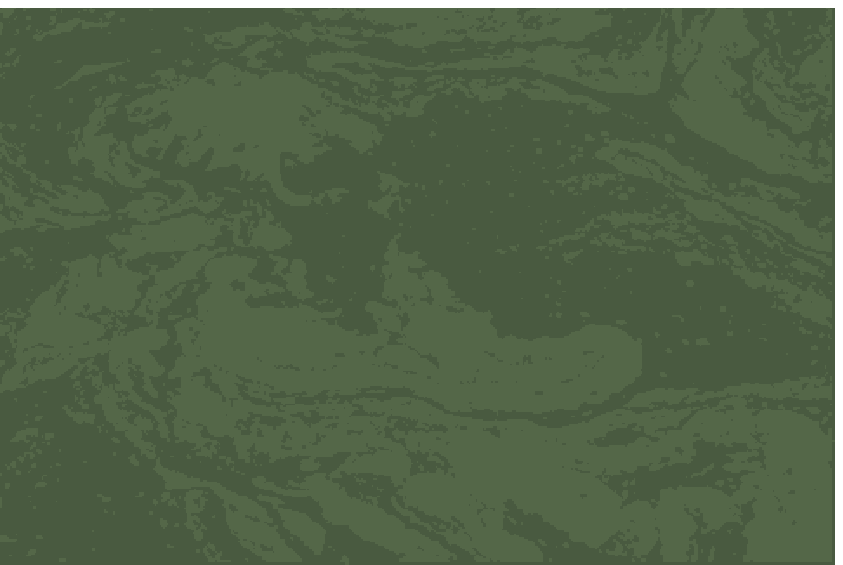}
\end{minipage}
\begin{minipage}[t]{0.23\linewidth}
\centering
\includegraphics[width=1\textwidth]{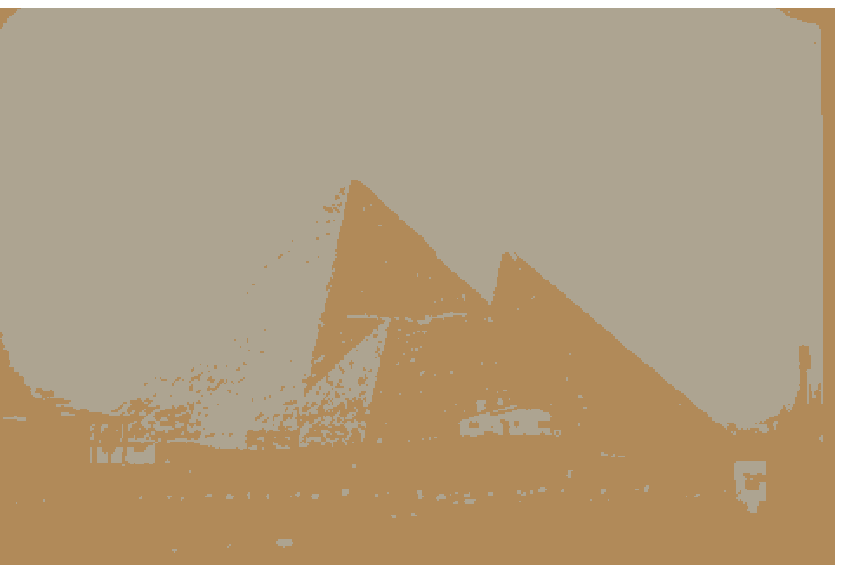}
\end{minipage}\\
\begin{minipage}[t]{0.23\linewidth}
\centering
\includegraphics[width=1\textwidth]{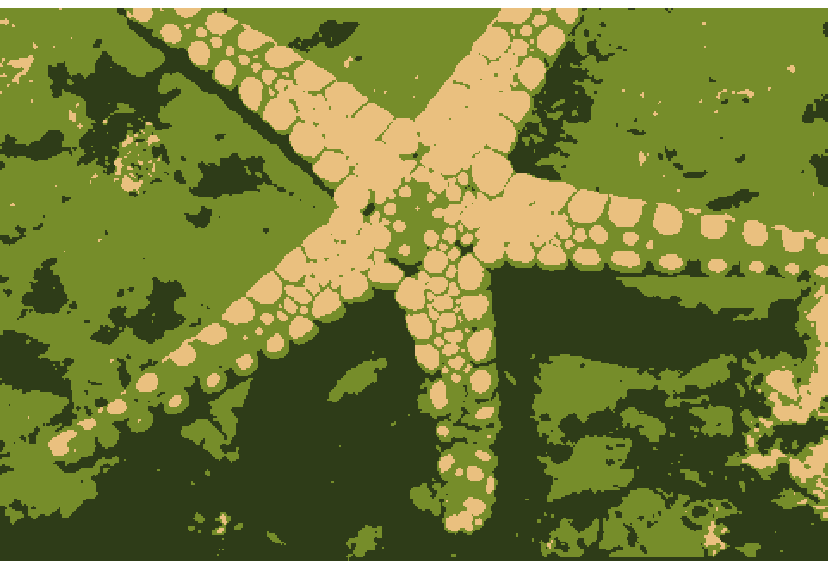}
\end{minipage}
\begin{minipage}[t]{0.23\linewidth}
\centering
\includegraphics[width=1\textwidth]{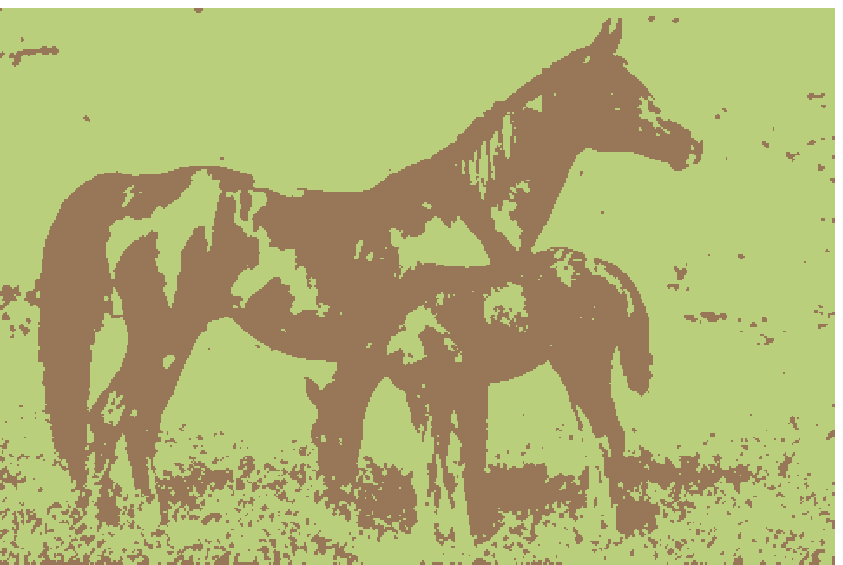}
\end{minipage}
\begin{minipage}[t]{0.23\linewidth}
\centering
\includegraphics[width=1\textwidth]{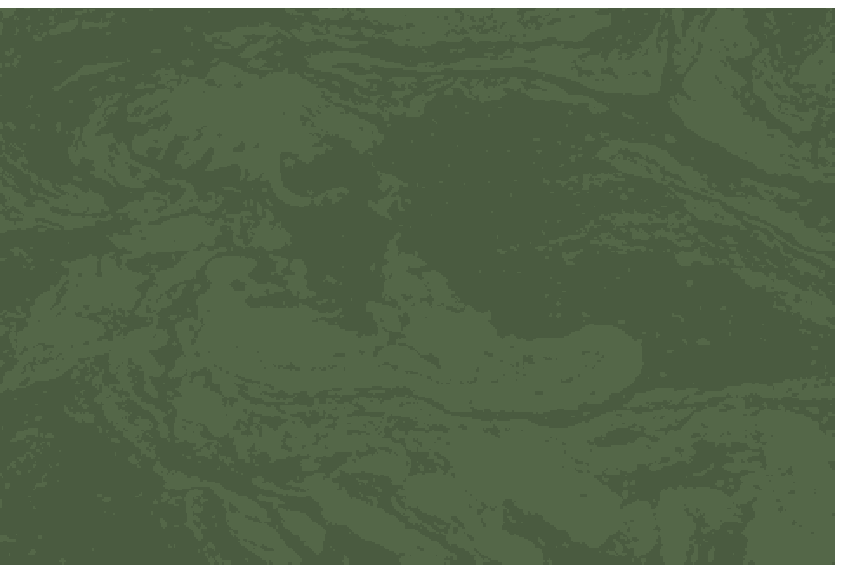}
\end{minipage}
\begin{minipage}[t]{0.23\linewidth}
\centering
\includegraphics[width=1\textwidth]{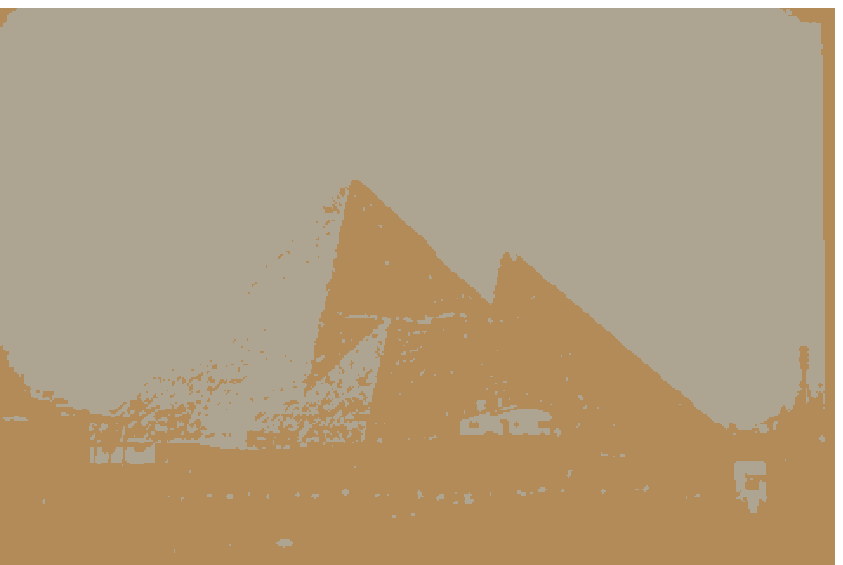}
\end{minipage}\\
\begin{minipage}[t]{0.23\linewidth}
\centering
\includegraphics[width=1\textwidth]{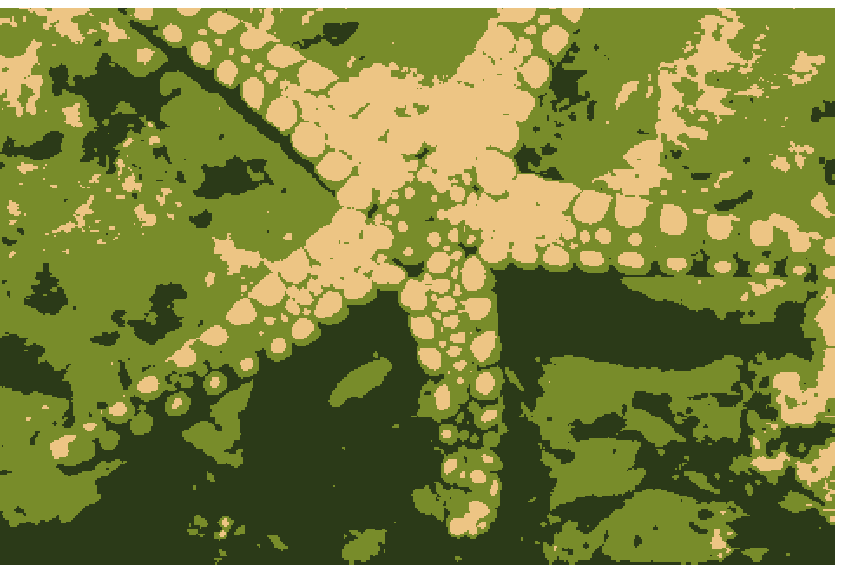}
\end{minipage}
\begin{minipage}[t]{0.23\linewidth}
\centering
\includegraphics[width=1\textwidth]{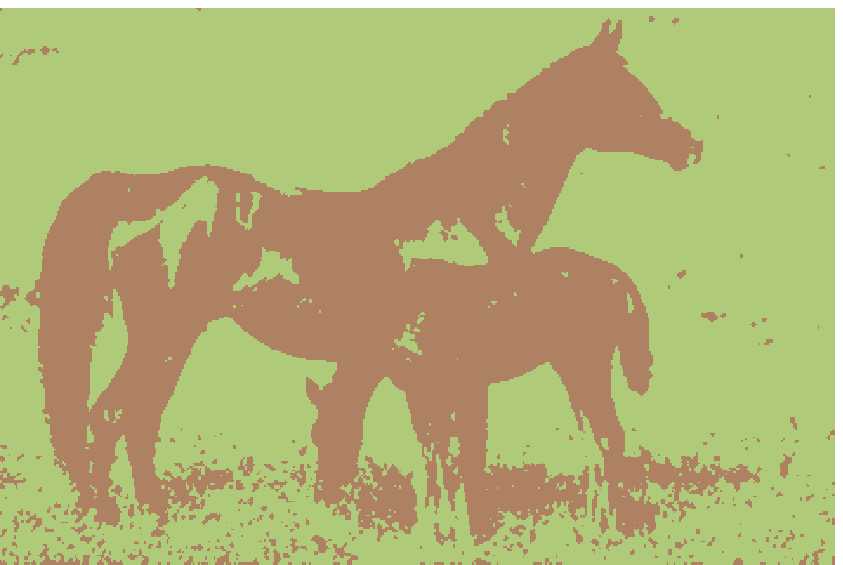}
\end{minipage}
\begin{minipage}[t]{0.23\linewidth}
\centering
\includegraphics[width=1\textwidth]{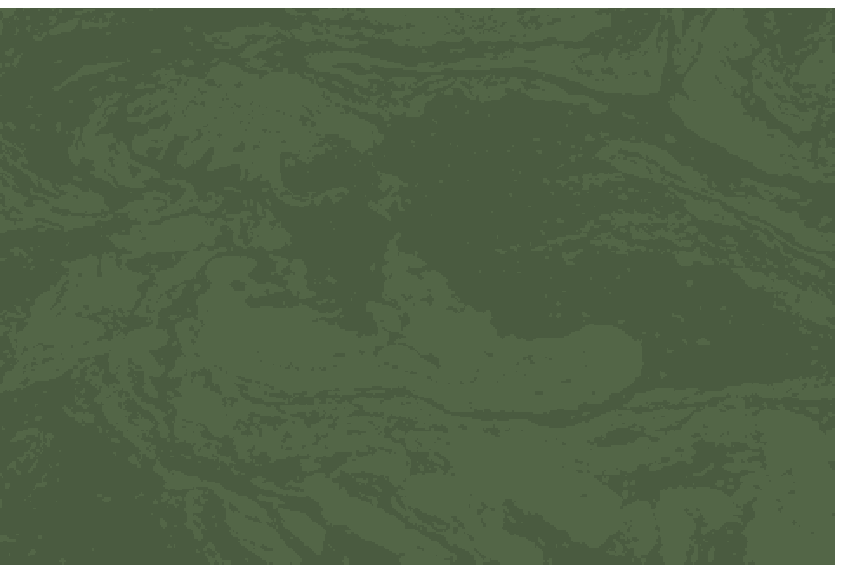}
\end{minipage}
\begin{minipage}[t]{0.23\linewidth}
\centering
\includegraphics[width=1\textwidth]{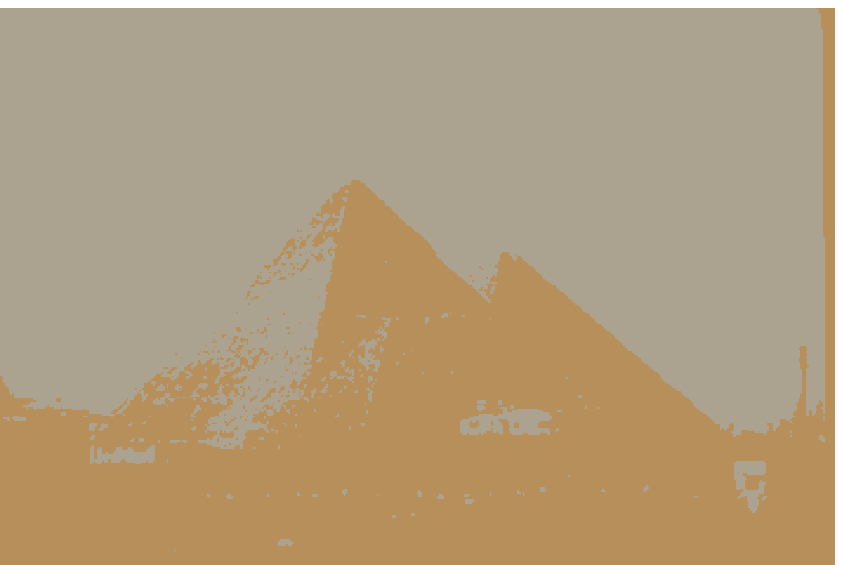}
\end{minipage}\\
\begin{minipage}[t]{0.23\linewidth}
\centering
\includegraphics[width=1\textwidth]{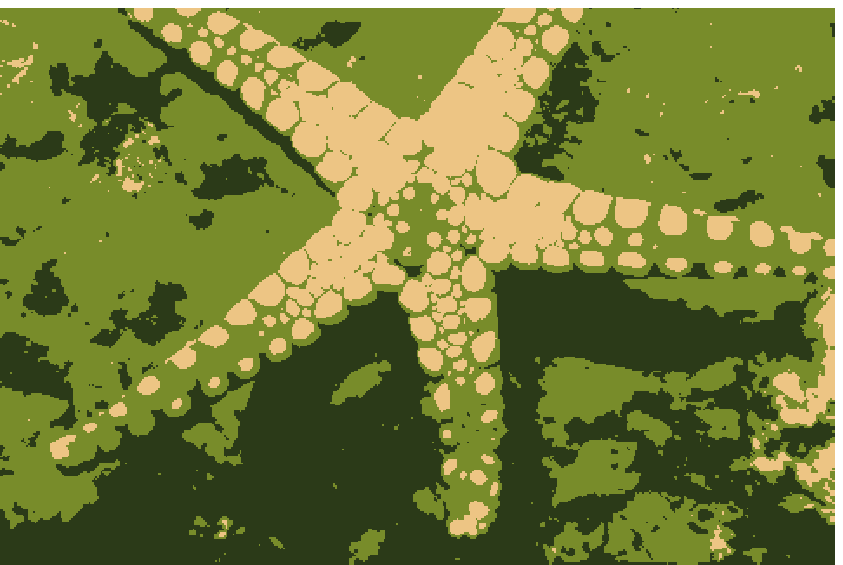}
\end{minipage}
\begin{minipage}[t]{0.23\linewidth}
\centering
\includegraphics[width=1\textwidth]{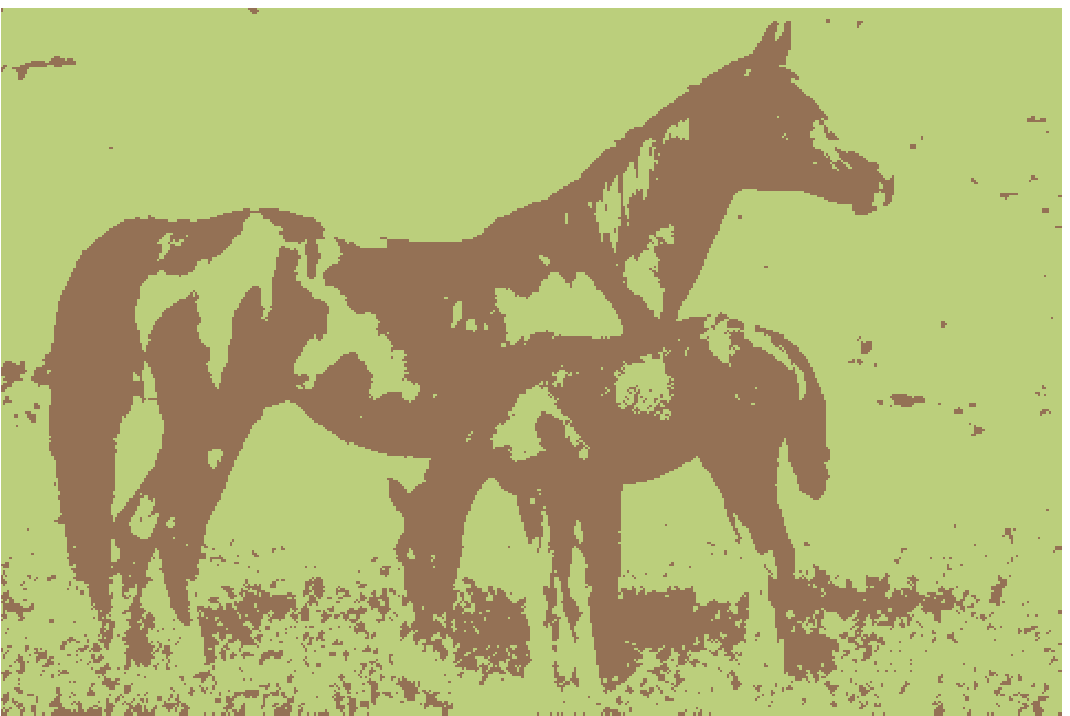}
\end{minipage}
\begin{minipage}[t]{0.23\linewidth}
\centering
\includegraphics[width=1\textwidth]{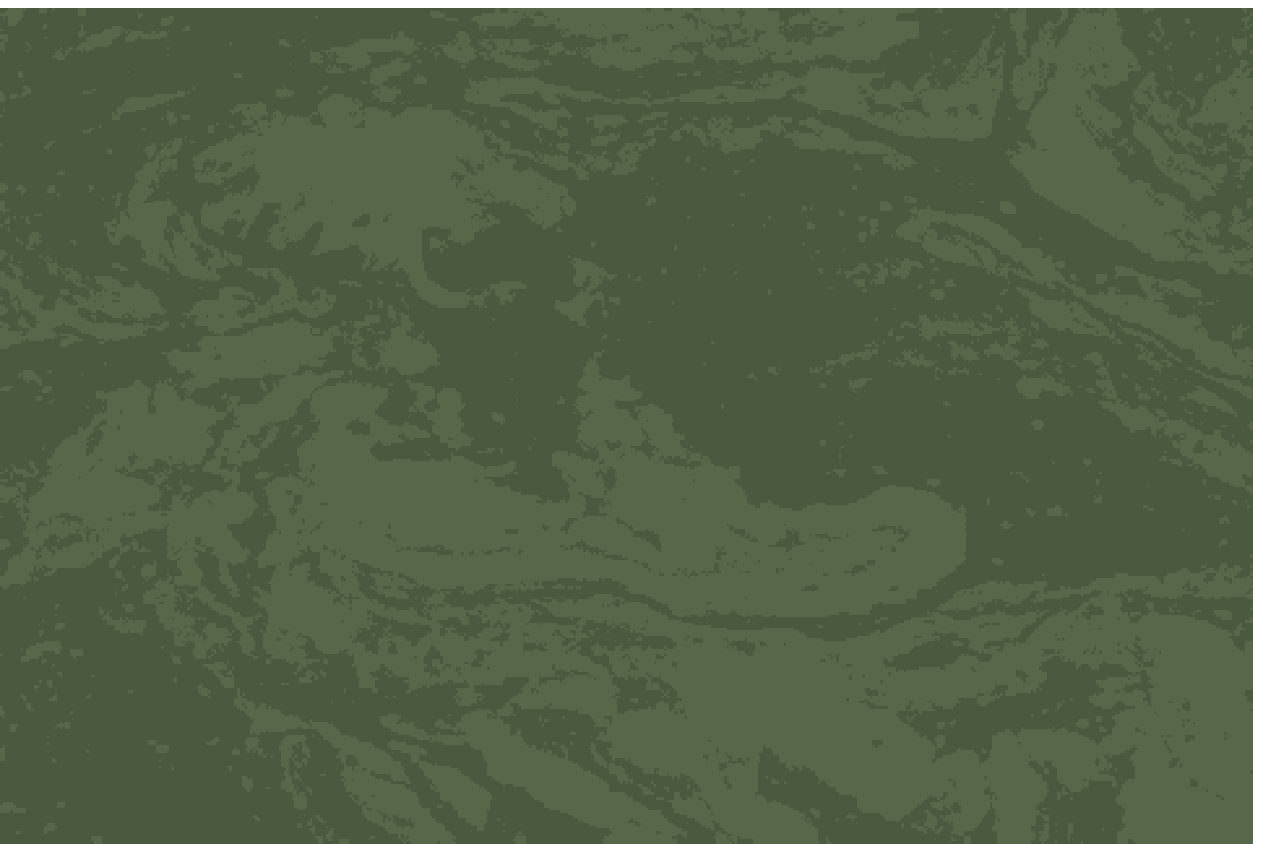}
\end{minipage}
\begin{minipage}[t]{0.23\linewidth}
\centering
\includegraphics[width=1\textwidth]{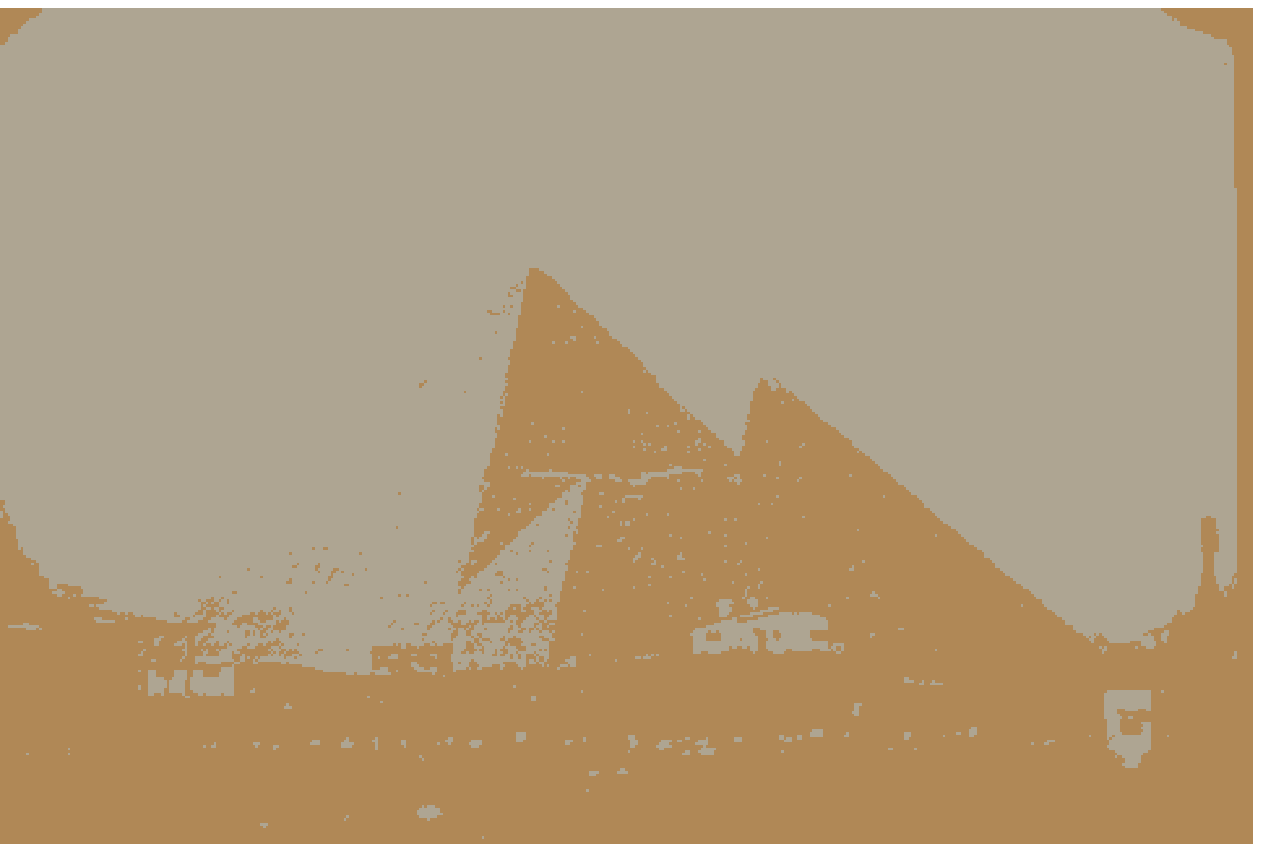}
\end{minipage}\\
\begin{minipage}[t]{0.23\linewidth}
\centering
\includegraphics[width=1\textwidth]{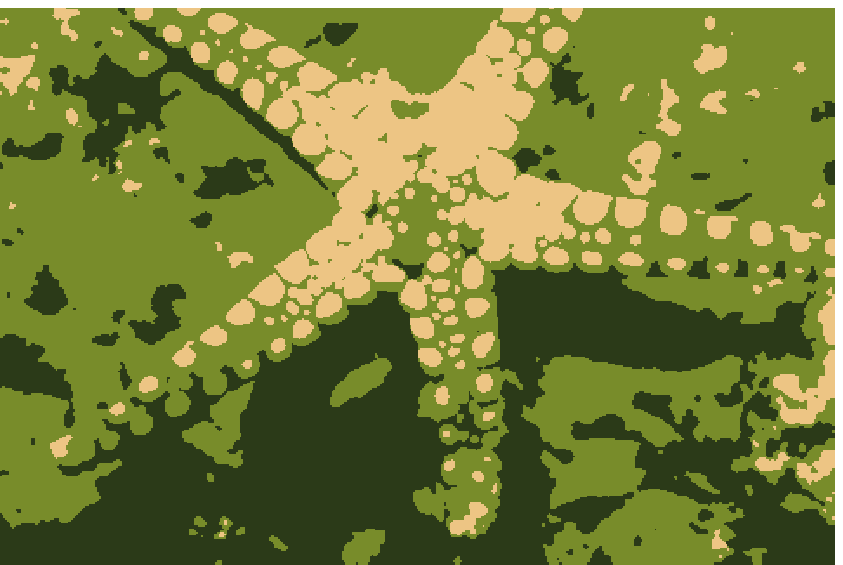}
\end{minipage}
\begin{minipage}[t]{0.23\linewidth}
\centering
\includegraphics[width=1\textwidth]{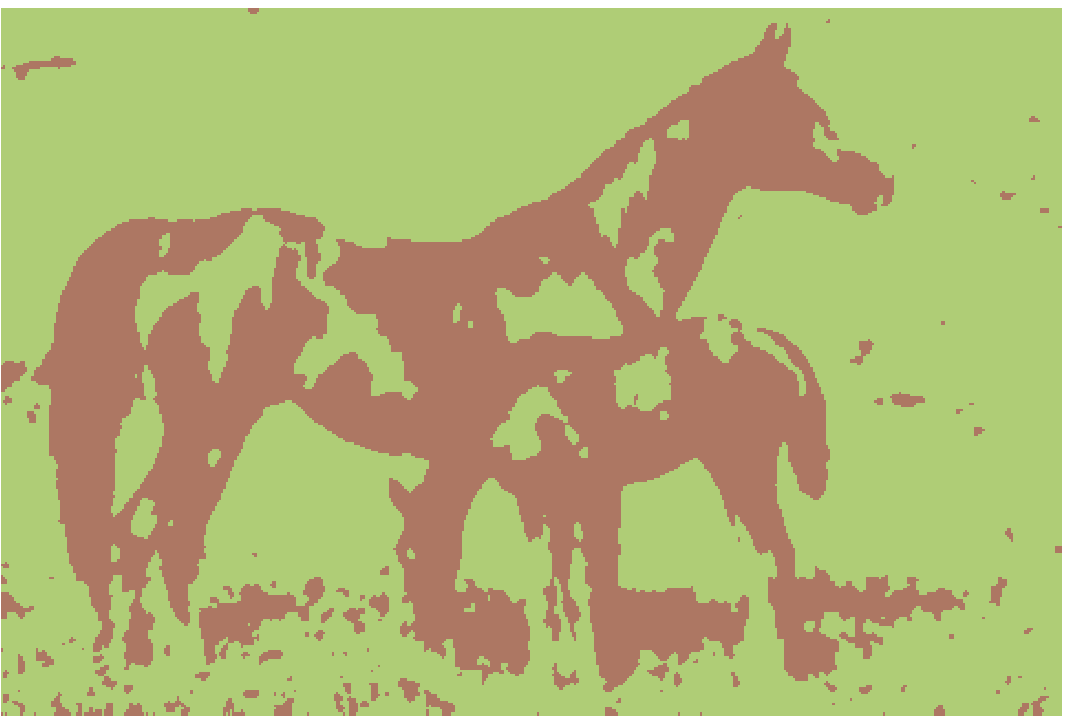}
\end{minipage}
\begin{minipage}[t]{0.23\linewidth}
\centering
\includegraphics[width=1\textwidth]{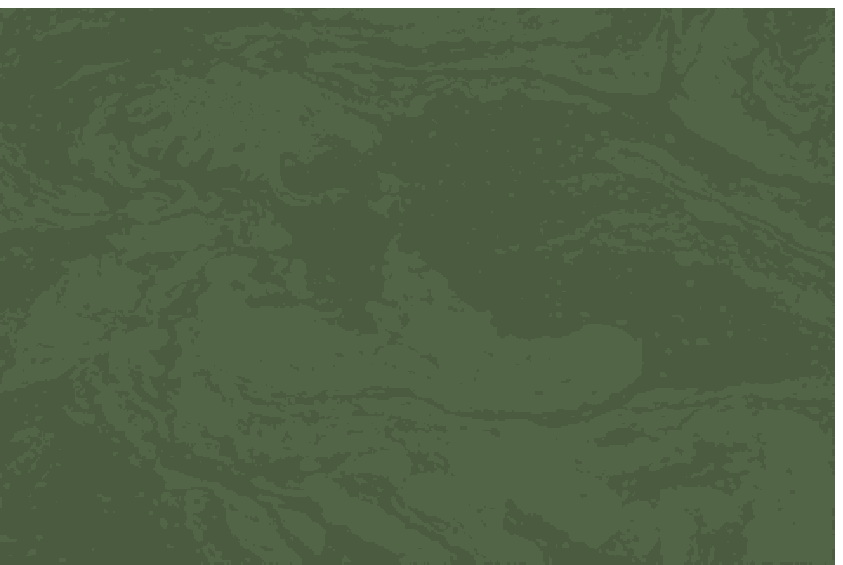}
\end{minipage}
\begin{minipage}[t]{0.23\linewidth}
\centering
\includegraphics[width=1\textwidth]{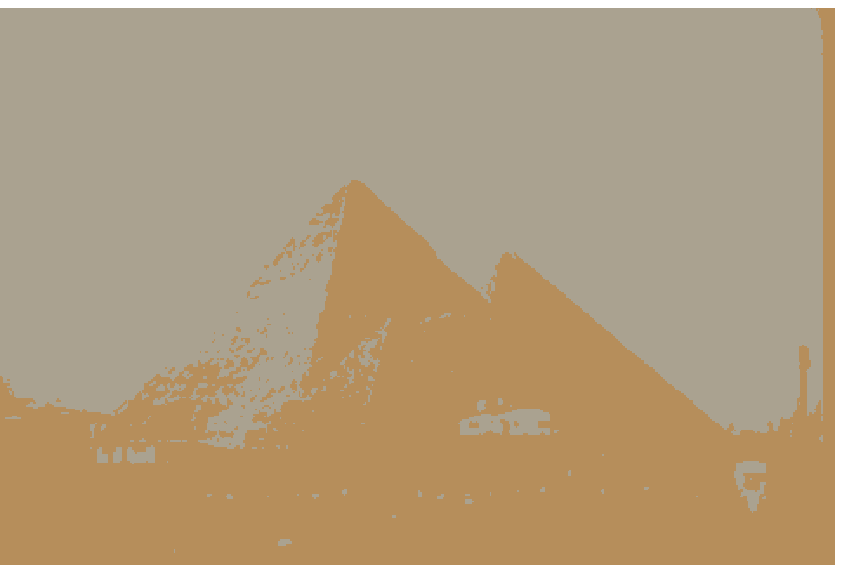}
\end{minipage}\\
\begin{minipage}[t]{0.23\linewidth}
\centering
\includegraphics[width=1\textwidth]{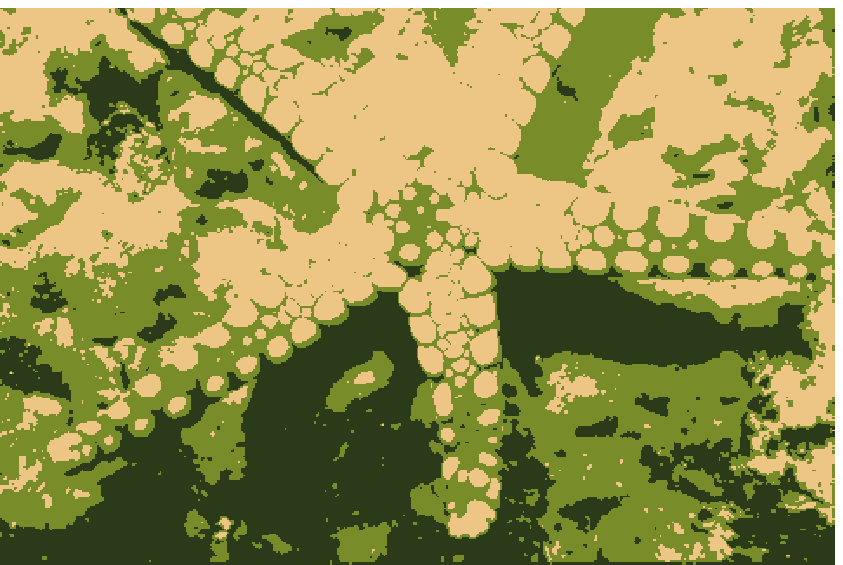}
\end{minipage}
\begin{minipage}[t]{0.23\linewidth}
\centering
\includegraphics[width=1\textwidth]{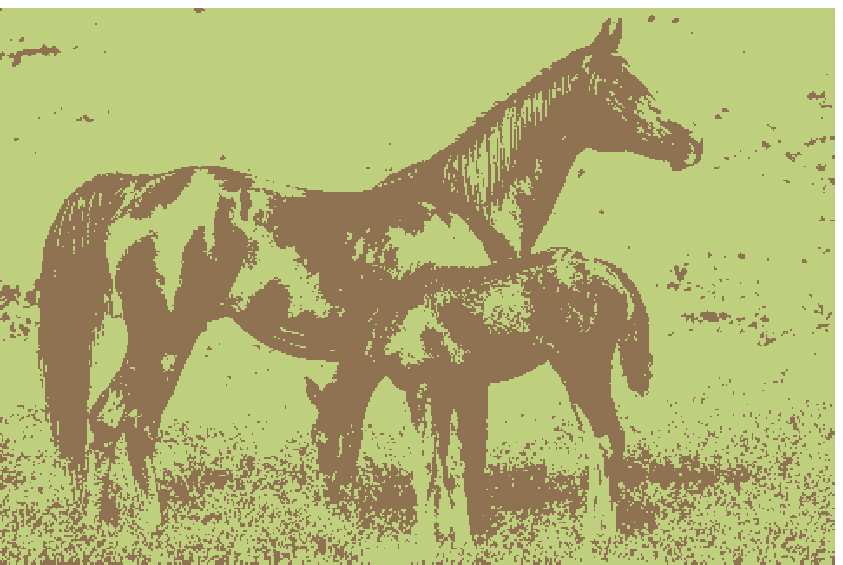}
\end{minipage}
\begin{minipage}[t]{0.23\linewidth}
\centering
\includegraphics[width=1\textwidth]{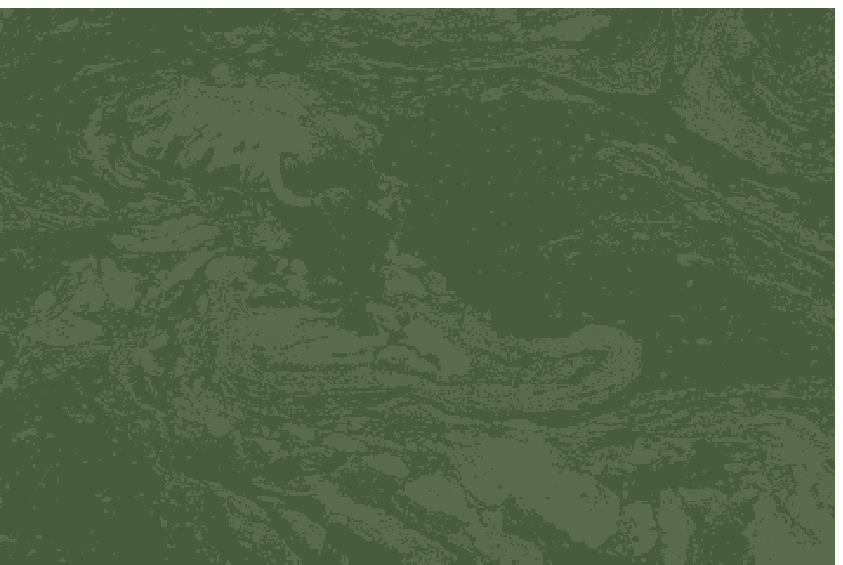}
\end{minipage}
\begin{minipage}[t]{0.23\linewidth}
\centering
\includegraphics[width=1\textwidth]{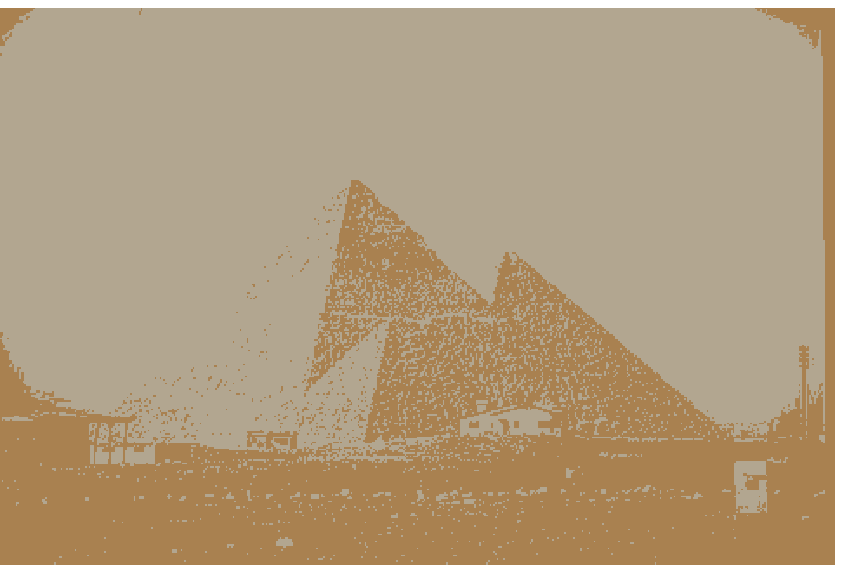}
\end{minipage}\\
\begin{minipage}[t]{0.23\linewidth}
\centering
\includegraphics[width=1\textwidth]{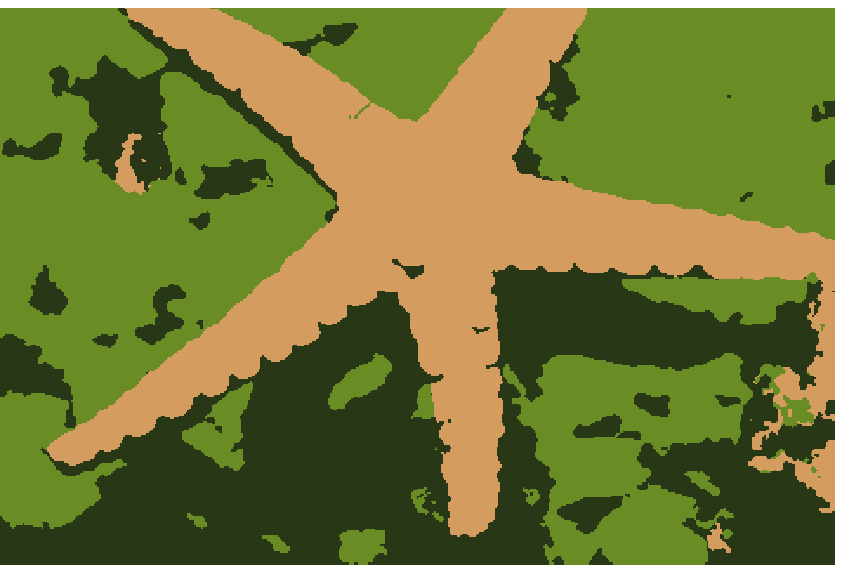}
\end{minipage}
\begin{minipage}[t]{0.23\linewidth}
\centering
\includegraphics[width=1\textwidth]{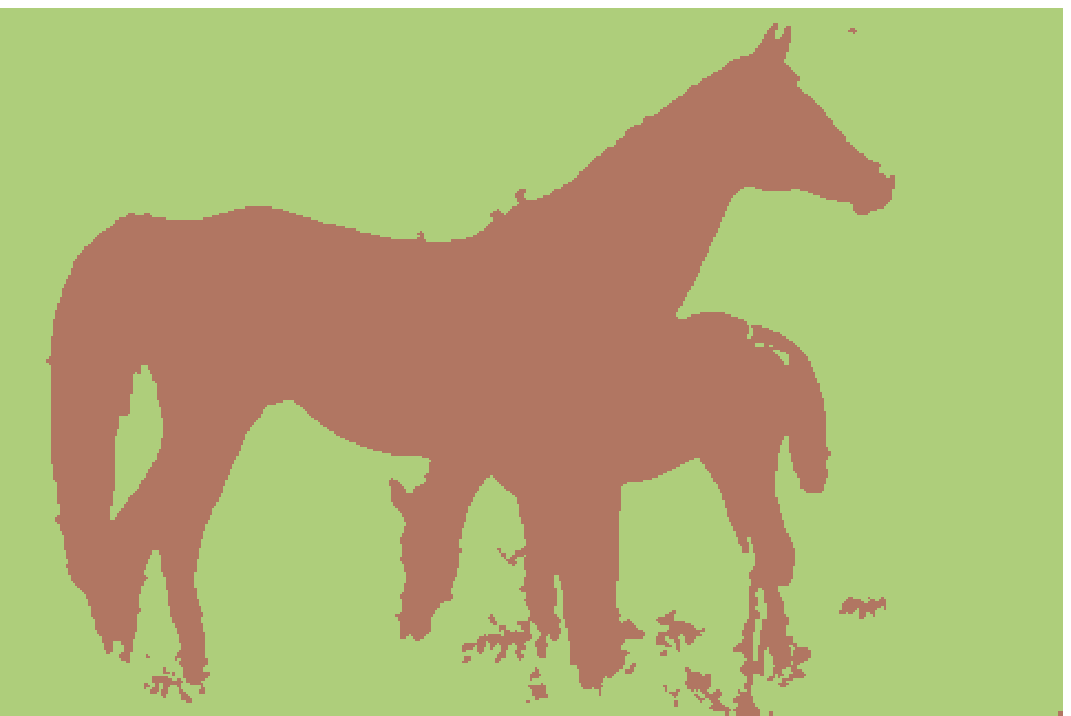}
\end{minipage}
\begin{minipage}[t]{0.23\linewidth}
\centering
\includegraphics[width=1\textwidth]{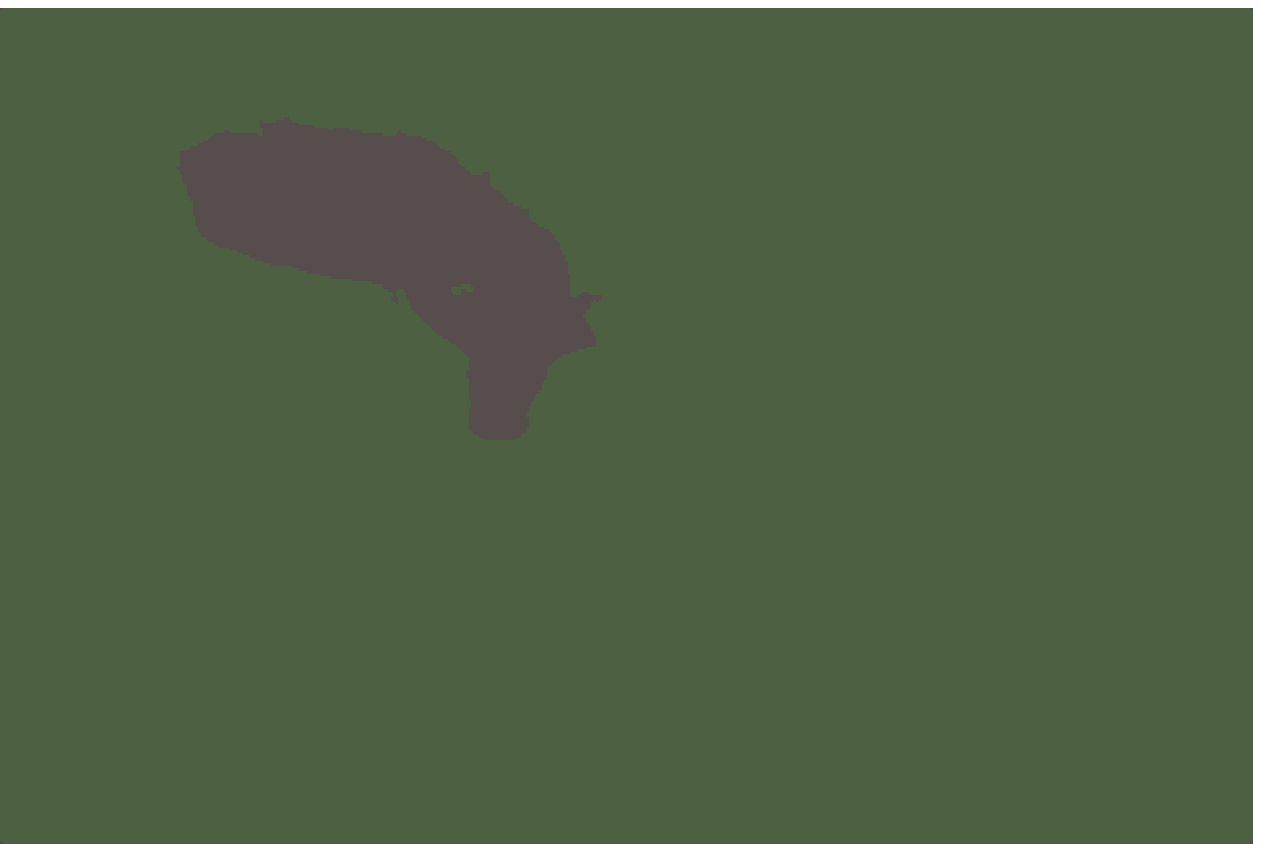}
\end{minipage}
\begin{minipage}[t]{0.23\linewidth}
\centering
\includegraphics[width=1\textwidth]{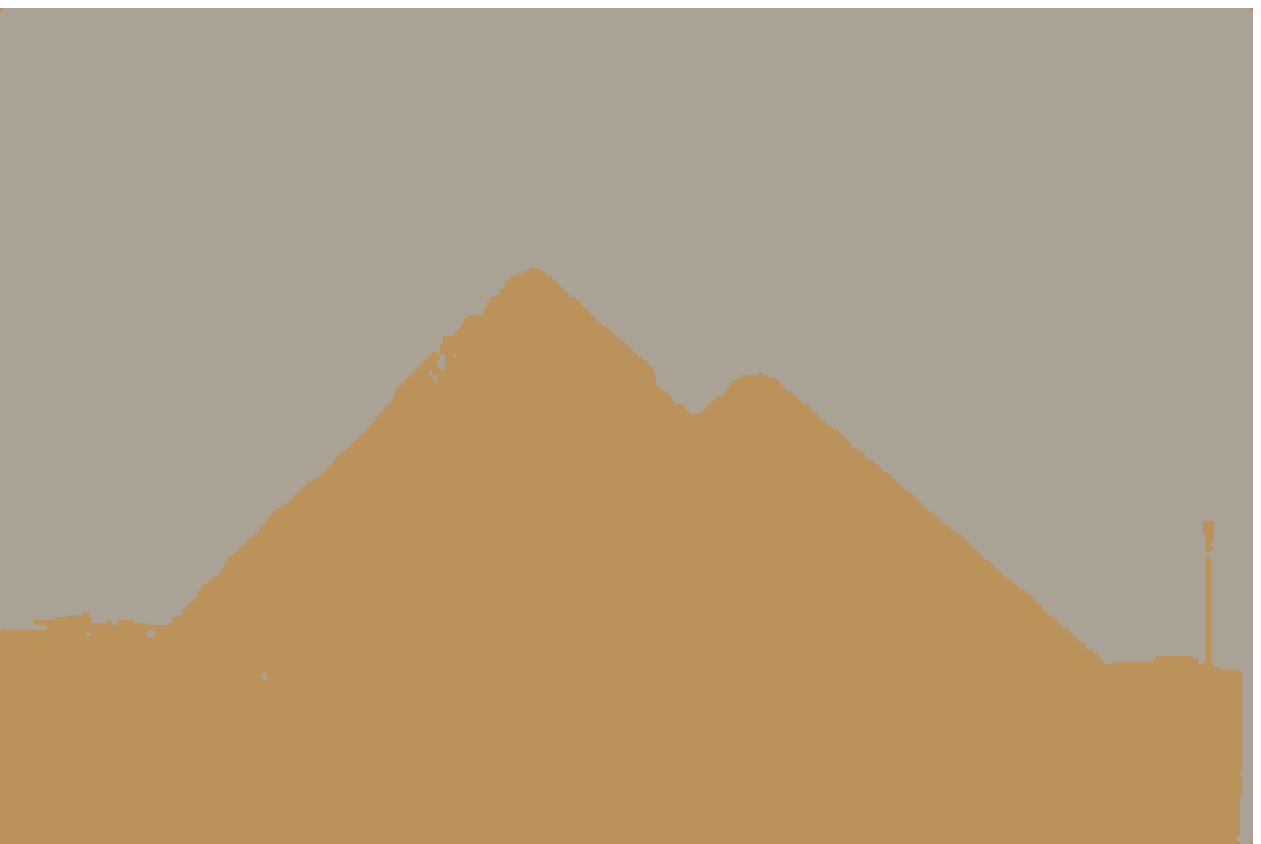}
\end{minipage}\\
\begin{minipage}[t]{0.23\linewidth}
\centering
\includegraphics[width=1\textwidth]{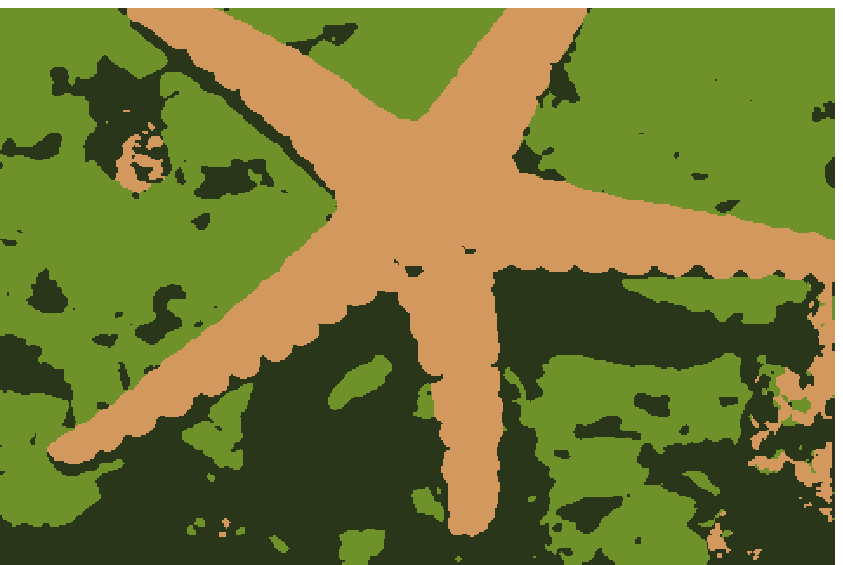}
\end{minipage}
\begin{minipage}[t]{0.23\linewidth}
\centering
\includegraphics[width=1\textwidth]{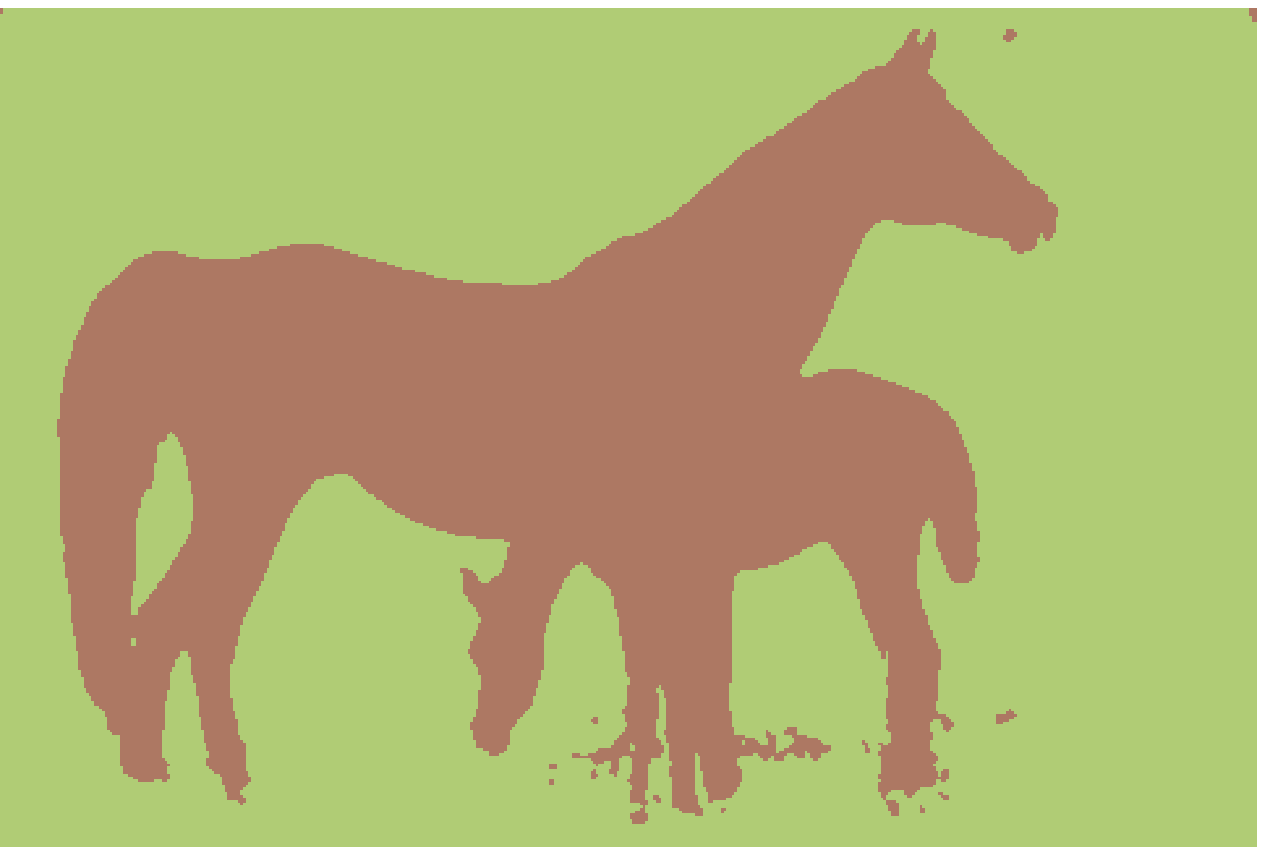}
\end{minipage}
\begin{minipage}[t]{0.23\linewidth}
\centering
\includegraphics[width=1\textwidth]{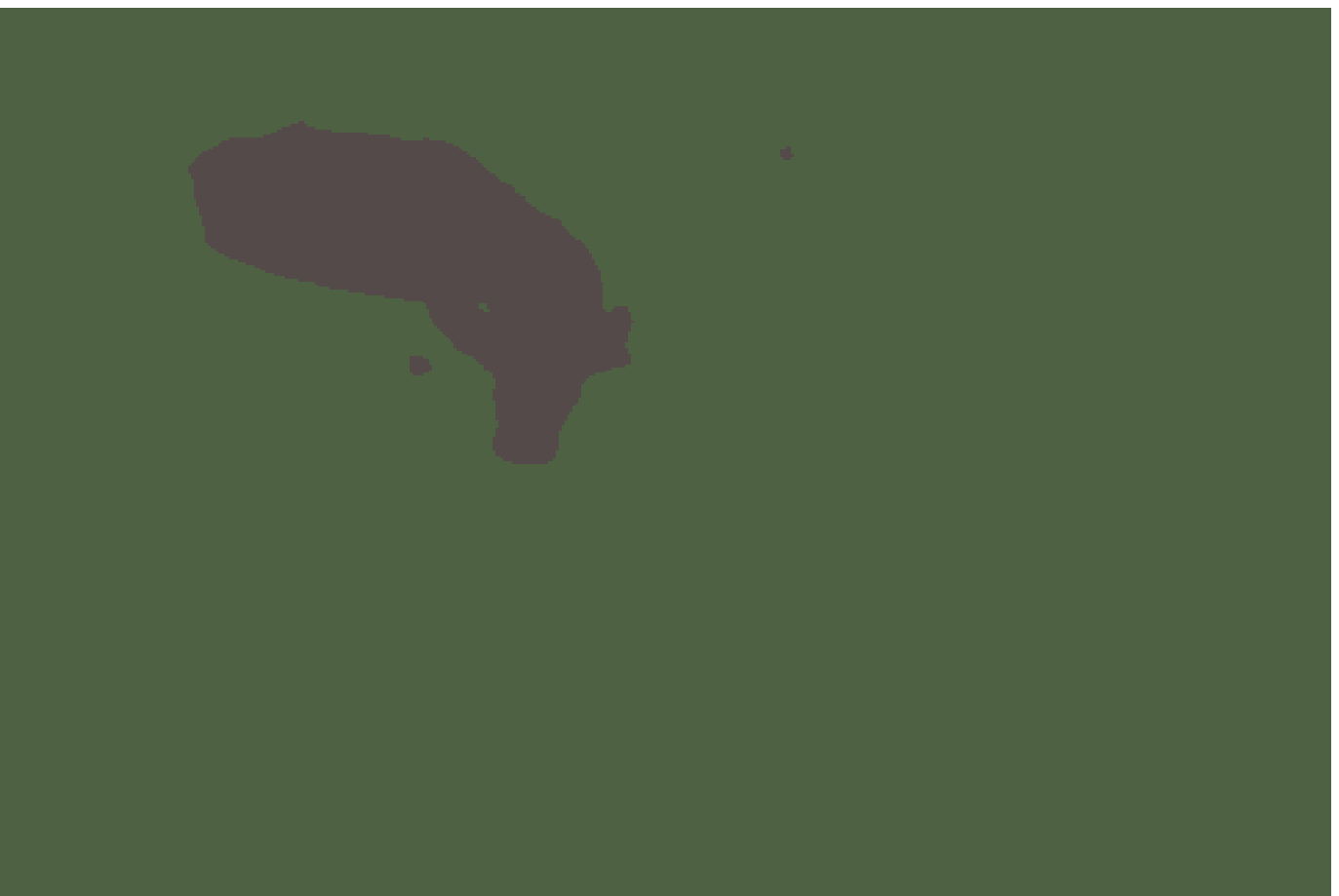}
\end{minipage}
\begin{minipage}[t]{0.23\linewidth}
\centering
\includegraphics[width=1\textwidth]{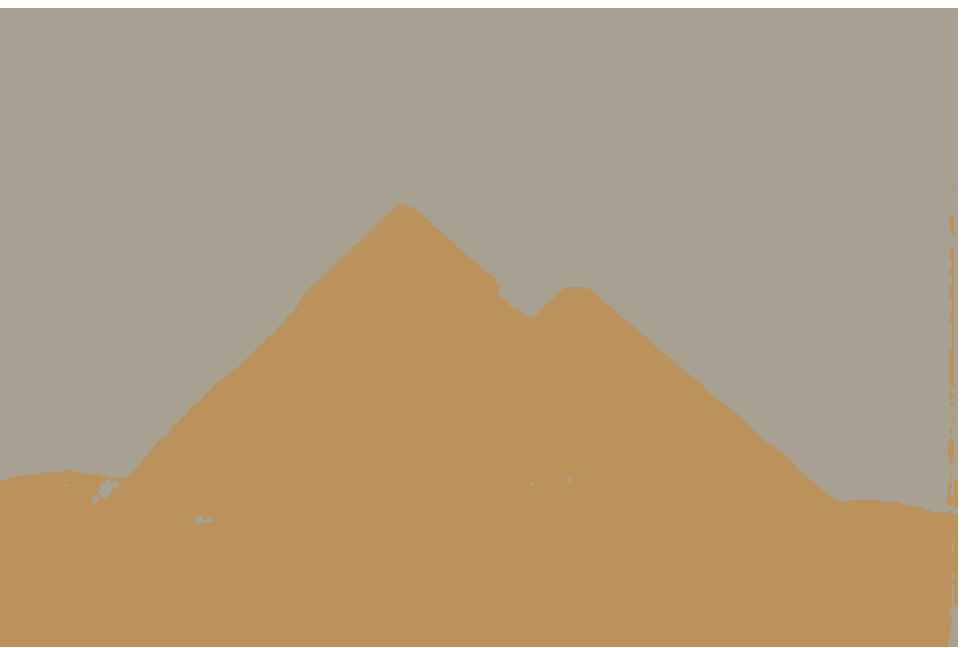}
\end{minipage}\\
\begin{minipage}[t]{0.23\linewidth}
\centering
\includegraphics[width=1\textwidth]{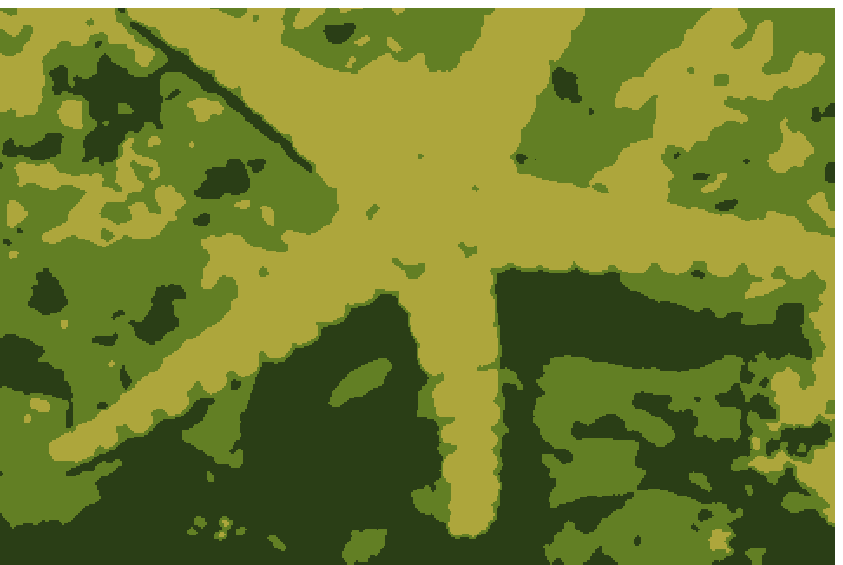}
\end{minipage}
\begin{minipage}[t]{0.23\linewidth}
\centering
\includegraphics[width=1\textwidth]{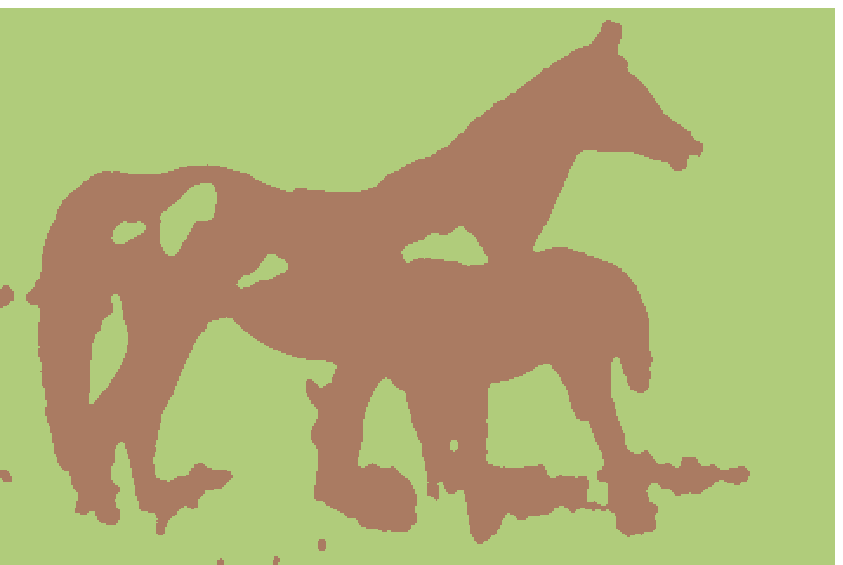}
\end{minipage}
\begin{minipage}[t]{0.23\linewidth}
\centering
\includegraphics[width=1\textwidth]{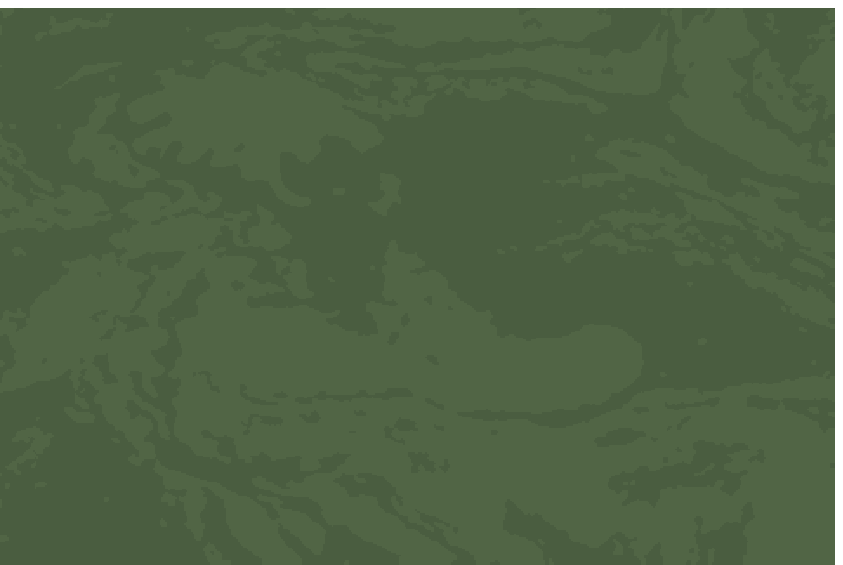}
\end{minipage}
\begin{minipage}[t]{0.23\linewidth}
\centering
\includegraphics[width=1\textwidth]{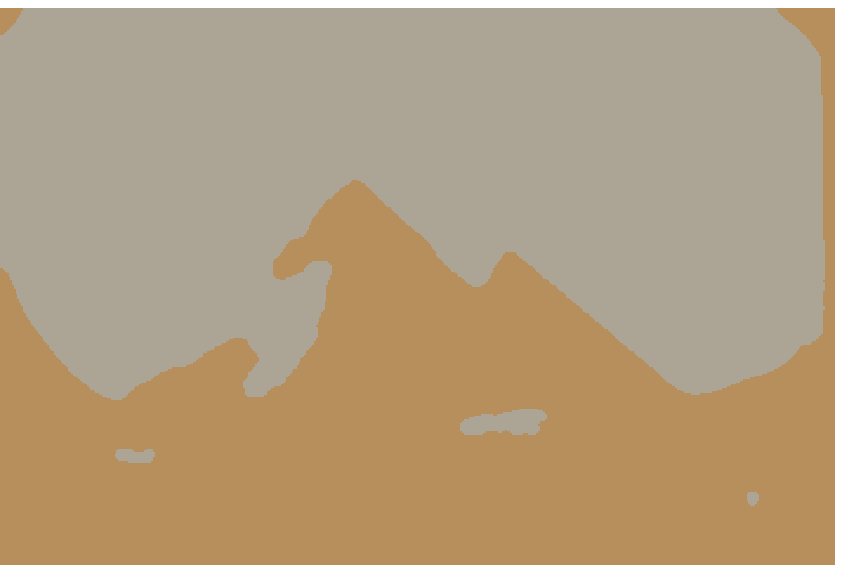}
\end{minipage}\\
\begin{minipage}[t]{0.23\linewidth}
\centering
\includegraphics[width=1\textwidth]{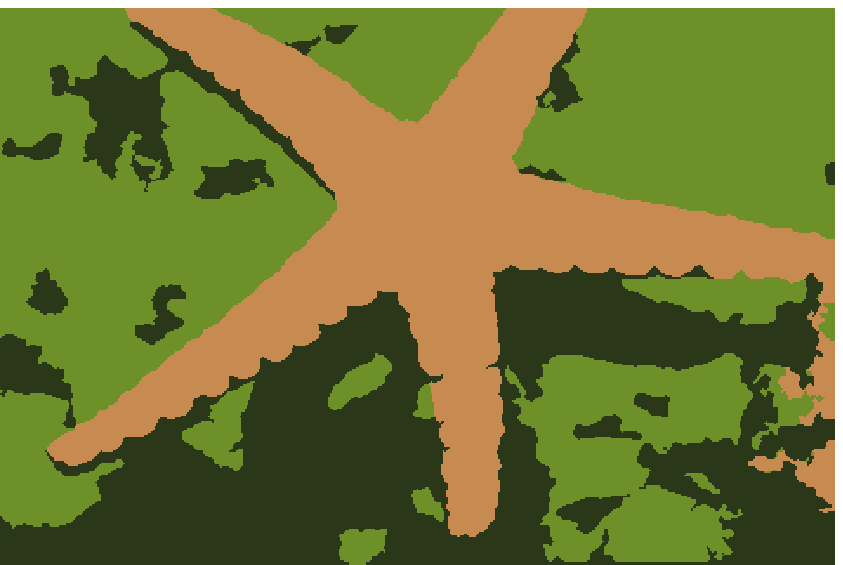}
\end{minipage}
\begin{minipage}[t]{0.23\linewidth}
\centering
\includegraphics[width=1\textwidth]{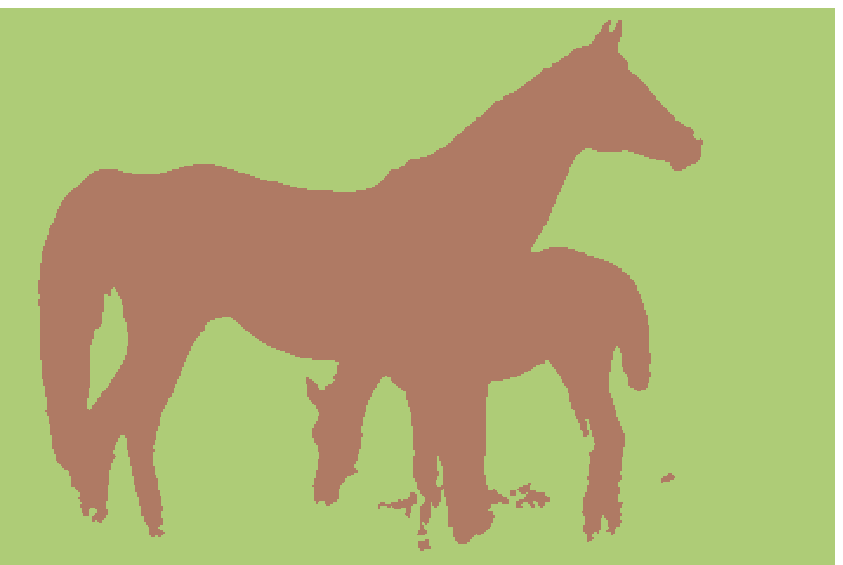}
\end{minipage}
\begin{minipage}[t]{0.23\linewidth}
\centering
\includegraphics[width=1\textwidth]{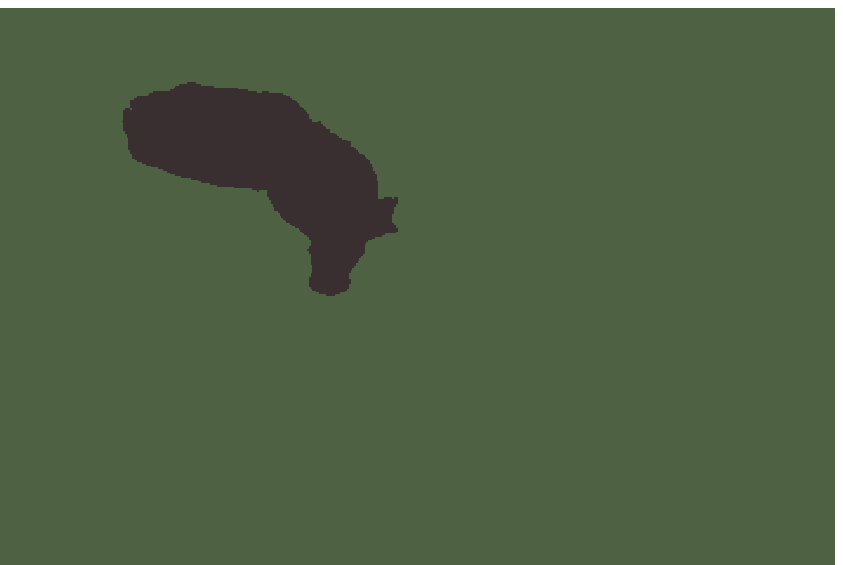}
\end{minipage}
\begin{minipage}[t]{0.23\linewidth}
\centering
\includegraphics[width=1\textwidth]{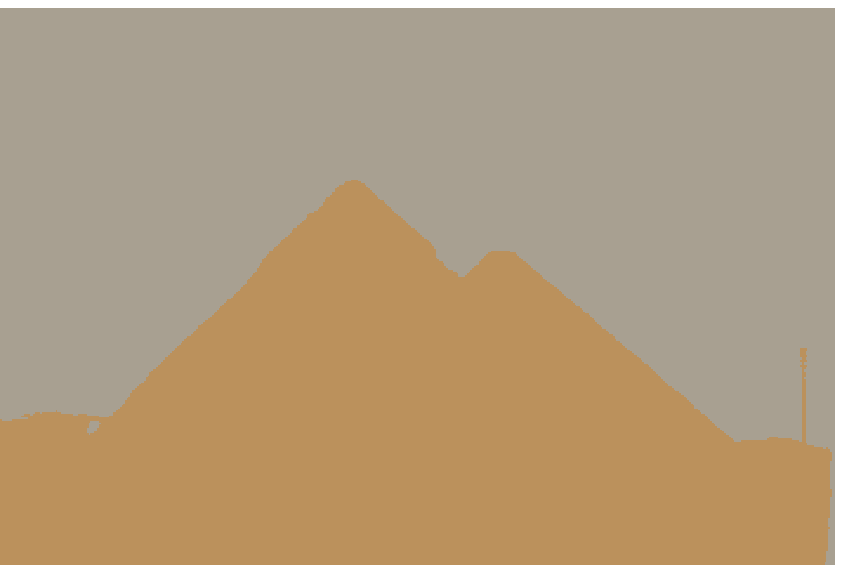}
\end{minipage}
\caption{Segmentation results for four color images in BSDS300. From top to bottom: original mages and results of FCM\_S1, FCM\_S2, FGFCM, FLICM, KWFLICM, ARKFCM, FRFCM, WFCM, DSFCM\_N, and LRFCM.}
\end{figure}

According to Fig. 8, it is found that LRFCM exhibits the best segmentation effects among all algorithms. It can not only retain true contours but also suppress clutter in color images. In contrast, FCM\_S1, FCM\_S2, FGFCM, FLICM, KWFLICM and ARKFCM achieve unsatisfactory visual results. The six algorithms cannot preserve clear contours while losing important image details. Superior to them, FRFCM and WFCM retain a large proportion of shapes. However, there are still a small amount of clutter in the segmentation results of FRFCM and WFCM. In addition, DSFCM\_N has unstable segmentation performance. As shown in the penultimate row of Fig. 8, even though DSFCM\_N has slightly better ability to track the contours of the first two images, it cannot exhibit good segmentation effects for the last two images.

Besides simulated color images in BSDS300, we also try to segment real-world images borrowed from the NASA Earth Observation database: \url{http://neo.sci.gsfc.nasa.gov/}. Generally speaking, there exist different levels of unknown noise in sampled images, which result from bit errors appearing in satellite measurements. Here, we select two real-world images. The corresponding segmentation results are illustrated in Figs. 9 and 10.

\begin{figure}[htb]
\centering
\begin{minipage}[t]{0.3\linewidth}
\centering
\includegraphics[width=1\textwidth]{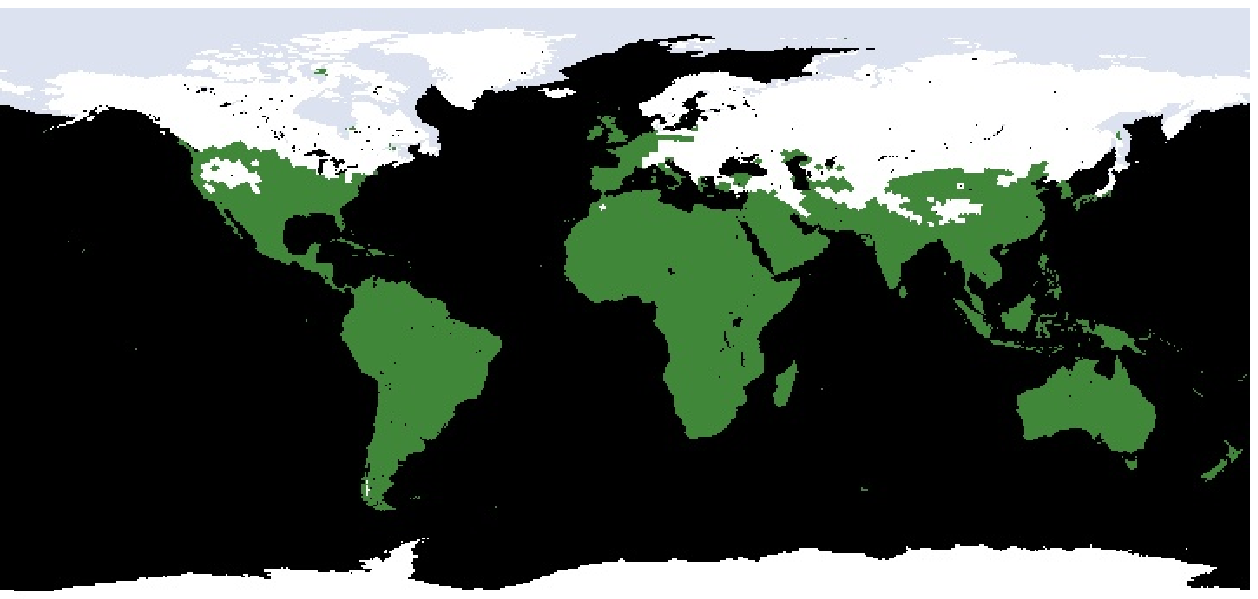}
\centerline{(a)}
\end{minipage}
\begin{minipage}[t]{0.3\linewidth}
\centering
\includegraphics[width=1\textwidth]{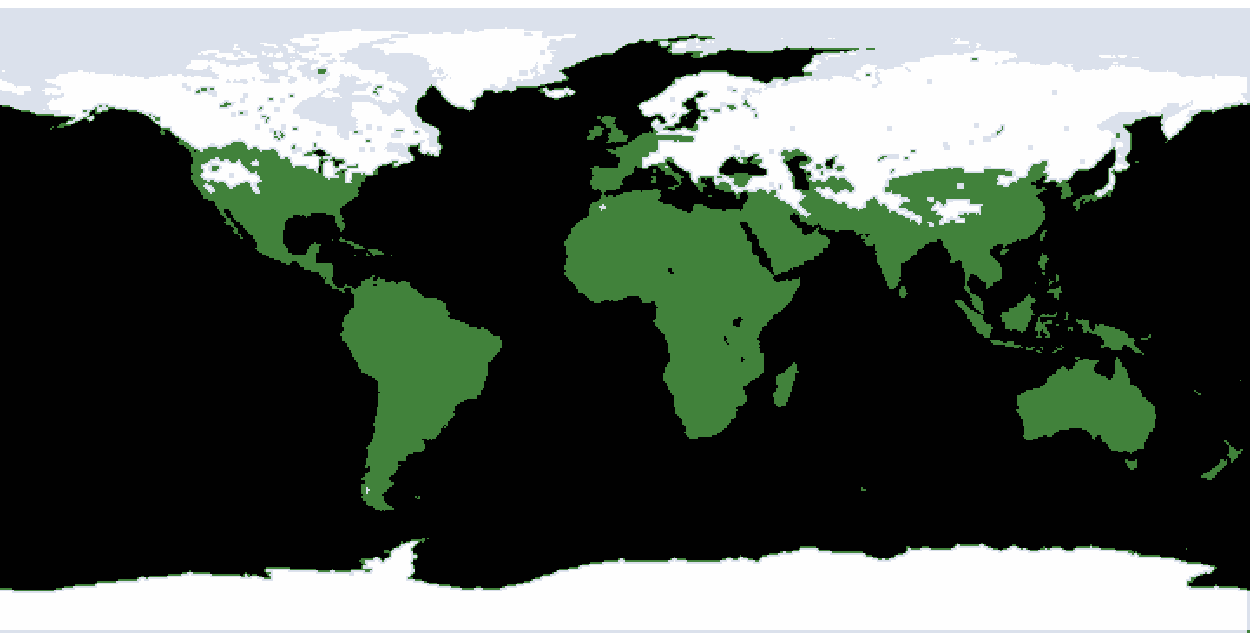}
\centerline{(b)}
\end{minipage}
\begin{minipage}[t]{0.3\linewidth}
\centering
\includegraphics[width=1\textwidth]{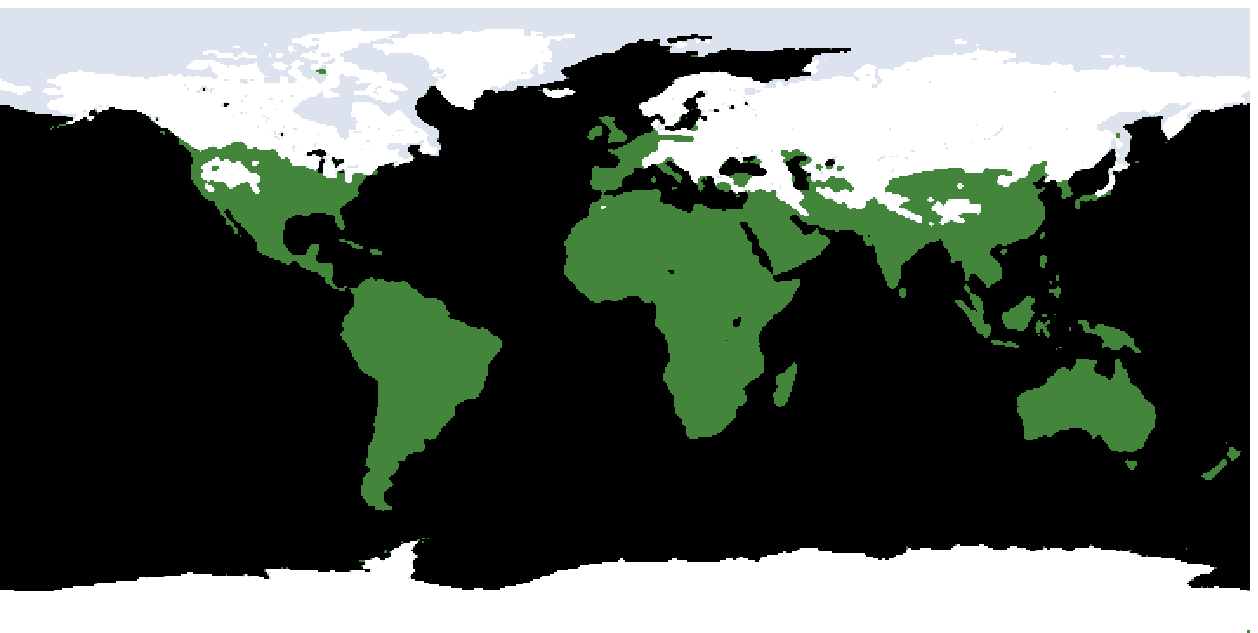}
\centerline{(c)}
\end{minipage}
\begin{minipage}[t]{0.3\linewidth}
\centering
\includegraphics[width=1\textwidth]{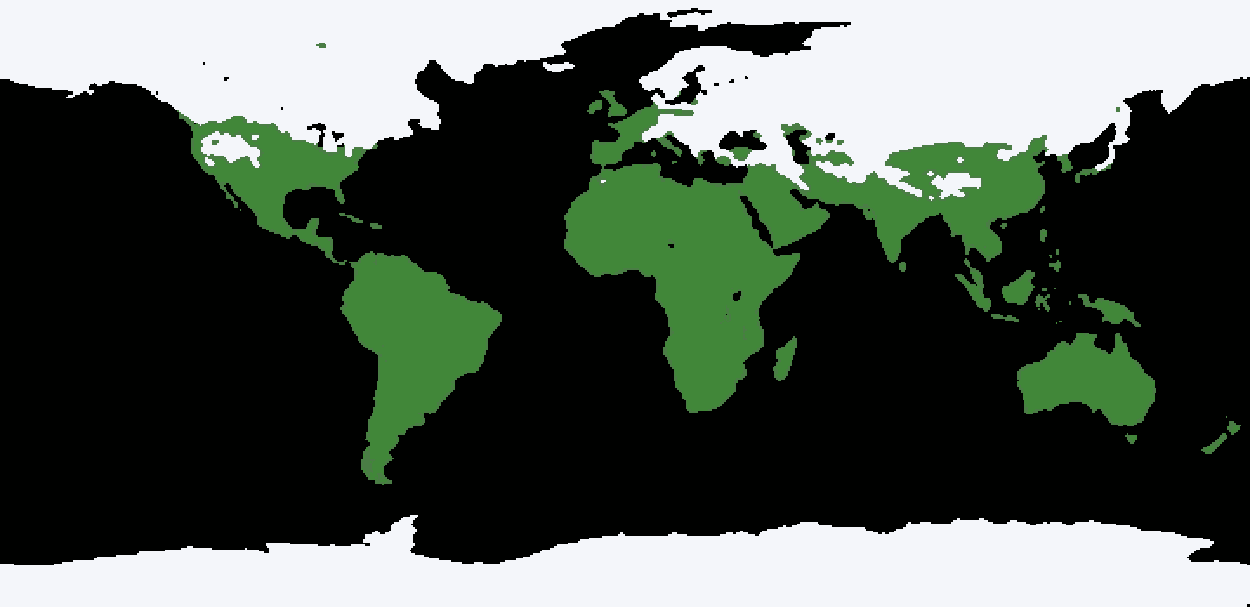}
\centerline{(d)}
\end{minipage}
\begin{minipage}[t]{0.3\linewidth}
\centering
\includegraphics[width=1\textwidth]{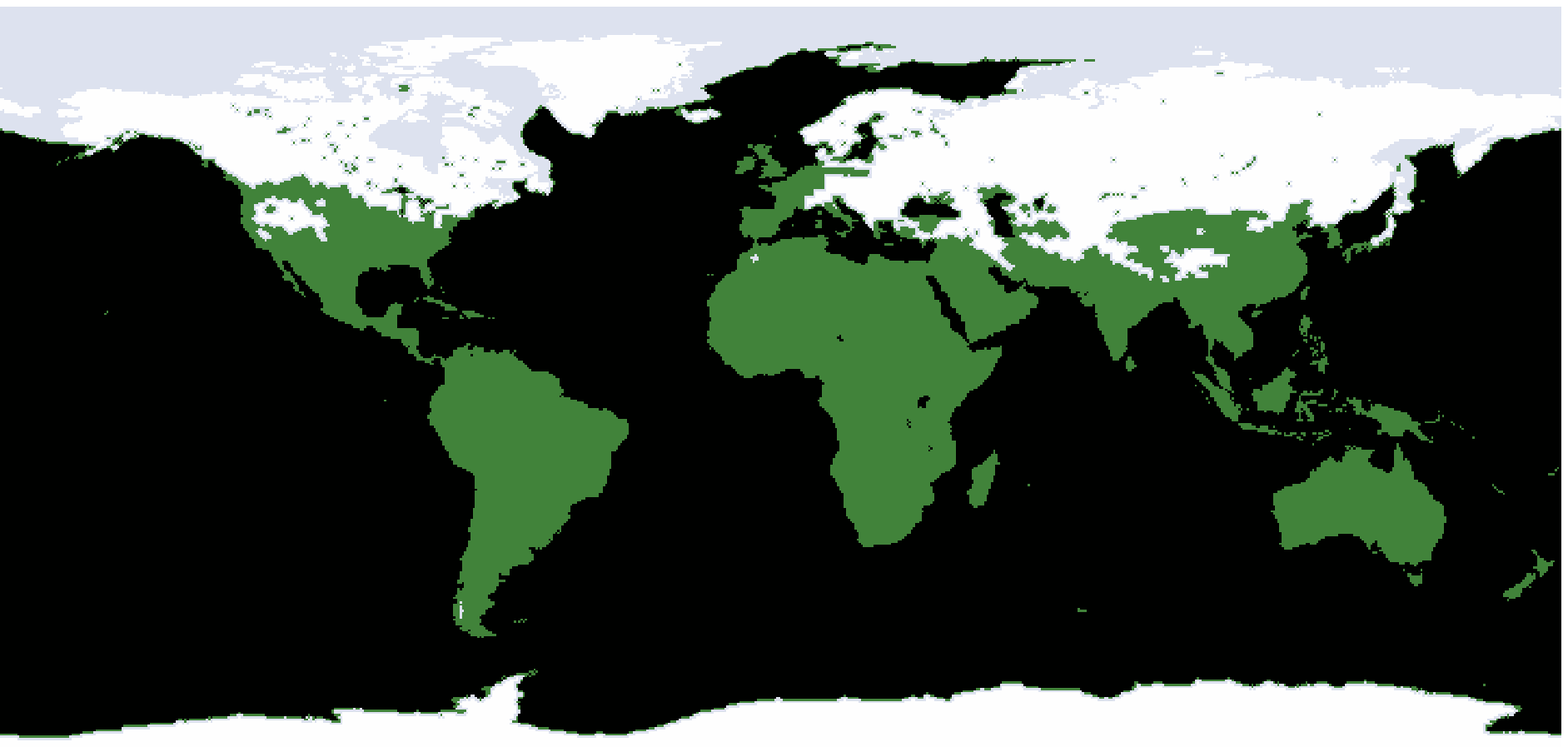}
\centerline{(e)}
\end{minipage}
\begin{minipage}[t]{0.3\linewidth}
\centering
\includegraphics[width=1\textwidth]{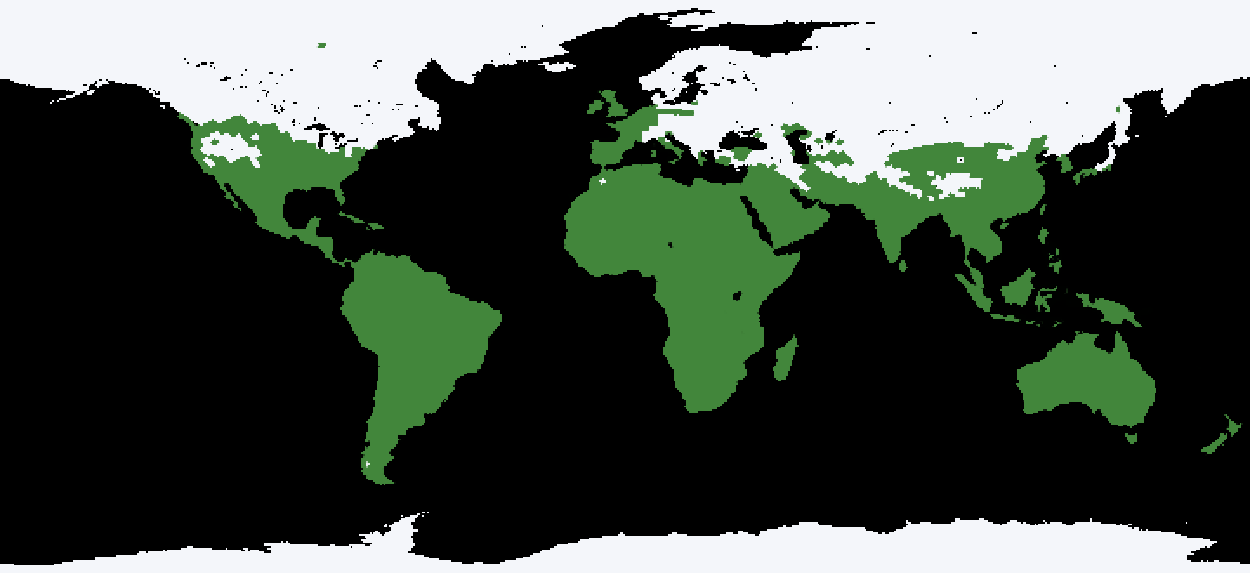}
\centerline{(f)}
\end{minipage}
\begin{minipage}[t]{0.3\linewidth}
\centering
\includegraphics[width=1\textwidth]{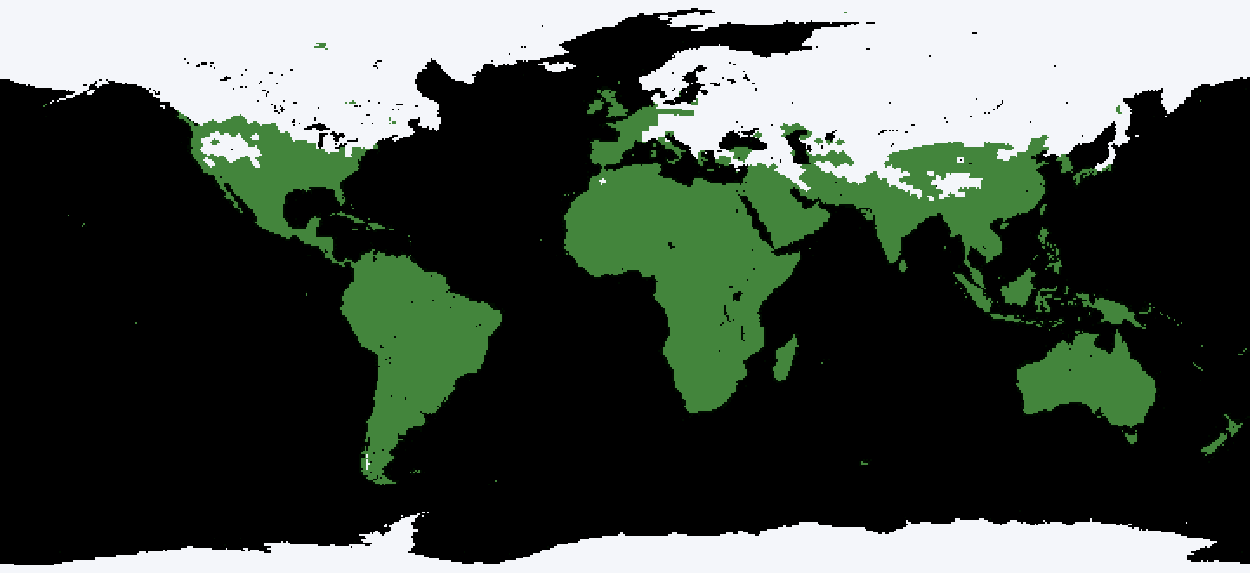}
\centerline{(g)}
\end{minipage}
\begin{minipage}[t]{0.3\linewidth}
\centering
\includegraphics[width=1\textwidth]{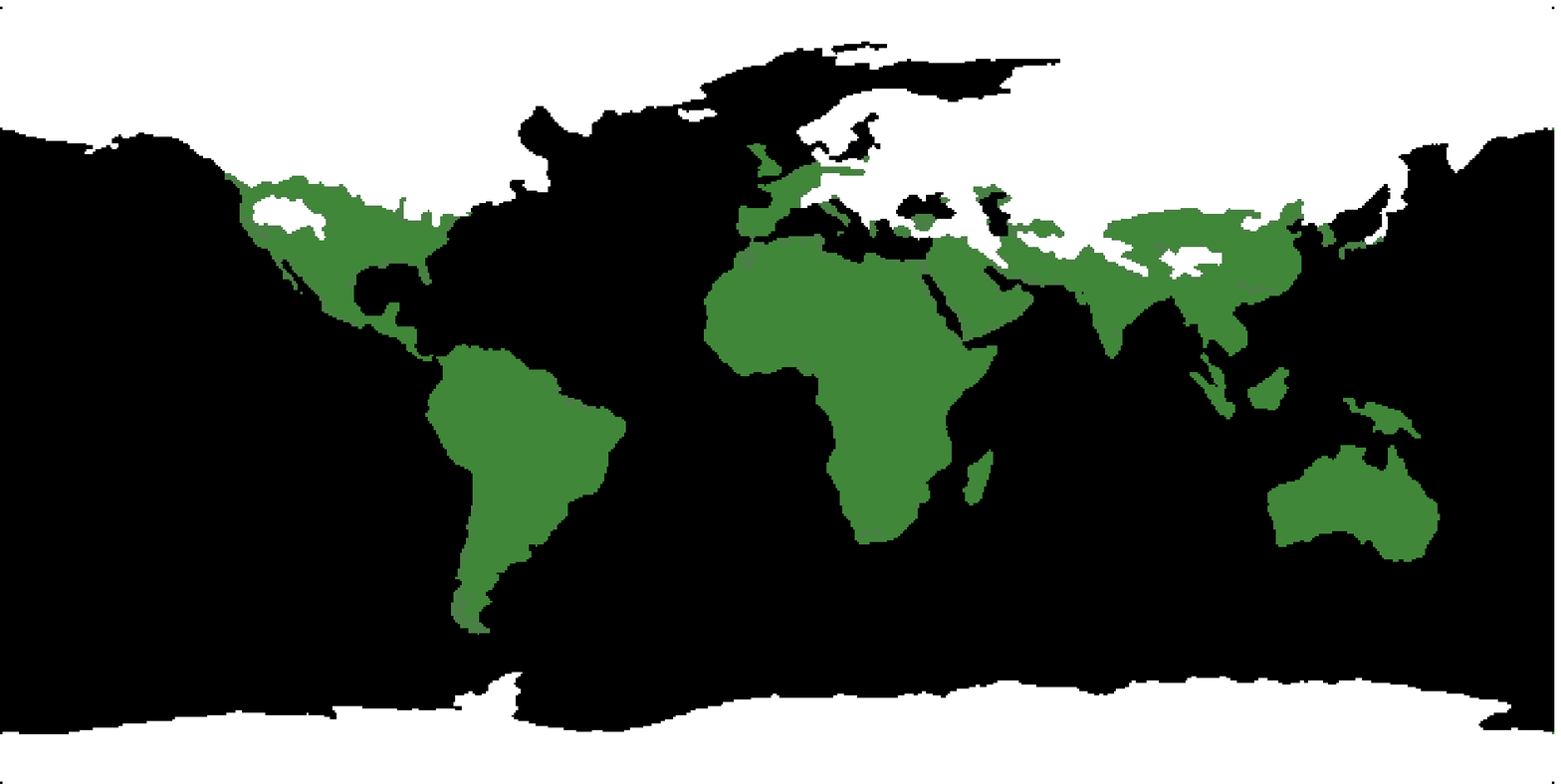}
\centerline{(h)}
\end{minipage}
\begin{minipage}[t]{0.3\linewidth}
\centering
\includegraphics[width=1\textwidth]{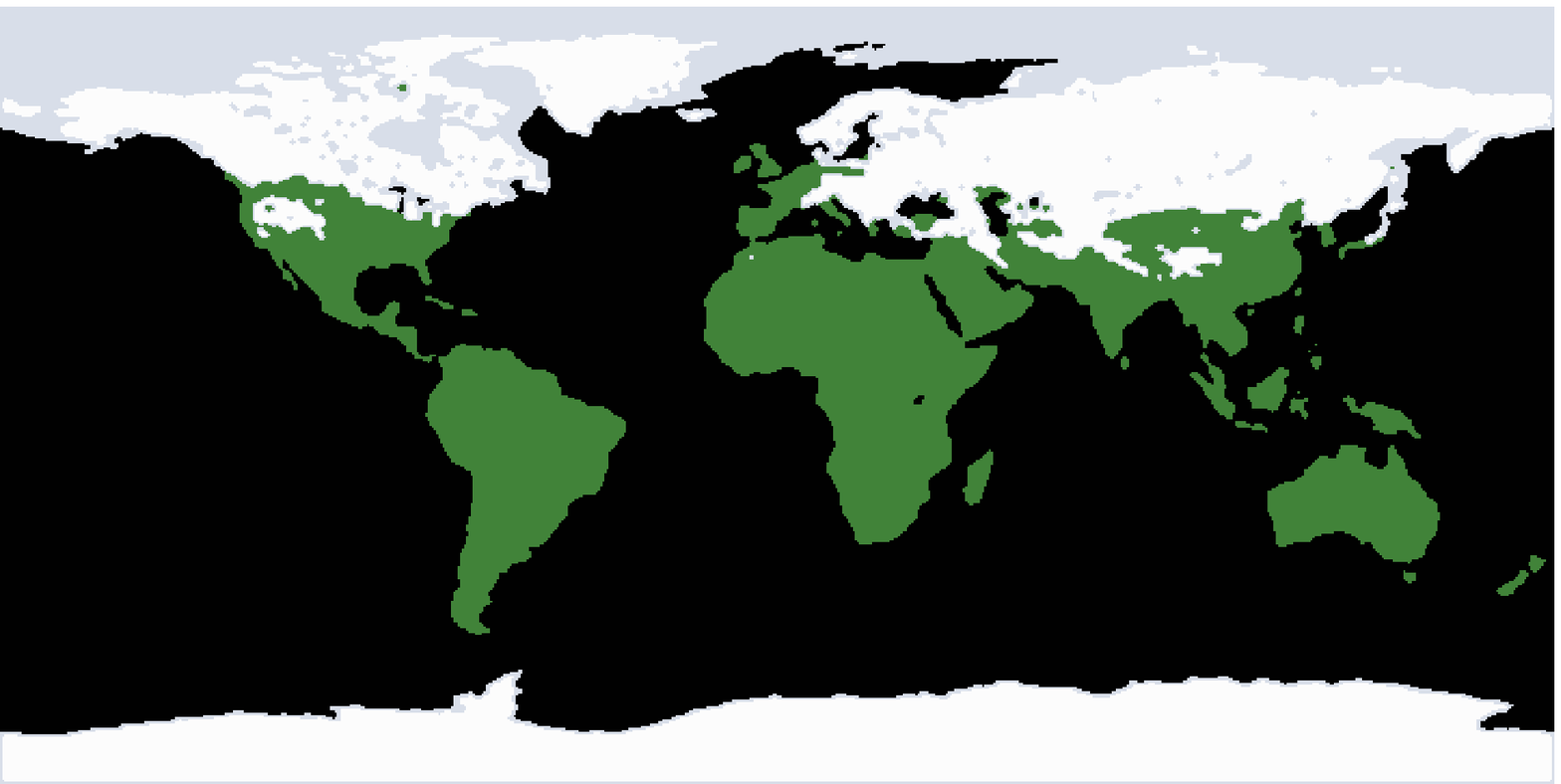}
\centerline{(i)}
\end{minipage}
\begin{minipage}[t]{0.3\linewidth}
\centering
\includegraphics[width=1\textwidth]{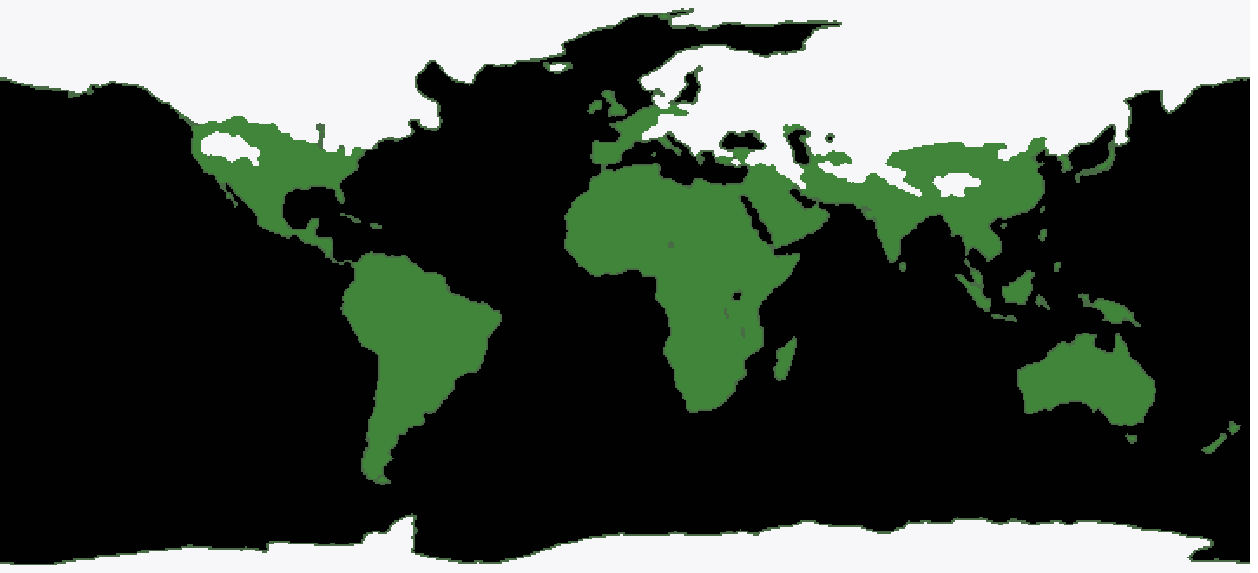}
\centerline{(j)}
\end{minipage}
\begin{minipage}[t]{0.3\linewidth}
\centering
\includegraphics[width=1\textwidth]{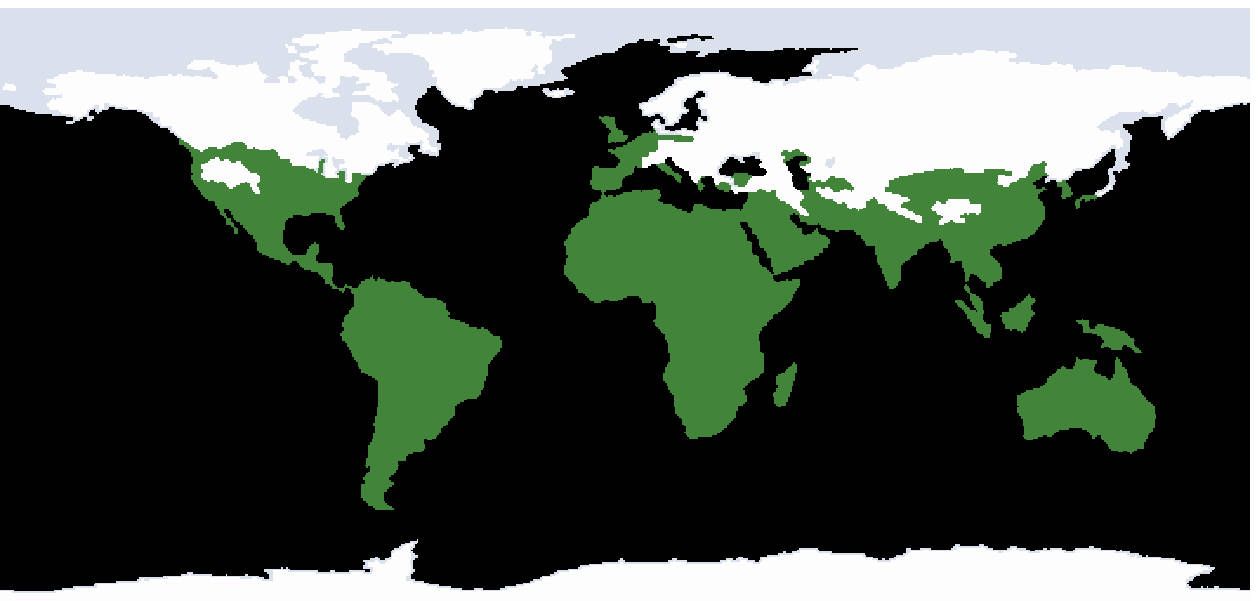}
\centerline{(k)}
\end{minipage}
\caption{Segmentation results for the first real-world image. From (a) to (k): original image and results of FCM\_S1, FCM\_S2, FGFCM, FLICM, KWFLICM, ARKFCM, FRFCM, WFCM, DSFCM\_N, and LRFCM.}
\end{figure}

Fig. 9 shows the results for segmenting a real-world image showing sea ice and snow extent. The colors represent where the land and ocean are covered by snow and ice per week (here is February 7--14, 2015). The number of clusters is set to 2. Obviously, most of algorithms, i.e., FCM\_S1, FCM\_S2, FGFCM, FLICM, KWFLICM, ARKFCM, and WFCM, cannot fully suppress unknown noise. Both FRFCM and DSFCM\_N have an aptitude for noise suppression. However, they produce incorrect clusters. Compared with nine peers, LRFCM can remove unknown noise and preserve image contours as well.

\begin{figure}[htb]
\centering
\begin{minipage}[t]{0.3\linewidth}
\centering
\includegraphics[width=1\textwidth]{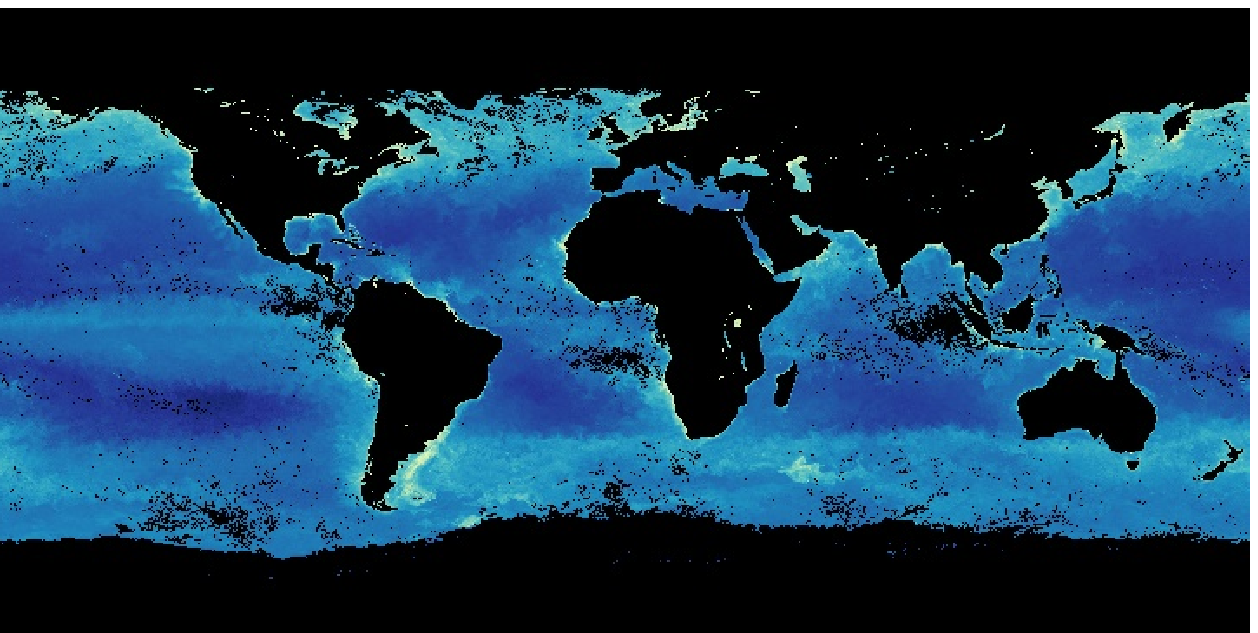}
\centerline{(a)}
\end{minipage}
\begin{minipage}[t]{0.3\linewidth}
\centering
\includegraphics[width=1\textwidth]{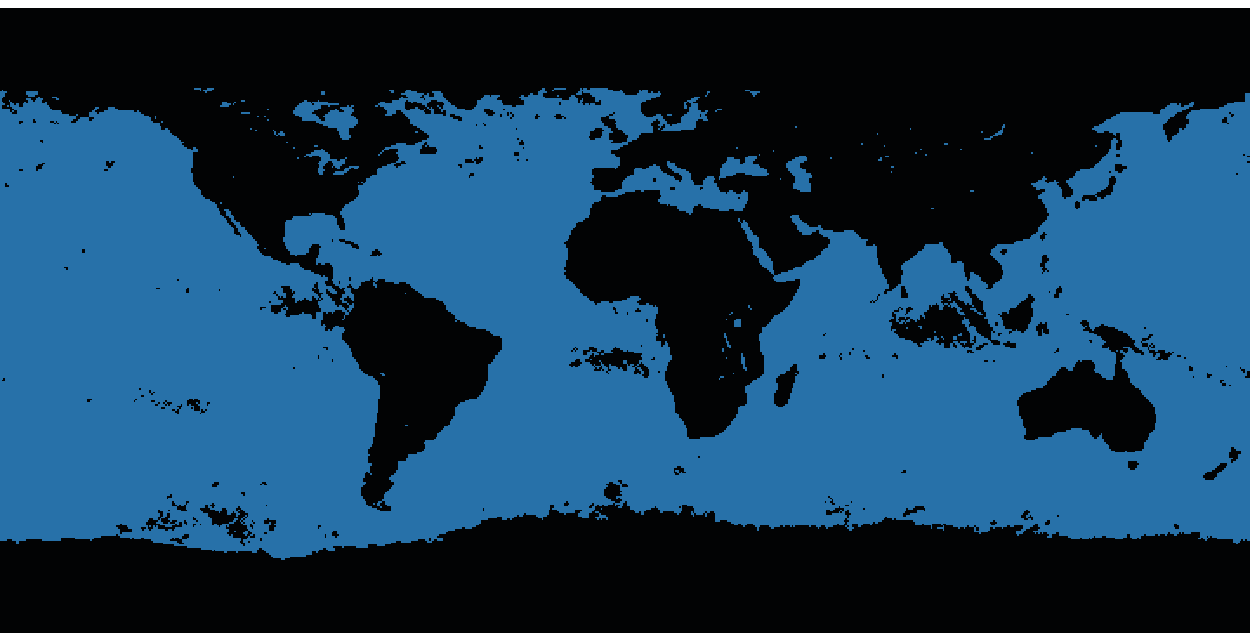}
\centerline{(b)}
\end{minipage}
\begin{minipage}[t]{0.3\linewidth}
\centering
\includegraphics[width=1\textwidth]{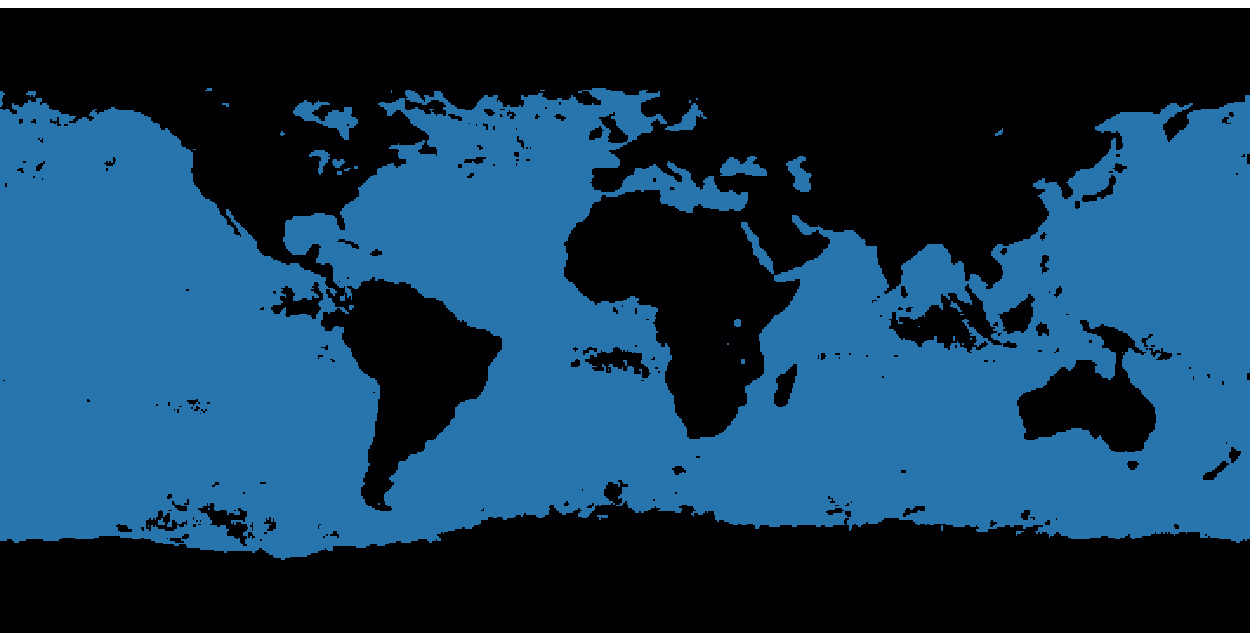}
\centerline{(c)}
\end{minipage}
\begin{minipage}[t]{0.3\linewidth}
\centering
\includegraphics[width=1\textwidth]{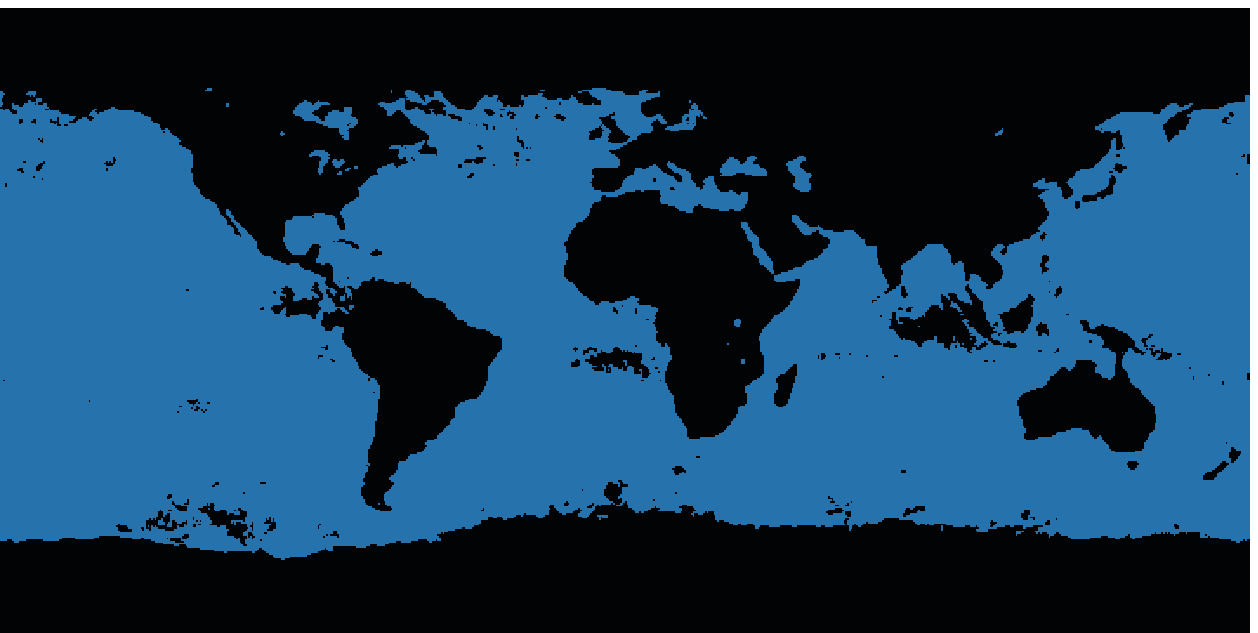}
\centerline{(d)}
\end{minipage}
\begin{minipage}[t]{0.3\linewidth}
\centering
\includegraphics[width=1\textwidth]{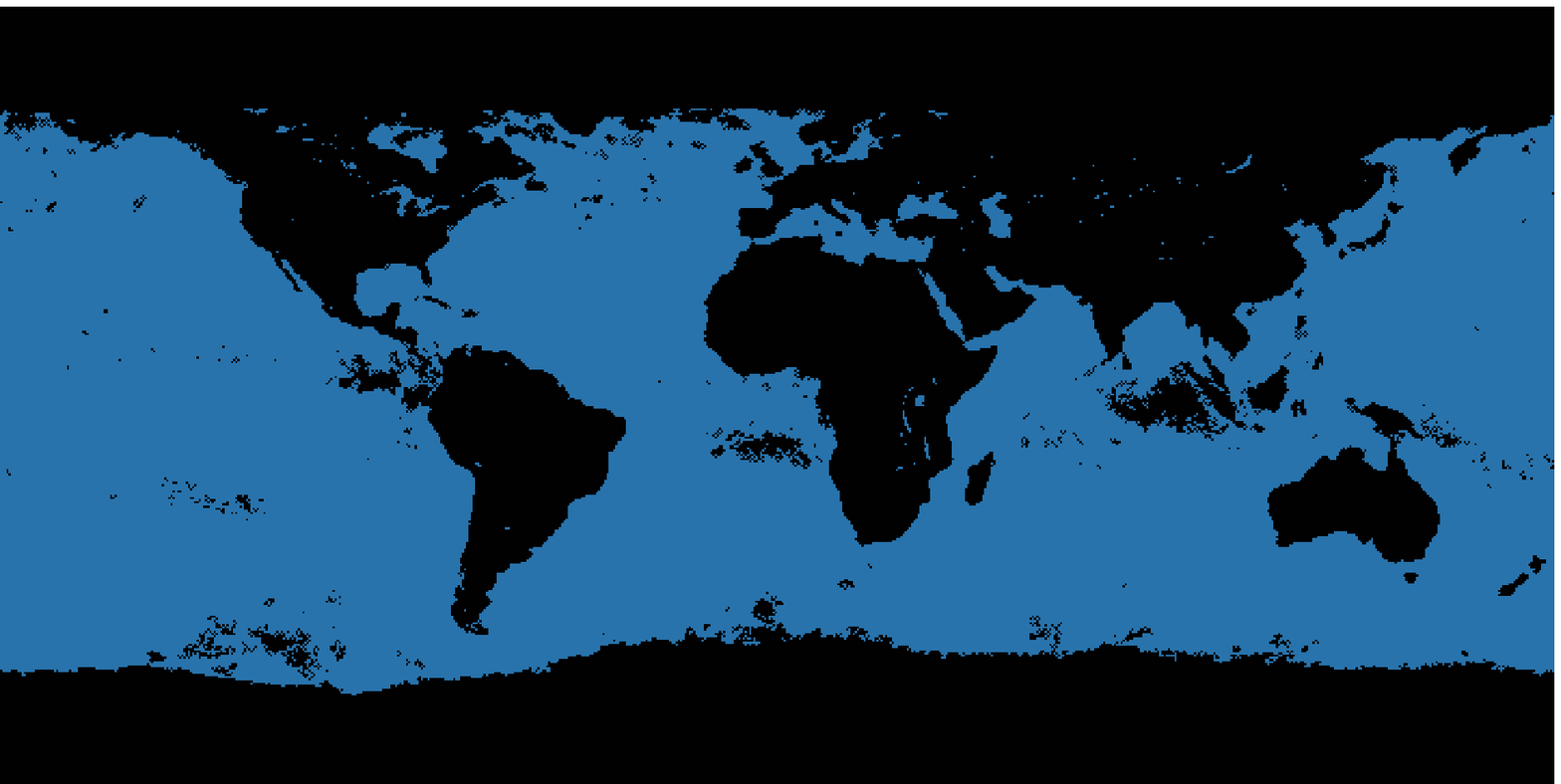}
\centerline{(e)}
\end{minipage}
\begin{minipage}[t]{0.3\linewidth}
\centering
\includegraphics[width=1\textwidth]{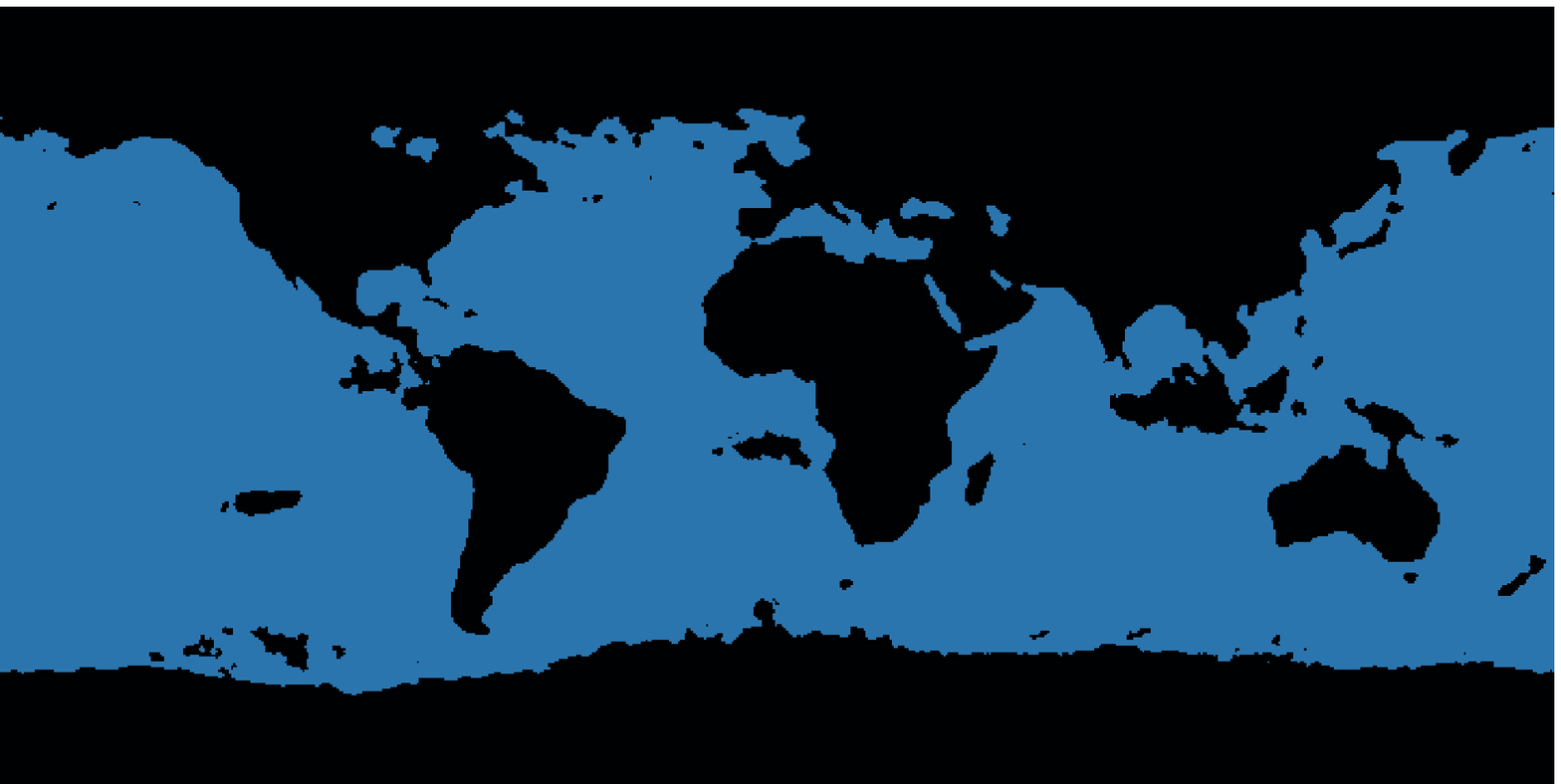}
\centerline{(f)}
\end{minipage}
\begin{minipage}[t]{0.3\linewidth}
\centering
\includegraphics[width=1\textwidth]{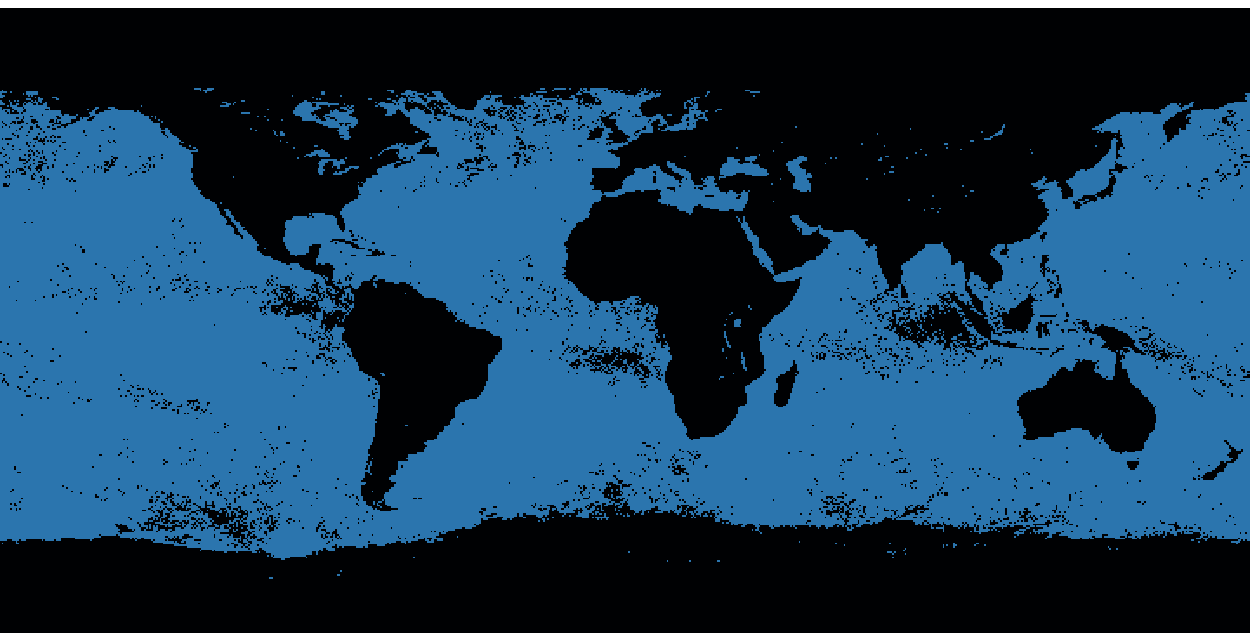}
\centerline{(g)}
\end{minipage}
\begin{minipage}[t]{0.3\linewidth}
\centering
\includegraphics[width=1\textwidth]{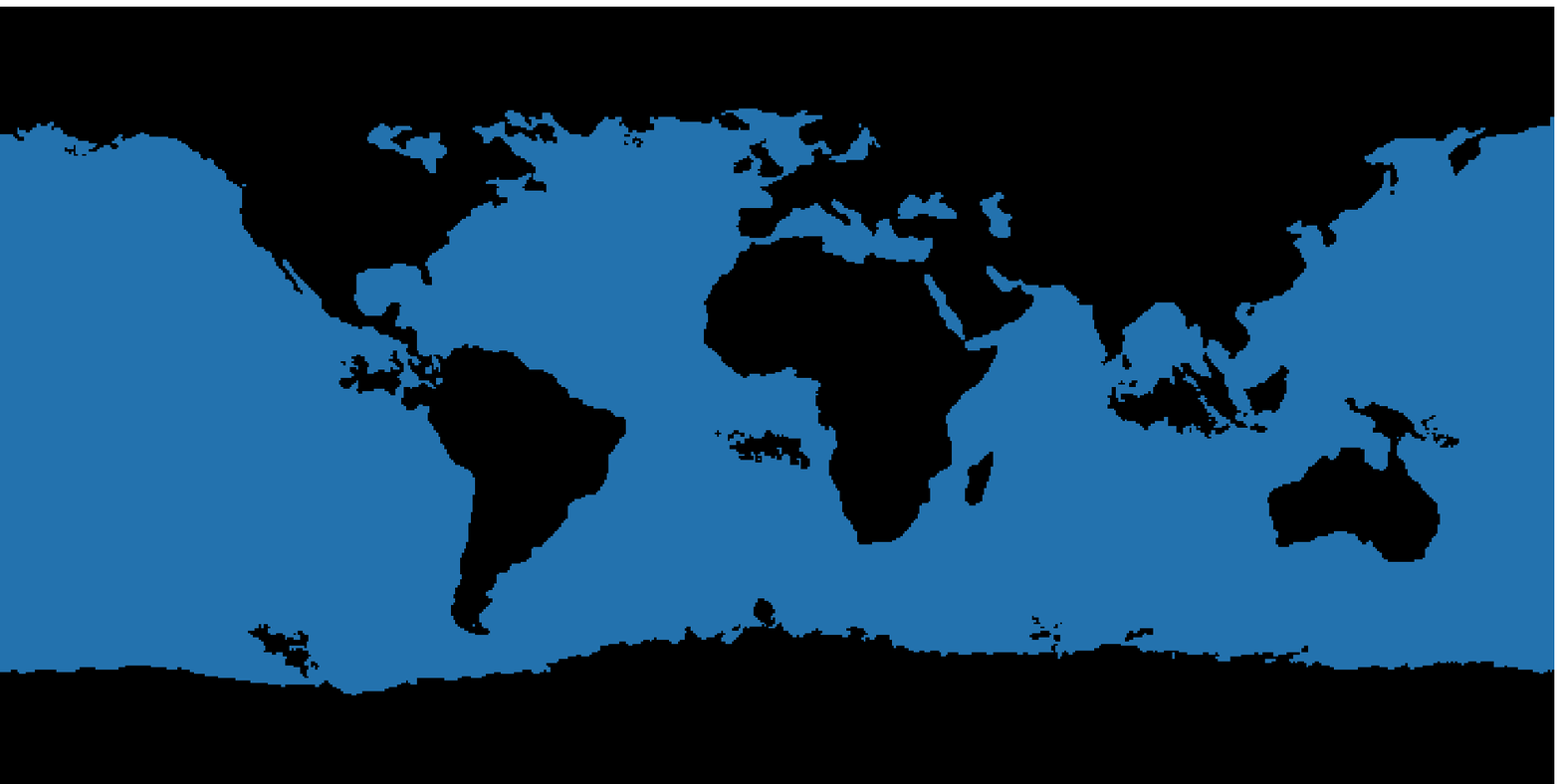}
\centerline{(h)}
\end{minipage}
\begin{minipage}[t]{0.3\linewidth}
\centering
\includegraphics[width=1\textwidth]{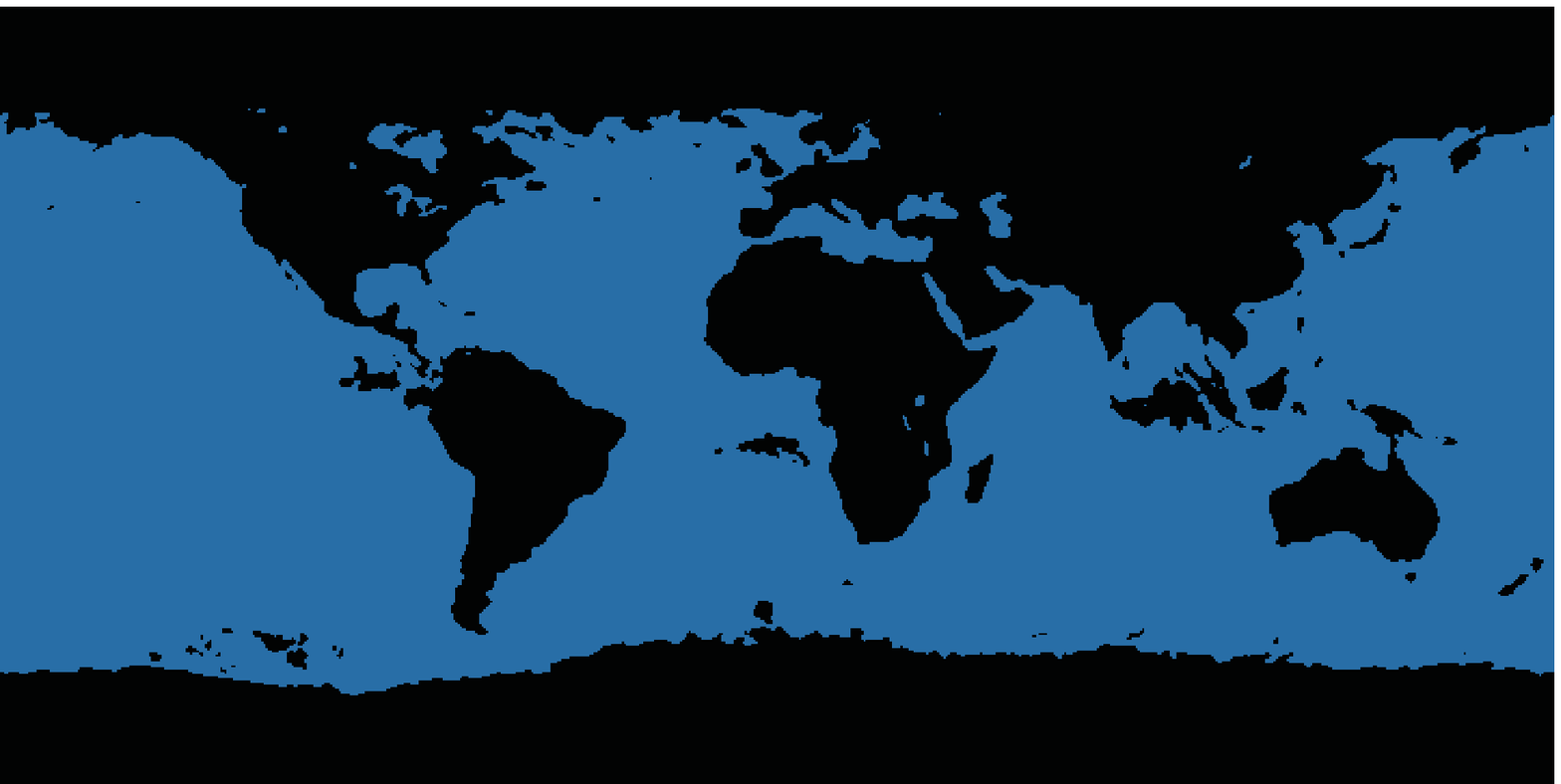}
\centerline{(i)}
\end{minipage}
\begin{minipage}[t]{0.3\linewidth}
\centering
\includegraphics[width=1\textwidth]{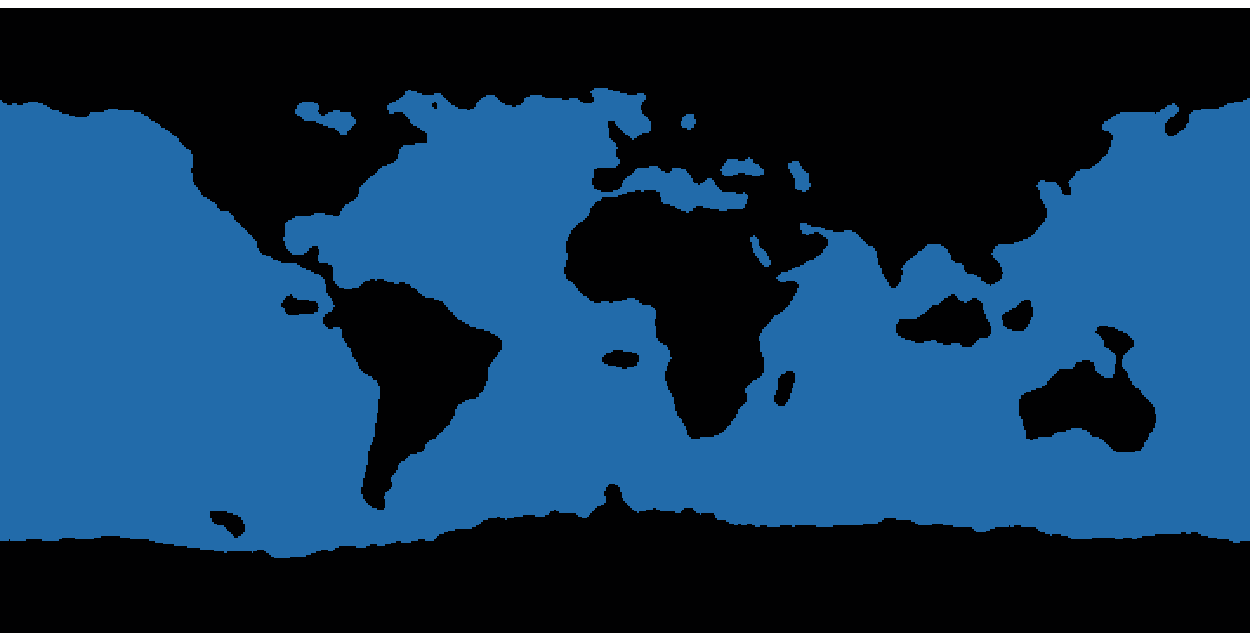}
\centerline{(j)}
\end{minipage}
\begin{minipage}[t]{0.3\linewidth}
\centering
\includegraphics[width=1\textwidth]{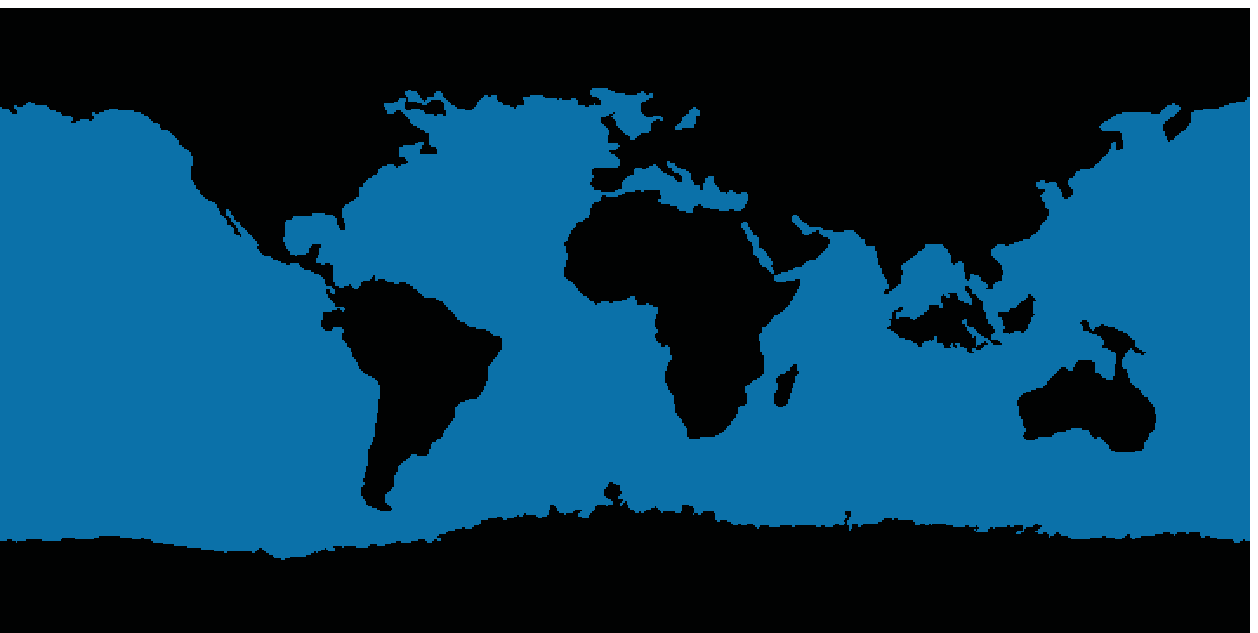}
\centerline{(k)}
\end{minipage}
\caption{Segmentation results for the second real-world image. From (a) to (k): original image and results of FCM\_S1, FCM\_S2, FGFCM, FLICM, KWFLICM, ARKFCM, FRFCM, WFCM, DSFCM\_N, and LRFCM.}
\end{figure}

Fig. 10 shows the segmentation results on a real-world image showing chlorophyll concentration. The colors represent where and how much phytoplankton are growing over a span of days. The black areas show where the satellite could not measure phytoplankton. The number of clusters is set to 2. By analyzing Fig. 10, the unknown noise is not completely eliminated according to the segmentation results of FCM\_S1, FCM\_S2, FGFCM, FLICM, and ARKFCM. Even though the remaining comparative algorithms exhibit a good capacity of noise suppression, they forge several topology changes in the form of black patches. LRFCM makes up the shortcomings of other comparative algorithms.

\subsection{Quantitative Comparisons and Analysis}

Previously, we report the visual comparison results between LRFCM and nine comparative algorithms. Besides the visual comparisons mentioned above, we proceed with quantitative comparisons by adopting two objective evaluation indicators i.e., segmentation accuracy (SA) \cite{Li2011} and entropy-based information (EI) \cite{Zhang2004}. SA is often used to assess the performance of segmenting images with known ground truth. The larger the SA value, the better the segmentation effect. Formally speaking, we have
\[{\rm SA}=\frac{\sum\limits_{i=1}^{c}|S_{i} \cap G_{i} | }{K} ,\]
where $S_{i} $ denotes the pixel set belonging to the $i$-th cluster in a segmented image, while $G_{i} $ is the pixel set belonging to the $i$-th cluster in the ground truth. $|\cdot |$ denotes the cardinality of a set.

For evaluating the performance for segmenting images without ground truth, we use:
\[\textrm{EI}=E_{1} (g'')+E_{2} (g''),\]
where $E_{1} (g'')$ stands for the expected region entropy of $g''$, i.e.,
\[E_{1} (g'')=\sum _{i=1}^{c}\left(|S_{i} |E(S_{i} )/|g''|\right) \]
while the entropy for $S_{i}$ is:
\[E(S_{i} )=-\sum _{z\in {\rm {\mathcal V}}_{i} }\left(|S_{i} (z)|/|S_{i} |\cdot \log \left(|S_{i} (z)|/|S_{i} |\right)\right) ,\]
where $S_{i} (z)$ is the subset of $S_{i} $, and its values equal \textit{z}. ${\rm {\mathcal V}}_{i} $ denotes the set including all gray level values in $S_{i} $. In addition, $E_{2} (g'')$ denotes the layout entropy of $g''$, i.e.,
\[E_{2} (g'')=-\sum _{i=1}^{c}\left(|S_{i} |\cdot\log \left(|S_{i} |/|g''|\right)/|g''|\right) .\]

The essence of indictor EI is to minimize the uniformity across all clusters by maximizing the uniformity of pixels within each segmented cluster. Hence, the better segmentation effect is associated with a smaller EI value.

For real-world images, we clarify that each image represents a specific scene. Since there exists unknow noise in these images, their reference (original) images are missing. In the sequel, indicator EI cannot directly used. To address this issue, we randomly shoot each scene for 50 times within the time span 2000--2019, which generates the mean image. It is used as the noise-free (reference) image. The calculated SA and EI values corresponding to visual results illustrated above are given in TABLE I. The results of LRFCM are highlighted by the bold letter.

\begin{table*}[htb]
  \centering
  \caption{Segmentation Results on Different Images}
    \begin{tabular}{cccccccccccc}
    \toprule
    Image &
      Indictor &
      FCM\_1 &
      FCM\_2 &
      FGFCM &
      FLICM &
      KWFLICM &
      ARKFCM &
      FRFCM &
      WFCM &
      DSFCM\_N &
      LRFCM
      \\
    \midrule
    Fig. 4 &
      SA (\%) &
      99.7803  &
      99.9125  &
      99.7375  &
      99.6948  &
      99.9878  &
      99.3713  &
      99.9832  &
      99.1226  &
      99.6643  &
      \textbf{99.9919}
      \\
    Fig. 5 &
      EI &
      1.6331  &
      1.5281  &
      1.6040  &
      1.6584  &
      1.4856  &
      1.6896  &
      1.4848  &
      1.5383  &
      1.4829  &
      \textbf{1.4821}
      \\
    Fig. 6 &
      SA (\%) &
      98.8823  &
      98.8899  &
      98.2101  &
      98.8874  &
      98.5411  &
      98.9154  &
      98.6506  &
      98.3986  &
      98.6022  &
      \textbf{99.2158}
      \\
    Fig. 7 &
      SA (\%) &
      95.4732  &
      98.6913  &
      95.6438  &
      98.7219  &
      98.5360  &
      96.3974  &
      98.5742  &
      98.5292  &
      95.6921  &
      \textbf{99.0177}
      \\
    Fig. 8 column 1 &
      EI &
      2.3696  &
      2.3676  &
      2.3932  &
      2.3928  &
      2.3811  &
      2.3564  &
      2.4688  &
      2.5204  &
      2.4806  &
      \textbf{2.5233}
      \\
    Fig. 8 column 2 &
      EI &
      2.2468  &
      2.2813  &
      2.2689  &
      2.2336  &
      2.2598  &
      2.2386  &
      2.1503  &
      2.1483  &
      2.3065  &
      \textbf{2.1389}
      \\
    Fig. 8 column 3 &
      EI &
      1.6699  &
      1.6650  &
      1.7014  &
      1.6556  &
      1.6176  &
      1.6386  &
      1.5173  &
      1.5176  &
      1.6895  &
      \textbf{1.5107}
      \\
    Fig. 8 column 4 &
      EI &
      1.9991  &
      1.9953  &
      2.0231  &
      2.1135  &
      2.1404  &
      1.9212  &
      2.0376  &
      2.0404  &
      2.1353  &
      \textbf{2.0338}
      \\
    Fig. 9 &
      EI &
      0.8713  &
      0.8516  &
      0.8840  &
      0.8592  &
      0.8312  &
      0.8286  &
      0.8146  &
      0.8172  &
      0.8725  &
      \textbf{0.8102}
      \\
    Fig. 10 &
      EI &
      1.4995  &
      1.5016  &
      1.5107  &
      1.4885  &
      1.4994  &
      1.4851  &
      1.3700  &
      1.3952  &
      1.5058  &
      \textbf{1.3520}
      \\
    \bottomrule
    \end{tabular}%
  \label{tab:addlabel}%
\end{table*}%

As TABLE I illustrates, LRFCM universally achieves the larger SA values than other nine algorithms when segmenting synthetic images with ground truth and medical images. In particular, the SA value of LRFCM comes up to 99.9919\% for the first synthetic image shown in Fig. 4. In addition, for the remaining images without ground truth, the EI values of LRFCM are universally smaller than those of other comparative algorithms, which indicates that LRFCM acquires better uniformity in segmented images. Note that the EI value of LRFCM is down to 0.8102 for the first real-world image illustrated in Fig. 9. In the light of the quantitative comparison results, we can conclude that LRFCM performs better than other FCM-related algorithms.

\subsection{Computing Overheads}
In order to exhibit the segmentation efficiency of LRFCM, we compare its computing overhead with its peers'. To ensure a fair comparison, we clarify that all experiments are completed with MATLAB running on a laptop with Intel(R) Xeon(R) W-2133 CPU of (3.60 GHz) and 32.0 GB RAM. More specifically, the computing overheads of copying with different images are shown in TABLE II. Moreover, the relevant results are intuitively illustrated in Fig. 11.

\begin{table*}[htbp]
  \centering
  \caption{Computing Overheads (in Seconds) on Different Images}
    \begin{tabular}{ccccccccccc}
    \toprule
    Image &
      FCM\_1 &
      FCM\_2 &
      FGFCM &
      FLICM &
      KWFLICM &
      ARKFCM &
      FRFCM &
      WFCM &
      DSFCM\_N &
      LRFCM
      \\
    \midrule
    Fig. 4 &
      18.6680  &
      22.6857  &
      1.3419  &
      3.3132  &
      19.7650  &
      6.3432  &
      0.2116  &
      3.5431  &
      4.6476  &
      \textbf{5.2133}
      \\
    Fig. 5 &
      12.2593  &
      13.2292  &
      1.4452  &
      3.0137  &
      22.2290  &
      4.3681  &
      0.2349  &
      2.7422  &
      3.0718  &
      \textbf{4.8040}
      \\
    Fig. 6 &
      12.9031  &
      19.7388  &
      1.3488  &
      3.6465  &
      34.3670  &
      4.4450  &
      0.2231  &
      3.6140  &
      2.9405  &
      \textbf{4.4232}
      \\
    Fig. 7 &
      11.9403  &
      14.8711  &
      1.3243  &
      3.5901  &
      43.9280  &
      4.5385  &
      0.2186  &
      3.9549  &
      3.6503  &
      \textbf{4.0271}
      \\
    Fig. 8 column 1 &
      116.6982  &
      114.3302  &
      4.9993  &
      5.5291  &
      345.2450  &
      14.0508  &
      2.3727  &
      6.2930  &
      16.8765  &
      \textbf{15.6644}
      \\
    Fig. 8 column 2 &
      39.0402  &
      49.9470  &
      3.2155  &
      4.3944  &
      78.7260  &
      9.3824  &
      0.9292  &
      4.8455  &
      13.4970  &
      \textbf{6.3864}
      \\
    Fig. 8 column 3 &
      35.3154  &
      38.4504  &
      3.2965  &
      4.5136  &
      80.6440  &
      8.2679  &
      1.6146  &
      9.3819  &
      12.5005  &
      \textbf{13.2142}
      \\
    Fig. 8 column 4 &
      70.9929  &
      70.5976  &
      3.1221  &
      6.4445  &
      138.6771  &
      8.4303  &
      0.9848  &
      5.3156  &
      18.3904  &
      \textbf{17.9287}
      \\
    Fig. 9 &
      81.6814  &
      82.1016  &
      3.8677  &
      6.8782  &
      298.4403  &
      23.4606  &
      4.9722  &
      6.2085  &
      36.3222  &
      \textbf{32.0798}
      \\
    Fig. 10 &
      44.5094  &
      41.0240  &
      2.7372  &
      4.8374  &
      53.9473  &
      13.3736  &
      1.4567  &
      6.4906  &
      18.6411  &
      \textbf{15.0150}
      \\
    \bottomrule
    \end{tabular}%
  \label{tab:addlabel}%
\end{table*}%

\begin{figure}[htb]
\centering
\begin{minipage}[t]{0.95\linewidth}
\centering
\includegraphics[width=1\textwidth]{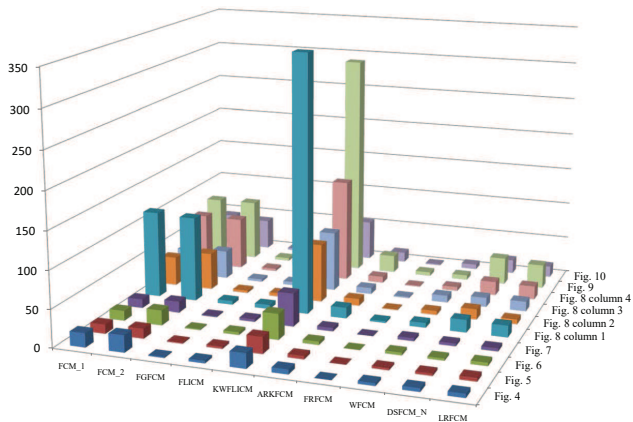}
\end{minipage}
\vspace*{-1em}
\caption{Computing overheads (in seconds) on different images.}
\end{figure}

As TABLE II and Fig. 11 indicate, for gray image segmentation, all algorithms have acceptable computing overheads. Among them, KWFLICM is the most time-consuming. In contrast, FRFCM takes the least time due to the usage of gray level histograms. Especially, LRFCM runs faster than most of its peers. When copying with color image segmentation, all algorithms incur more computational costs. Obviously, the computing overhead of KWFLICM is much higher than those of other algorithms. FRFCM still exhibits high computational efficiency. Although LRFCM is slightly slower than several comparative algorithms, its good segmentation performance makes up this shortcoming.

\subsection{Ablation Studies and Analysis}
In this subsection, we provide ablation experiments to show the effects of four key components involved in LRFCM, i.e., image filtering, feature extraction, $\ell_{0} $ regularization, and label smoothing. We impose the mixed Gaussian and impulse noise ($\textrm{standard~deviation}=30$, $\textrm{density}=20\%$) on Fig. 4(a). We set the number of clusters to 4. The analysis results are presented in TABLE III. We here clarify that symbol $\times$ represents its corresponding component is absent while symbol $\surd$ indicates the component is activated.

\begin{table}[htb]
  \centering
  \caption{Investigation of Each Component in LRFCM}
  \scriptsize
    \begin{tabular}{cccccc}
    \toprule
    \tabincell{c}{Image\\ filtering}  &
      \tabincell{c}{Feature\\extraction}  &
      \tabincell{c}{$\ell_0$\\regularization}  &
      \tabincell{c}{Label\\smoothing}  &
     {SA (\%)} &
      {Iterations}
      \\
    \midrule
      $\times$ &
      $\times$ &
      $\times$ &
      $\times$ &
      91.9734  &
      27
      \\
    $\surd$ &
      $\times$ &
      $\times$ &
      $\times$ &
      97.1793  &
      24
      \\
    $\times$ &
      $\surd$ &
      $\times$ &
      $\times$ &
      92.5247  &
      25
      \\
      $\times$ &
      $\times$ &
      $\surd$ &
      $\times$ &
      98.4379  &
      99
      \\
     $\times$ &
     $\times$ &
     $\times$ &
      $\surd$ &
      92.4194  &
      26
      \\
    $\surd$ &
      $\surd$ &
      $\surd$ &
      $\times$ &
      99.8536  &
      80
      \\
    $\surd$ &
      $\surd$ &
      $\times$ &
      $\surd$ &
      97.7691  &
      25
      \\
    $\surd$ &
      $\times$ &
      $\surd$ &
      $\surd$ &
      99.6954  &
      82
      \\
      $\times$ &
      $\surd$ &
      $\surd$ &
      $\surd$ &
      98.9722  &
      85
      \\
    $\surd$ &
      $\surd$ &
      $\surd$ &
      $\surd$ &
      99.9919  &
      80
      \\
    \bottomrule
    \end{tabular}%
  \label{tab:addlabel}%
\end{table}%

As shown in TABLE III, ten combinations of the four key components are tested. When all components are not present, the SA value is only 91.9734\%. If we consider each component alone, the SA values are increased by 5.2059\%, 0.5513\%, 6.4645\%, and 0.4460\%, respectively. Hence, the $\ell _{0} $ regularization exhibits the greatest impact on the improvement of FCM. Nevertheless, it also results in a high computing overhead. Since MR and tight wavelet fames optimize the data characteristic in advance, the computational efficiency of the proposed algorithm is still very high.

\section{Conclusions}
\label{V}
In this work, we propose a comprehensive FCM-related algorithm for image segmentation by taking advantage of various mathematical tools. By preprocessing an observed image by using MR, its weighted sum image with good properties is first generated. With the use of tight wavelet frames, the feature set associated with the weighted sum image is taken as data for clustering, which is adaptive for the analysis of image data. More importantly, an $\ell_{0}$ regularization-based FCM algorithm is proposed, which implies that the favorable estimation of the residual is obtained and the ideal value participates in clustering. In fact, the sparsity imposed on the residual is beneficial to acquire more suitable estimation, which is positive for the segmentation performance. As a post-processing step, MR is also applied to filter the obtained labels, which implies that the performance of the $\ell_{0}$ regularization-based FCM is improved. Finally, many supporting experiments are conducted to show that the proposed algorithm is superior to other FCM-related algorithms even though its running time is slightly more than a couple of compared algorithms.

Although experimental results illustrate that the proposed algorithm is effective and practical, there exist some open issues worth pursuing. For example, can one apply the proposed algorithm to non-flat domains, such as remote sensing \cite{Xu2017}, computer networks, ecological systems \cite{Wang2019ecological}, and transportation networks \cite{Lv2017}? How can one automatically select the numbers of clusters?


%

%
%

\ifCLASSOPTIONcaptionsoff
  \newpage
\fi



%

%

%
\begin{IEEEbiography}[{\includegraphics[width=1in,height=1.25in,clip,keepaspectratio]{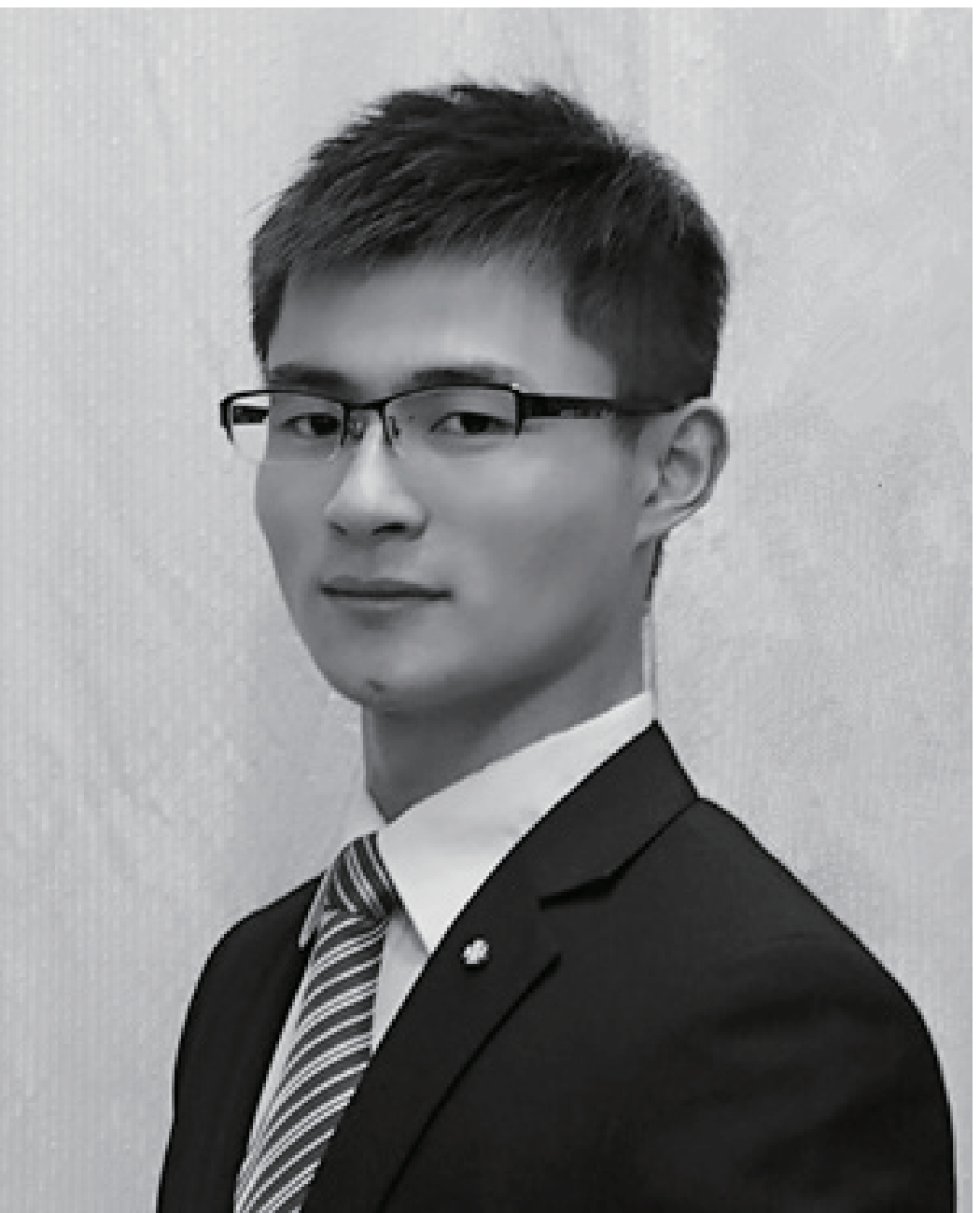}}]{Cong Wang}
received the B.S. degree in automation and the M.S. degree in mathematics from Hohai University, Nanjing, China, in 2014 and 2017, respectively. He is currently pursuing the Ph.D. degree in mechatronic engineering, Xidian University, Xi'an, China.

He was a Visiting Ph.D. Student with the Department of Electrical and Computer Engineering, University of Alberta, Edmonton, AB, Canada. He is currently a Research Assistant at the School of Computer Science and Engineering, Nanyang Technological University, Singapore. His current research interests include wavelet analysis and its applications, granular computing, and pattern recognition and image processing.
\end{IEEEbiography}

\begin{IEEEbiography}[{\includegraphics[width=1in,height=1.25in,clip,keepaspectratio]{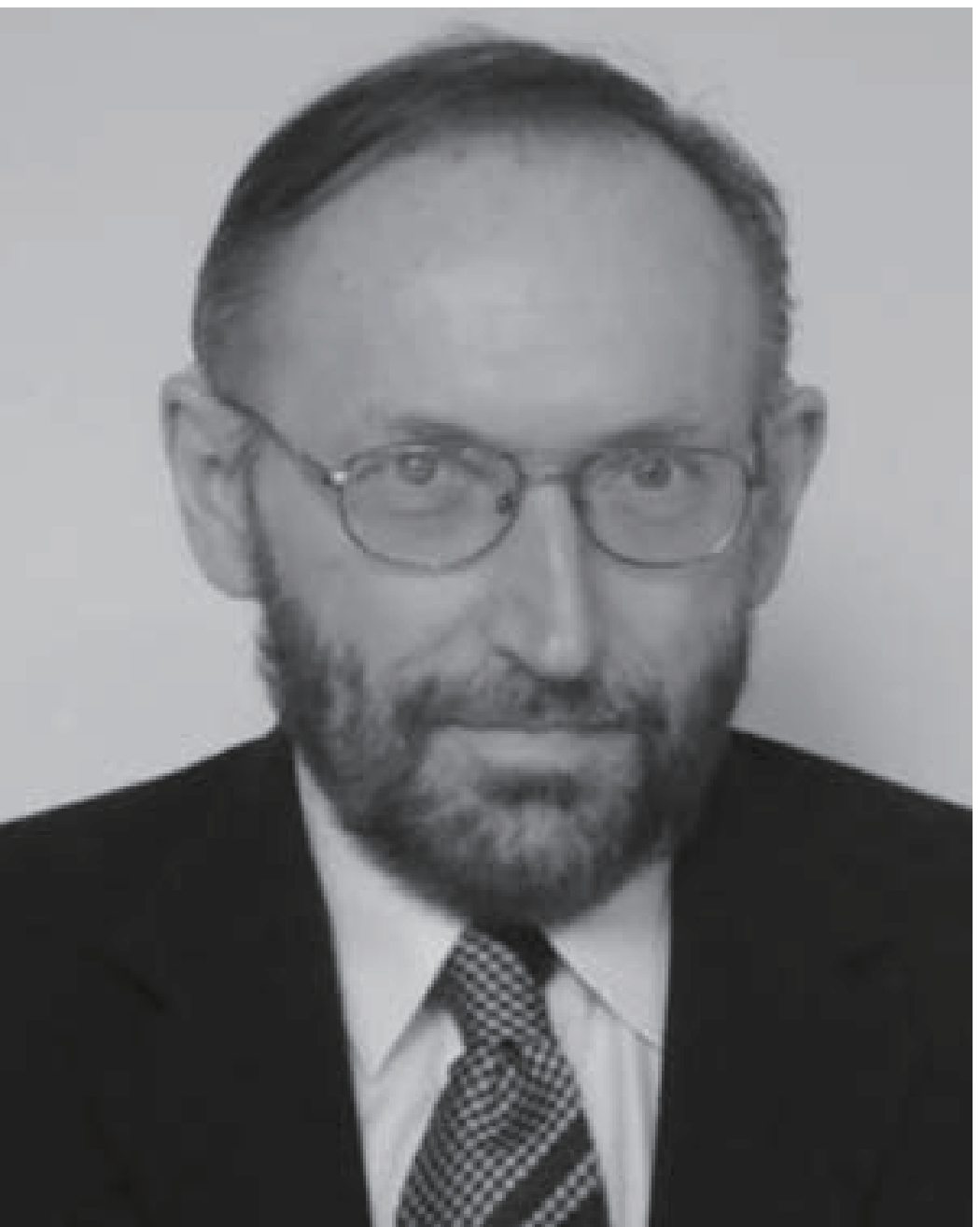}}]{Witold Pedrycz}(M'88-SM'90-F'99) received the MS.c., Ph.D., and D.Sci., degrees from the Silesian University of Technology, Gliwice, Poland.

He is a Professor and the Canada Research Chair in Computational Intelligence with the Department of Electrical and Computer Engineering, University of Alberta, Edmonton, AB, Canada. He is currently with the School of Electro-Mechanical Engineering, Xidian University, Xi'an 710071, China, and the Faculty of Engineering, King Abdulaziz University, Jeddah 21589, Saudi Arabia. He is also with the Systems Research Institute of the Polish Academy of Sciences, Warsaw, Poland. He is a foreign member of the Polish academy of Sciences. He has authored 15 research monographs covering various aspects of computational intelligence, data mining, and software engineering. His current research interests include computational intelligence, fuzzy modeling, and granular computing, knowledge discovery and data mining, fuzzy control, pattern recognition, knowledge-based neural networks, relational computing, and software engineering. He has published numerous papers in the above areas.

Dr. Pedrycz was a recipient of the IEEE Canada Computer Engineering Medal, the Cajastur Prize for Soft Computing from the European Centre for Soft Computing, the Killam Prize, and the Fuzzy Pioneer Award from the IEEE Computational Intelligence Society. 
He is a fellow of the Royal Society of Canada.
\end{IEEEbiography}

\begin{IEEEbiography}[{\includegraphics[width=1in,height=1.25in,clip,keepaspectratio]{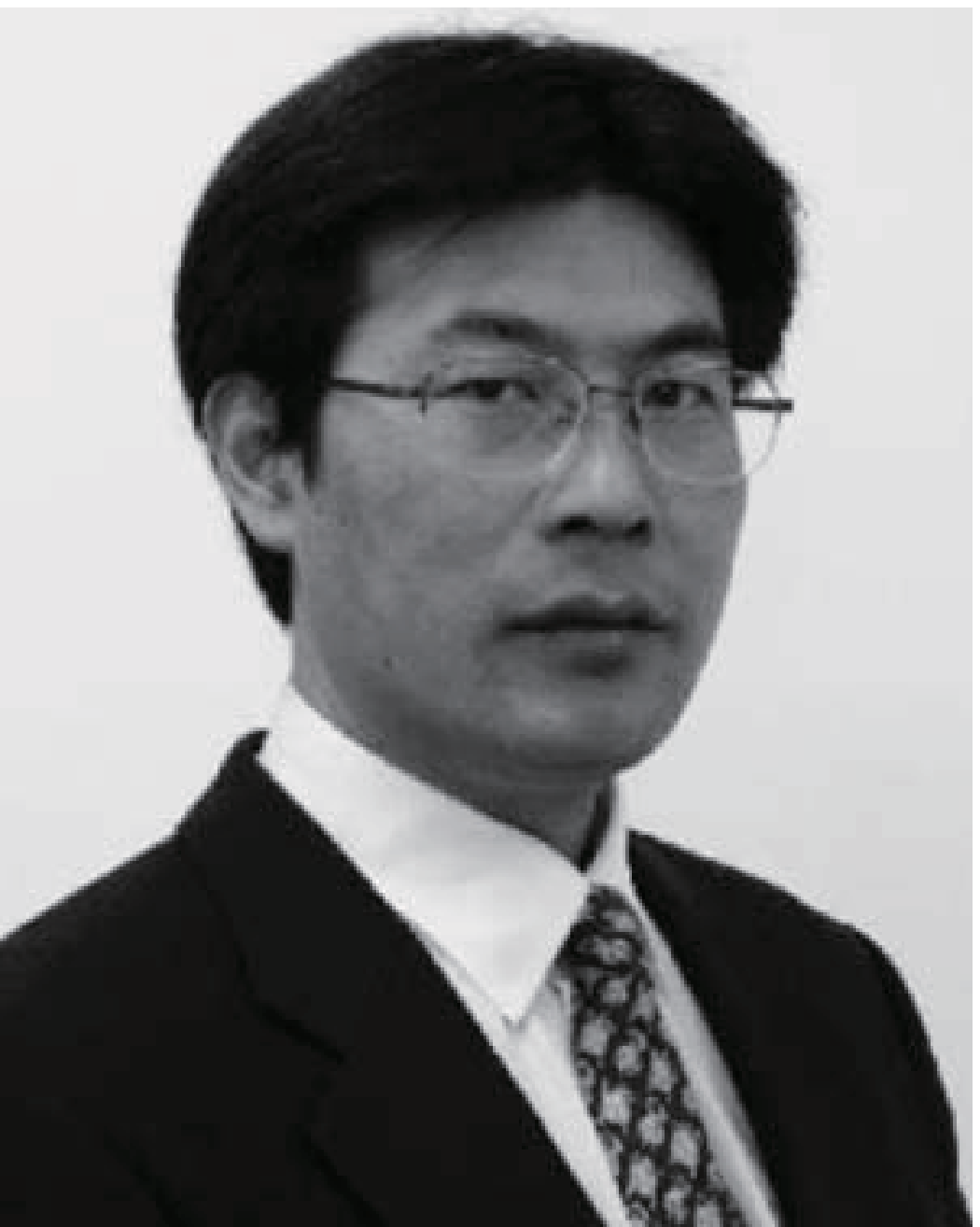}}]{ZhiWu Li}
(M'06-SM'07-F'16) 
joined Xidian University in 1992. He is currently with the Institute of Systems Engineering, Macau University of Science and Technology, Macau, China. He was a Visiting Professor with the University of Toronto, Toronto, ON, Canada, the Technion-Israel Institute of Technology, Haifa, Israel, the Martin-Luther University of Halle-Wittenburg, Halle, Germany, Conservatoire National des Arts et M\'{e}tiers, Paris, France, and Meliksah Universitesi, Kayseri, Turkey. His current research interests include Petri net theory and application, supervisory control of discrete-event systems, workflow modeling and analysis, system reconfiguration, game theory, and data and process mining.

\end{IEEEbiography}

\begin{IEEEbiography}[{\includegraphics[width=1in,height=1.25in,clip,keepaspectratio]{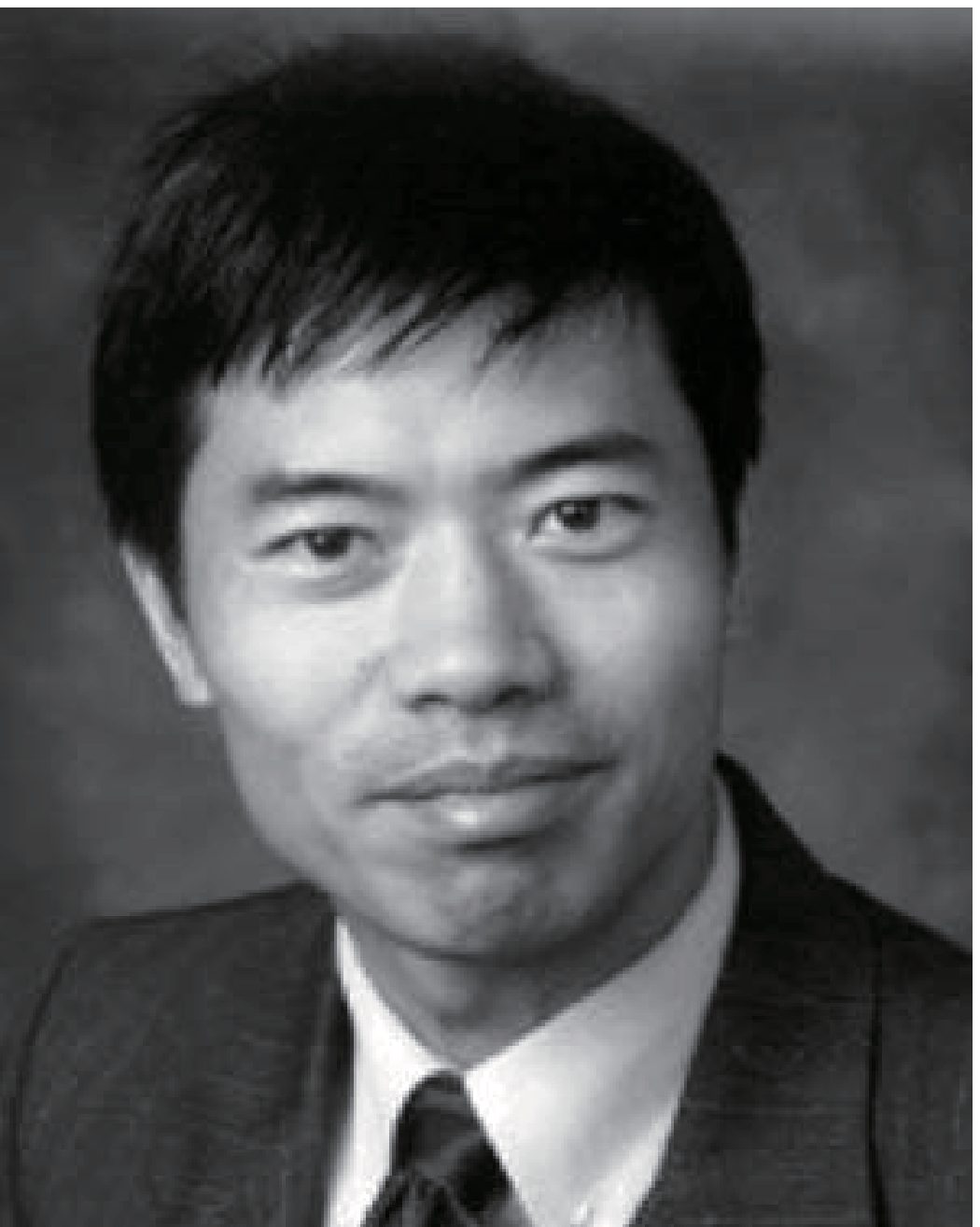}}]{MengChu Zhou}
(S'88-M'90-SM'93-F'03) 
joined New Jersey Institute of Technology (NJIT), Newark, NJ in 1990, and is now a Distinguished Professor of Electrical and Computer Engineering. 
His research interests are in Petri nets, intelligent automation, Internet of Things, big data, web services, and intelligent transportation.

He has over 800 publications including 12 books, 500+ journal papers (400+ in IEEE \textsc{Transactions}), 12 patents and 29 book-chapters. 
He is a Fellow of International Federation of Automatic Control (IFAC), American Association for the Advancement of Science (AAAS) and Chinese Association of Automation (CAA).
\end{IEEEbiography}

\begin{IEEEbiography}[{\includegraphics[width=1in,height=1.25in,clip,keepaspectratio]{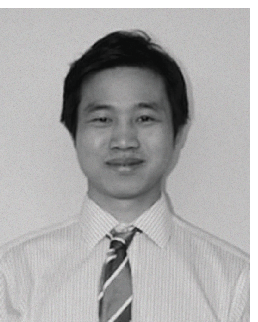}}]{Jun Zhao}
(S'10-M'15) is currently an Assistant Professor in the School of Computer Science and Engineering at Nanyang Technological University (NTU) in Singapore. He received a PhD degree in Electrical and Computer Engineering from Carnegie Mellon University (CMU) in the USA (advisors: Virgil Gligor, Osman Yagan; collaborator: Adrian Perrig) and a bachelor's degree from Shanghai Jiao Tong University in China. Before joining NTU first as a postdoc with Xiaokui Xiao and then as a faculty member, he was a postdoc at Arizona State University as an Arizona Computing PostDoc Best Practices Fellow (advisors: Junshan Zhang, Vincent Poor). His research interests include blockchains, security, and privacy with applications to the Internet of Things and deep learning. One of his first-authored papers was shortlisted for the best student paper award in IEEE International Symposium on Information Theory (ISIT) 2014, a prestigious conference in information theory.
\end{IEEEbiography}



\end{document}